\documentclass[conference,compsoc]{IEEEtran}
\IEEEoverridecommandlockouts
\usepackage{cite,soul}
\usepackage{amsmath,amssymb,amsfonts}
\usepackage{algorithmic}
\usepackage{graphicx}
\usepackage{textcomp}
\usepackage{xcolor}
\def\BibTeX{{\rm B\kern-.05em{\sc i\kern-.025em b}\kern-.08em
    T\kern-.1667em\lower.7ex\hbox{E}\kern-.125emX}}

\usepackage[utf8]{inputenc} 
\usepackage[T1]{fontenc}    
\usepackage{hyperref}       
\usepackage{url}            
\usepackage{booktabs}       
\usepackage{amsfonts}       
\usepackage{nicefrac}       

\usepackage[utf8]{inputenc} 
\usepackage[T1]{fontenc}    

\usepackage{subfig}
\usepackage{tabularx}
\usepackage{graphicx}

\begin{document}

\title{Accuracy-Privacy Trade-off in Deep Ensemble: \\ A Membership Inference Perspective
}

\author{\IEEEauthorblockN{Shahbaz Rezaei}
\IEEEauthorblockA{
\textit{University of California}\\
Davis, CA, USA \\
srezaei@ucdavis.edu}
\and
\IEEEauthorblockN{Zubair Shafiq}
\IEEEauthorblockA{
\textit{University of California}\\
Davis, CA, USA \\
zshafiq@ucdavis.edu}
\and
\IEEEauthorblockN{Xin Liu}
\IEEEauthorblockA{
\textit{University of California}\\
Davis, CA, USA \\
xinliu@ucdavis.edu}
\and
}

\maketitle

\begin{abstract}
Deep ensemble learning has been shown to improve accuracy by training multiple neural networks and averaging their outputs. Ensemble learning has also been suggested to defend against membership inference attacks that undermine privacy. In this paper, we empirically demonstrate a trade-off between these two goals, namely accuracy and privacy (in terms of membership inference attacks), in deep ensembles. Using a wide range of datasets and model architectures, we show that the effectiveness of membership inference attacks increases when ensembling improves accuracy. We analyze the impact of various factors in deep ensembles and demonstrate the root cause of the trade-off. Then, we evaluate common defenses against membership inference attacks based on regularization and differential privacy. We show that while these defenses can mitigate the effectiveness of membership inference attacks, they simultaneously degrade ensemble accuracy. We illustrate similar trade-off in more advanced and state-of-the-art ensembling techniques, such as snapshot ensembles and diversified ensemble networks. Finally, we propose a simple yet effective defense for deep ensembles to break the trade-off and, consequently, improve the accuracy and privacy, simultaneously. 
\end{abstract}


\section{Introduction}

Ensemble learning has been shown to improve classification accuracy of neural networks in particular, and machine learning classifiers in general \cite{kondratyuk2020ensembling,
kuncheva2003measures,sagi2018ensemble}. The most commonly used approach for deep models involves averaging the output of multiple neural networks (NN) that are independently trained on the same dataset with different random initialization, called \textbf{deep ensemble} \cite{lobacheva2020power}. Such a simple approach has been extensively used in practice to improve accuracy \cite{lee2015m, wang2020multiple}. Notably, a majority of the top performers in machine learning benchmarks, such as the ImageNet Large Scale Visual Recognition Challenge \cite{russakovsky2015imagenet}, have adopted some form of ensemble learning \cite{lee2015m,szegedy2015going,he2016deep}.

Interestingly, a few recent papers argue using ensemble learning to achieve a different goal rather than improving accuracy, that is, to defend against membership inference attack \cite{huang2020damia, li2021membership, rahimian2020sampling, yang2020defending}. In membership inference attack, the goal of an attacker is to infer whether a sample has been used to train a model--i.e., whether the sample belongs to the train set. In literature, several forms of ensemble learning (different from deep ensembles), such as partitioning, has been used to defend against privacy-harming membership inference (MI) attacks. Membership inference attacks generally use the prediction confidence of NN models to infer membership status of a sample \cite{salem2018ml, shokri2017membership, truex2019demystifying, yeom2018privacy} by leveraging the insight that trained models may output higher prediction confidence on train samples than non-train samples \cite{choo2020label}. The intuition behind using ensemble learning approaches, like partitioning, to defend against MI attacks is that training each model on a different subset of data makes the ensemble less prone to overfitting \cite{salem2018ml}. 
While the idea is discussed in \cite{huang2020damia, li2021membership, rahimian2020sampling, yang2020defending}, none of these papers theoretically or empirically demonstrate the usefulness of \textit{deep ensembles}, in particular, as a defense mechanism.


\begin{figure}
\centering
\includegraphics[width = 0.8\linewidth]{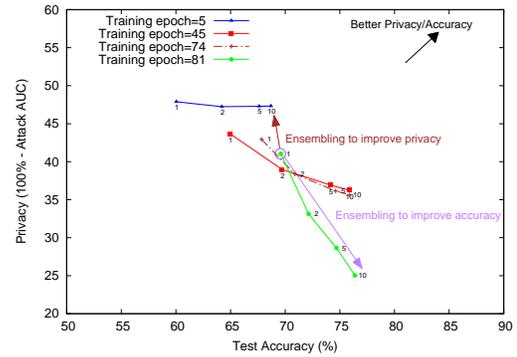}
\caption{Trade-off between accuracy and privacy on an AlexNet model trained on CIFAR10. Each curve contains four points corresponding to ensembles comprising of 1, 2, 5, and 10 base models (from left to right). Using the single model trained for 81 epochs as a baseline, there are two choices: (1) making an ensemble of these models to achieve the highest accuracy possible but worse privacy (purple arrow), or (2) making an ensemble of less overfitted models (epoch \#5) to achieve slightly lower accuracy of a single model but better privacy (brown arrow).}
\label{fig-intro}
\vspace{-.1in}
\end{figure}

In this paper, we show that these two goals of ensemble learning, namely improving accuracy and defending against MI attack, do not trivially sum up in a unified solution in deep ensembles. Figure~\ref{fig-intro} illustrates accuracy and privacy trade-off by plotting accuracy and membership inference attack effectiveness for ensembles comprising of varying number of base models (1, 2, 5, and 10) that are trained for different numbers of epochs (5, 45, 74 and 81). The training epoch is chosen such that the accuracy of a single model best aligns with the accuracy of an ensemble. We make two key observations here. First, there is an increase in both accuracy and MI attack effectiveness as we go from a single model to ensembles comprising of an increasing number of base models. The trade-off is more noticeable for more accurate models trained for a larger number of epochs. Second, we can adapt the design of ensembles to suitably navigate the trade-off between accuracy and privacy. Starting with a single well-trained model (indicated by the the pink circle) achieving around $70\%$ test accuracy  as a baseline (for non-ensemble case), ensembling can be adopted to: (1) improve accuracy by using an ensemble of highly accurate models but at the cost of worse privacy\footnote{ Note that for complicated tasks, such as image classification, the common practice is to train deep models for a large number of epochs and avoid under-fitted models. That is because memorizing samples from long-tailed subpopulations are shown to be necessary to achieve close-to-optimal generalization error \cite{feldman2020does}.} (purple arrow); and (2) improve privacy by intentionally using an ensemble of \textit{under-fitted} models instead of a single model but at the cost of accuracy (brown arrow). However, these two objectives are not achieved simultaneously in deep ensembles.

To better study this phenomenon, we start with the most widely-used form of ensembling in deep models, that is, deep ensembles, and the most common type of membership inference attack based on confidence outputs. To understand the root cause of this trade-off, we show that using deep ensembles to improve accuracy exacerbates its susceptibility to membership inference attacks by making train and non-train samples more distinguishable. By analyzing the confidence averaging mechanism of deep ensembles, we investigate potential factors that enable membership inference. We show that the most influential factor is the level of correct agreement among models. Simply put, the number of models that correctly classify a train sample is often greater than the ones that correctly classify a test sample. This results in a wider confidence gap between train and non-train samples, when confidence values are averaged, enabling more effective membership inference attacks.

We further show that the difference in the level of correct agreement between train and non-train samples is correlated with models' generalization gap. Hence, a natural question to ask is "can deep ensembles that use less overfitted models mitigate privacy issues while achieving high accuracy?". To answer this question, we study several regularization techniques, common membership inference defenses, and a few other ensembling approaches. We again observe a privacy-accuracy trade-off pattern similar to that shown in Figure~\ref{fig-intro}.

Finally, using the insights obtained in the above analysis, we derive yet effective modification on deep ensembles that not only mitigate the privacy leakage issue in deep ensembles, but also improve privacy significantly. Instead of averaging confidence values, our approach outputs the confidence of the most confident model among the models that predict the same label as the entire ensemble. We show that this approach has several benefits: 1) It mitigates the effectiveness of the membership inference attacks to the point where the attack often performs similar to a random guess. 2) It can still achieve similar accuracy as of deep ensembles (averaging confidence values). 3) It does not require any change to the training process of base models. In other words, this can be easily adopted even for the systems whose base models have already been trained.

\textit{Summary of contributions:} 
In this paper, we perform a systematic empirical study of MI attacks on deep ensemble models. We start with an in-depth analysis of the most common ensembling technique and membership inference attack, and then we extend the results to various ensembling techniques and membership inference attacks. First, we show that when deep ensembles improve accuracy, it also leads to a different distribution shift in the prediction confidence of train and test samples, which in turn enables more effective membership inference. Second, we analyze various factors that potentially cause the prediction confidence of train and non-train samples to diverge. Among potential factors, we show that the most dominant factor is the level of correct agreement among models which indicates that more models in an ensemble agree on their prediction when a sample is a training sample. Hence, the aggregation of their prediction yields higher confidence output in comparison with non-train samples. We show that common defense mechanisms in membership inference literature, including differential privacy, MemGuard, MMD+Mixup, L1 and L2 regularization, as well as other ensemble training approaches, such as bagging, partitioning, 
can be used to mitigate effectiveness of MI attacks but at the cost of accuracy. We solve this trade-off issue by changing the fusing mechanism of deep ensembles which improves the accuracy and privacy, simultaneously. Although the main focus of the paper is on deep ensembles, we also cover bagging, partitioning, 
weighted averaging (Appendix \ref{appendix-all-epochs-stacking}), as well as more advanced and state-of-the-art ensembling techniques, such as snapshot ensembles \cite{huang2017snapshot} and diversified ensemble networks \cite{zhang2020diversified} (Appendix \ref{appendix-all-epochs-advanced-ensemble}). We observe similar trade-off.

\section{Background}
\subsection{Ensemble Learning}


In literature, ensemble learning refers to various approaches that combine multiple models to make a prediction. Models used to construct an ensemble are often called base learners. There are two main factors to construct an ensemble \cite{sagi2018ensemble}: 1) how base learners are trained to ensure diversity, such as random initialization, bagging, partitioning, etc., and 2) how the output of base learners are fused to obtain the final output, including majority voting, confidence averaging, stacking, etc. 

The most common forms of ensemble learning in classical machine learning algorithms are bagging, partitioning, boosting, and stacking. In \textbf{bagging}, several models are trained with different bootstrap samples of the training dataset. In other words, each model is trained on a randomly selected under-sampled version of the entire dataset. As a results, the diversity is ensured by varying training set of each model. The model outputs are often fused using majority voting or averaging. Random forest is a widely-used example of bagging of decision trees. In \textbf{partitioning}, similar to bagging, base models are trained on different subset of the entire dataset and fused with majority voting or averaging. However, unlike bagging, the training datasets are non-overlapping. \textbf{Boosting} is an ensemble learning technique that focuses on samples that were misclassified by previous trained models. In other words, the models are trained sequentially such that the second model aim to correct the prediction of the first model, the third model aims to correct the prediction of the second model, and so on. This is often done by changing the weight of each sample during the training. The most common boosting algorithms are AdaBoost, gradient boosting and XGBoost. \textbf{Stacking} is a meta-learning ensembling approach where a meta-learner is trained on top of the base models. Meta-learner is often a simple regression or a shallow neural network. The complexity is often shifted to base models. There are hundreds of variations of these methods in literature that are less common and the study of them is out of the scope of this paper.

Unlike classical machine learning domain where several popular ensemble methods exist and are equally used for different scenarios, there is only one heavily-used method for deep learning models, called \textbf{deep ensemble} \cite{kondratyuk2020ensembling}. In this method, 1) base models are initialized with random weights and trained on the same training dataset, and 2) their prediction confidence are fused through averaging to construct the final output. Unlike ensemble of traditional machine learning algorithms, in deep ensembles, the main source of diversity often comes only from random initialization of base learners \cite{fort2019deep}. In fact, other sources of diversity, such as bagging, have been shown to considerably degrade the overall accuracy of a deep ensemble \cite{lee2015m,lakshminarayanan2016simple}. Although some classical ensemble learning approaches, including bagging, partitioning, and stacking, can be easily adopted for deep learning models, they are rarely used due to the low accuracy in comparison with deep ensembles. 

Recently, a few promising ensemble methods for deep models have been proposed. In \textbf{snapshot ensemble} \cite{huang2017snapshot}, only one deep model is trained. Here, base models are snapshots of that single model at different epoch during the training. Specifically, every time the model is converged, an snapshot is taken and the training process continues by using a large learning rate to find a new local optimum. This process significantly reduces the training time of ensemble which is an important obstacle for training deep models on large dataset. In \textbf{diversified ensemble network} \cite{zhang2020diversified}, the output of base models are aggregated in a shared layer (similar to stacking) and are trained jointly. The main novelty is that it uses an additional loss term that ensures each model is optimized in different directions of diversity. 
Interestingly, unlike deep ensembles, they show that there is an optimum number of base models over which the diversified ensemble network's accuracy starts to degrade. To the best of our knowledge, there is no ensembling approach in literature for deep models that \textit{considerably} outperforms deep ensembles.

\subsection{Membership Inference}
Membership inference is a form of privacy leakage where the goal is to determine if a sample was used during the training of a target model. Samples used during training are often referred to as \textit{member} or \textit{train samples}, and other samples are referred to as \textit{non-member, non-train}, or \textit{test samples}. Majority of the membership inference attacks are built on the intuition that a trained model outputs greater confidence on member samples than non-member samples. Hence, they often use confidence values as an attack feature. The first MI attack on neural networks, refereed to as \textbf{Shokri}'s attack, was proposed in \cite{shokri2017membership} where the attacker trains an attack classifier to predict the membership status. The attack classifier takes the prediction confidence of the victim model as an input. The intuition is that the confidence values are often greater for train samples in comparison with non-train samples \cite{rezaei2020towards}. Hence, assuming that the attacker has access to a dataset with a similar distribution, she trains a set of shadow models to mimic the victim model. Since the membership status of the data with which the shadow models are trained are known to the attacker, she can use the data to train the attack classifier. Many papers use the same idea with different variations or less restrictive assumptions \cite{salem2018ml, liu2019socinf, song2019privacy, long2017towards, truex2019demystifying, long2018understanding, yeom2018privacy, rezaei2020towards, zou2020privacy, li2020label}. 

These confidence-based approaches suffer from low accuracy because they cannot distinguish hard member samples (for which the confidence is low) from easy non-member samples (for which the confidence is high). The state-of-the-art membership inference attacks aim to solve this issue by calibrating the confidence so that it takes the difficulty of the target sample into account \cite{watson2021importance, sablayrolles2019white, rezaei2022efficient}. The most recent state-of-the-art MI attack with difficulty calibration is proposed in \cite{watson2021importance}. In this \textbf{Watson attack}, the attacker trains multiple shadow models with datasets that exclude the target sample. Thus, the attacker can obtain the average confidence of the target sample when the sample is absent from the training data. During the attack, the attacker obtains the attack score by subtracting the obtained average confidence from the confidence output of the target model. Hence, the large attack score is an indication of a member sample. 
In \cite{rezaei2022efficient}, authors propose the \textit{subpopulation-based MI attack} that instead of calibrating the difficulty by training multiple shadow models, they use samples from the same subpopulation as the target sample to estimate the average model response. The intuition is that if the confidence of the victim model is significantly higher on the target samples in comparison with similar samples, it is an indication of a member sample. This attack achieves similar accuracy as other calibration-based attacks, while decreasing the training computation cost of shadow models significantly.
`

Most previous work built upon the idea of using prediction confidence to infer the membership status, except \cite{rezaei2020towards, choo2020label, rahimian2020sampling}. In \cite{rezaei2020towards}, the authors assumed white-box access to the target model and launched a series of MI attack based on confidence values, distance to the decision boundary, gradient w.r.t model weight, and gradient w.r.t input. Interestingly, none of those approaches significantly outperforms confidence-based attack. In \cite{choo2020label}, the authors proposed two attacks based on input transformation and distance to the boundary in a black-box setting. Similarly, in \cite{rahimian2020sampling}, they propose the \textbf{sampling attack} that randomly perturbs an input to obtain a set of random transformations of the input and uses the predicted labels to infer membership status. The intuition is that deep models are more robust on training samples. As a result, perturbed training samples are less likely to be labeled as a class different from the unperturbed training sample. The attack simply counts the number of perturbed samples that has different label than the unperturbed sample as an MI attack feature. In \cite{rahimian2020sampling} shows that DP-SGD can effectively defend against the sampling attack, at the cost of accuracy.

In this paper, we illustrate the results of Shokri's shadow model attack \cite{shokri2017membership}, Watson's MI attack with sample difficulty calibration \cite{watson2021importance}, and sampling attack \cite{rahimian2020sampling}. As shown in \cite{li2021membership}, most confidence-based and loss-based membership inference attacks that do not use sample difficulty calibration achieve similar performance. Thus, we only show the results of Shokri's original attack in this paper. Furthermore, the other variations of the membership inference attacks with sample difficulty \cite{rezaei2022efficient, sablayrolles2019white} achieve similar accuracy in comparison with \cite{watson2021importance}. Decision boundary-based attacks are extremely computational and query inefficient.
Gradient-based approach of \cite{rezaei2020towards} also needs full knowledge of the entire deep ensemble. We also show the accuracy of sampling attack \cite{rahimian2020sampling} because it only needs black-box access to the defender's model.

\subsection{Membership Inference Defenses}
Defense mechanisms against membership inference attacks can be summarized into two categories \cite{rahimian2020sampling}:

\textit{Generalization-based:}
Shokri \cite{shokri2017membership} was the first to correlate membership inference success with overfitting. Since then, many standard regularization techniques have been used to alleviate overfitting, such as L1 regularization \cite{choo2020label}, L2 regularization \cite{choo2020label, truex2019demystifying, nasr2019comprehensive, jia2019memguard, shokri2017membership}, differential privacy \cite{choo2020label, rahimian2020sampling}, dropout \cite{jia2019memguard}, and adversarial training \cite{nasr2018machine}. Interestingly, ensemble learning has also been proposed as a defense mechanism. In \cite{salem2018ml}, they proposed a combination of partitioning and stacking as a defense mechanism. The intuition is that training each model with different subset of data makes the entire ensemble model less prone to overfitting. Note that these defense mechanisms often degrade the accuracy of the target model (see Section \ref{sec-defenses}) \cite{choo2020label}.

\textit{Confidence-masking:} These defense mechanisms aim to reduce the amount of information that can be obtained from the output of a target model by perturbing \cite{jia2019memguard} or limiting the dimensionality of the output \cite{shokri2017membership, truex2019demystifying, choo2020label}. Most confidence-masking approaches manipulate confidence values post-training. As a result, the output values of these models do not reliably represent the "confidence" of the model. These approaches are built under the assumption that accurate prediction of confidence is not needed. However, many applications require accurate estimation of confidence. Moreover, if the accurate prediction of confidence is not required, then the trivial MI defense would be to only output the class label and avoid these confidence-masking defenses altogether. In this paper, we cover MemGuard-random defense as it has already shown to outperforms the other confidence-masking mechanisms \cite{li2021membership}.


\section{Threat Model}
Our threat model works under the scenario of machine-learning-as-a-service (MLaaS) where an ML prediction API is provided by an MLaaS provider. 
The API is accessible to users who can query the API with an input and obtain the prediction output. 
In this scenario, a malicious user, referred to as an \textit{attacker}, can query the API to obtain unintended information beyond the prediction output.
Specifically, the attacker aims to launch a membership inference attack to identify training samples used to train the MLaaS API's model. 
In this paper, we refer to the MLaaS provider as the \textit{defender} or \textit{victim}.

\subsection{Defender}
\textbf{Assumptions:} Here, we assume that the defender uses deep ensembles to improve the accuracy of the prediction model. 
The defender uses the training dataset, $D_{tr}$, to train multiple base models. 
The defender may or may not use all available training samples to train each model. 
As long as a training sample is used to train at least one base model, we label the sample as a member. 
Moreover, the defender provides an API access and returns prediction confidence values. 
In a multi-class classification task, the output is a vector of probabilities corresponding to each class, referred to as confidence values. 
The defender can train base models from scratch and, as a result, apply regularization techniques or membership inference defense mechanisms that requires modification of the training process.
The defender can also use any fusing technique rather than confidence averaging which is used in deep ensembles.


\textbf{Objectives:} The main objective of the defender is to mitigate the membership inference attack while benefiting from the ensemble learning's improved accuracy. Preferably, the solution should impose minimal computational cost at both training and inference time. 
We study several MI defense mechanisms, including MMD+Mixup \cite{li2021membership}, L1 and L2 regularization, and DP-SGD. 
Moreover, we investigate several ensembling approaches that suggested in literature as a defense mechanism, including bagging, and partitioning. 
Finally, we propose a simple solution that achieve both objectives of improving accuracy and privacy, simultaneously.

\subsection{Attacker}
\textbf{Objectives:} The attacker aims to launch a membership inference attack to identify samples of the defender's training dataset, $D_{tr}$. 

\textbf{Knowledge:} We make the following assumptions about the attacker's knowledge:

1) The attacker has the full knowledge of the classification task. Given that the purpose of the API is to provide a service to users, it is reasonable to assume that the task for which the API is developed is known to all users, including the attacker. This includes the number of classes, class labels, and the input shape.

2) We assume that the attacker has only black-box access to the defender's model. Although it is possible in some scenarios to approximate model parameters through model extraction methods \cite{tramer2016stealing}, it is of the model owner's interest to keep the model proprietary as otherwise it defies the whole point of MLaaS as a business model.

3) The attacker may know the model architecture, training algorithms, the type of ensemble method, and other training information. 
These information may sometimes be available via the API's documentation. 
Specifically, when the attacker needs to train shadow models, he uses the same architecture and training parameters as the defender's model.
But, as it is shown in \cite{salem2018ml}, the lack of this information barely changes the MI attack effectiveness, at least when original shadow-based MI attack of Shokri \cite{shokri2017membership} is used.

\textbf{Capabilities:} The attacker has the following capabilities:

1) The attacker can collect a dataset, $D_{s}$, that has the similar distribution as the $D_{tr}$ to train shadow models on the same task if needed. 

2) The attacker has computational resources to train an attack model, which takes some features from a target sample and outputs the membership status. In the simplest form, it can be a threshold on the output of the defender's confidence output \cite{yeom2018privacy}, or an ML model \cite{shokri2017membership}. 



\section{How Does Ensembling Increase Membership Inference Effectiveness?}
\label{sec-how-ensemble}

In this section, we thoroughly investigate the most widely-used deep models ensembling technique and membership inference attack, that is, deep ensembles and confidence-based attack. We mainly focus on distributions of confidence values when a deep ensemble is used and how it can lead to more distinguishable distributions. How an actual MI attack can use this feature is studied in Section \ref{sec-eval}. Furthermore, other forms of ensembling techniques and membership inference attacks are shown in Section \ref{sec-eval}.

\subsection{Confidence Distribution Shift}
\label{section-dist-change}

\begin{figure*}
\def\tabularxcolumn#1{m{#1}}
\begin{tabularx}{\linewidth}{@{}cXX@{}}
\begin{tabular}{ccc}
\subfloat[All samples (Non-EL)]
{\includegraphics[width=0.30\linewidth]{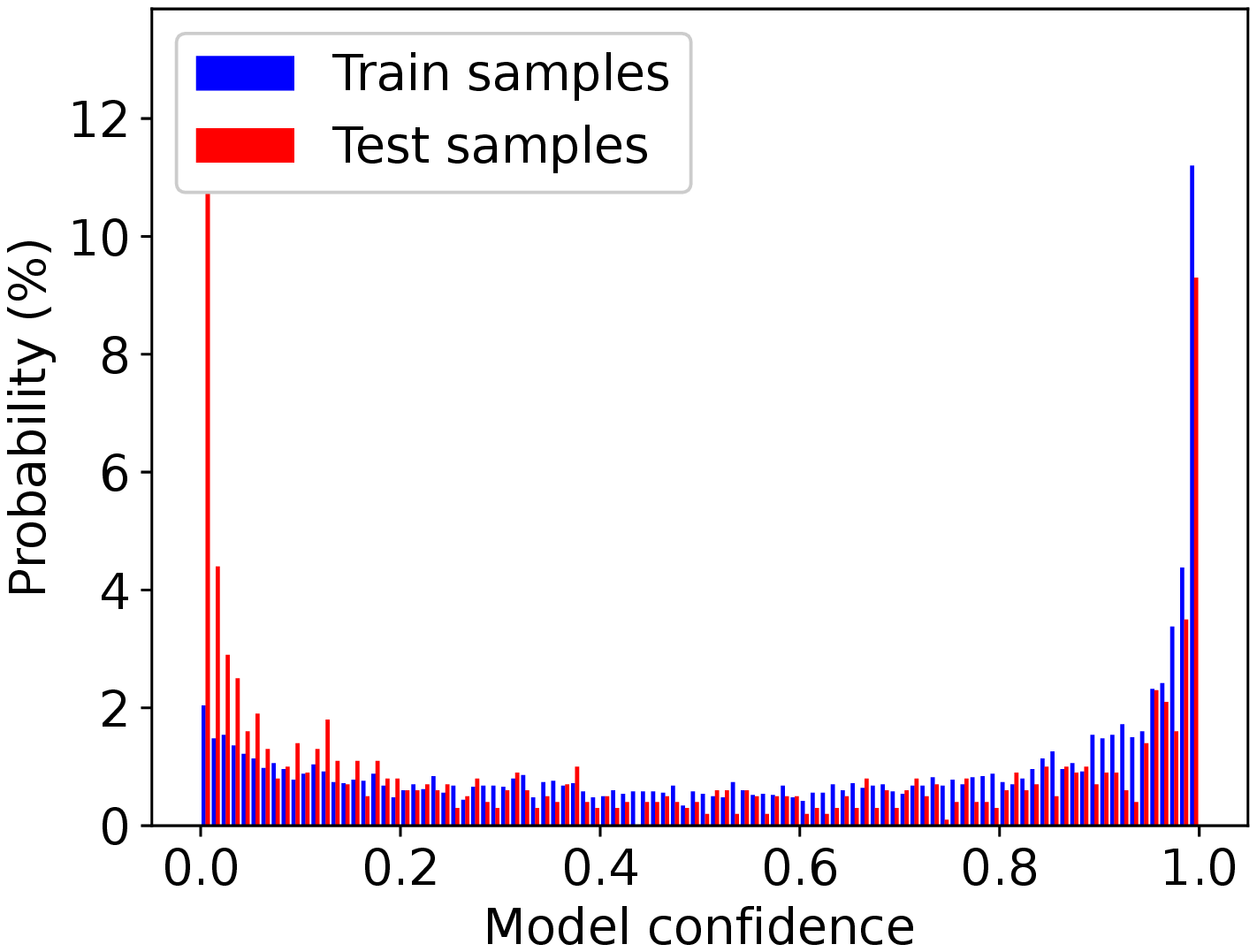}} 
   & 
\subfloat[Correctly classified samples (Non-EL)]
{\includegraphics[width=0.30\linewidth]{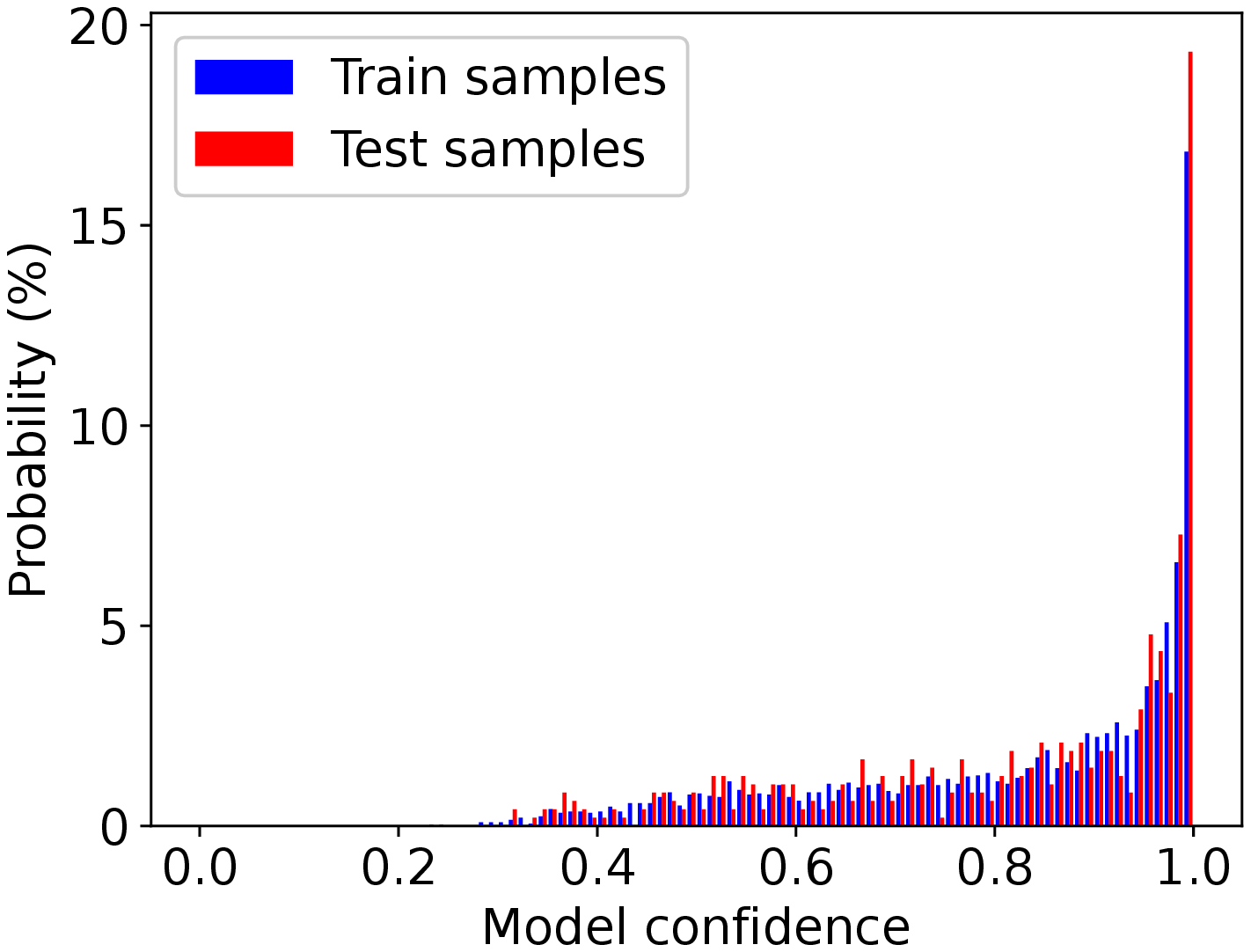}}
   &
\subfloat[Incorrectly classified samples (Non-EL)]
{\includegraphics[width=0.30\linewidth]{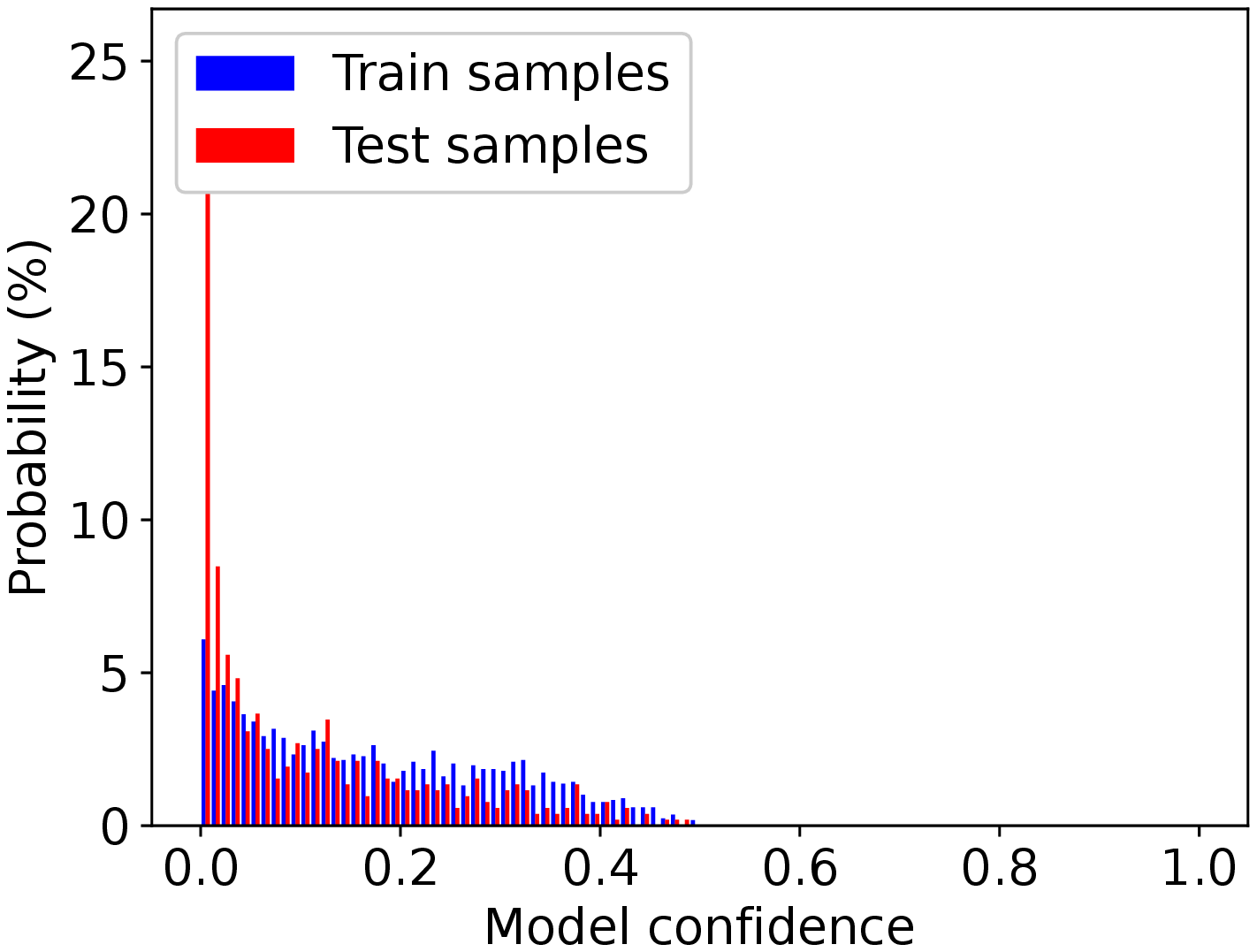}}\\

\subfloat[All samples (EL-10)]
{\includegraphics[width=0.30\linewidth]{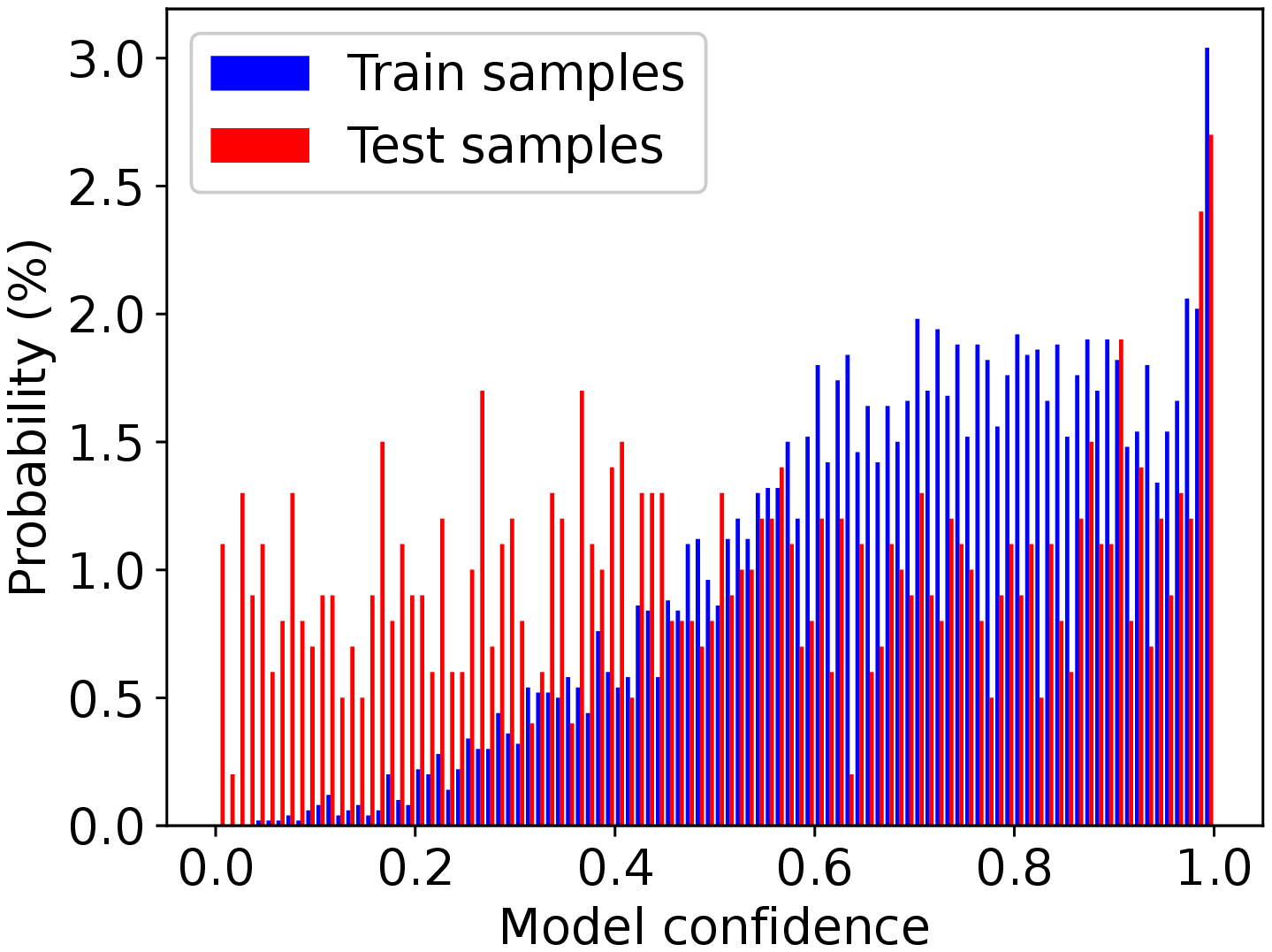}} 
   & 
\subfloat[Correctly classified samples (EL-10)]
{\includegraphics[width=0.30\linewidth]{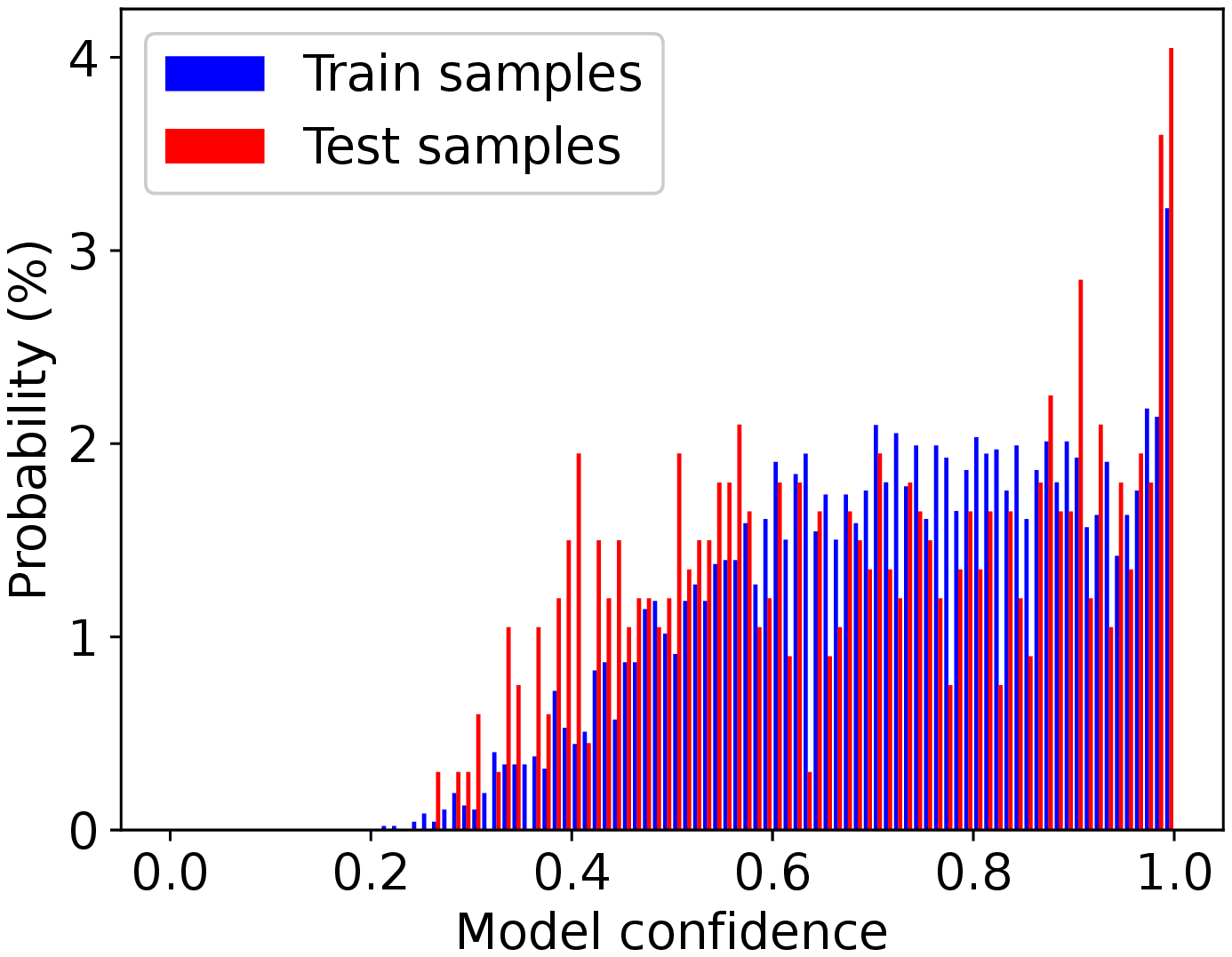}}
   & 
\subfloat[Incorrectly classified samples (EL-10)]
{\includegraphics[width=0.30\linewidth]{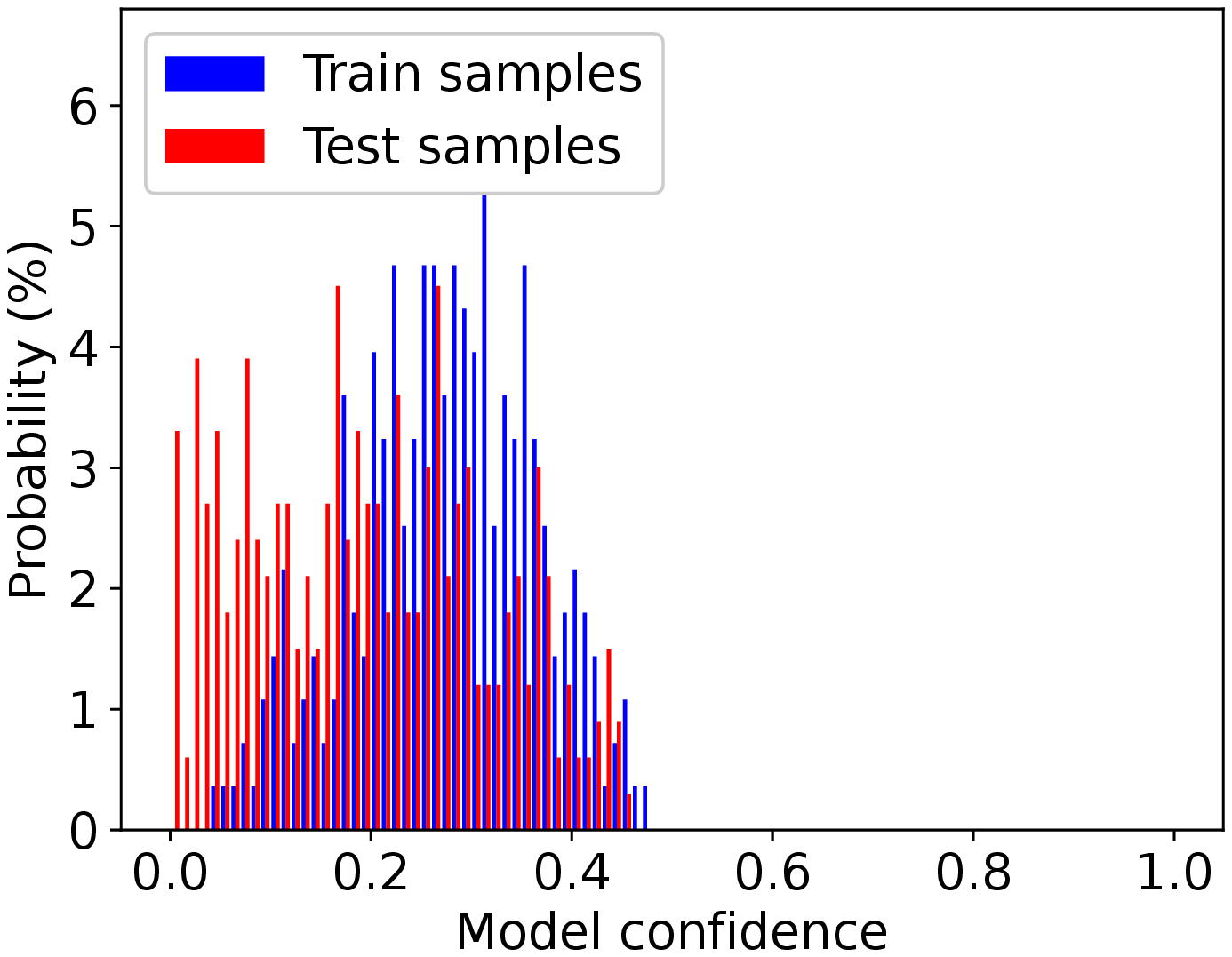}}\\

\subfloat[All samples (EL-10)]
{\includegraphics[width=0.30\linewidth]{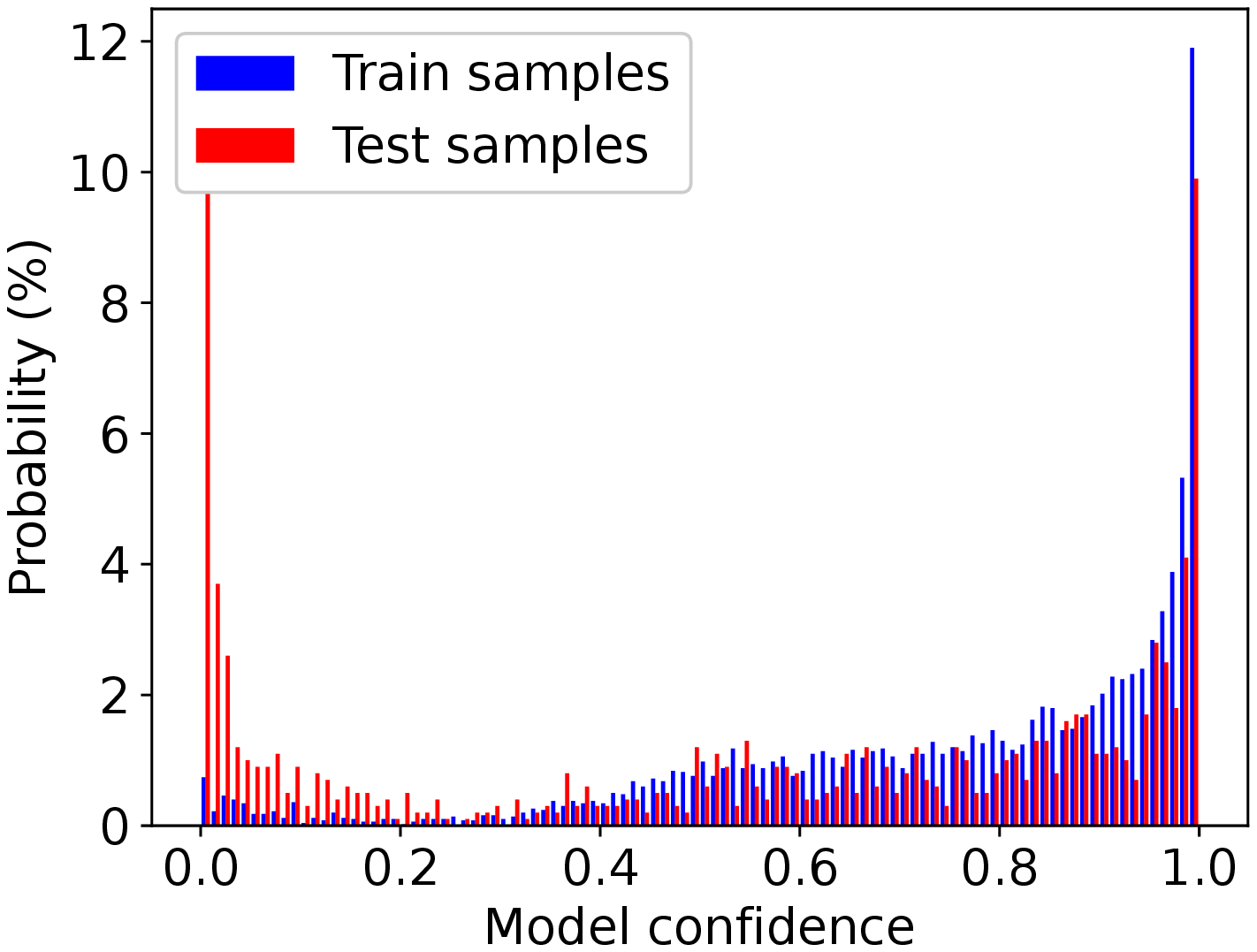}} 
   & 
\subfloat[Correctly classified samples (EL-10)]
{\includegraphics[width=0.30\linewidth]{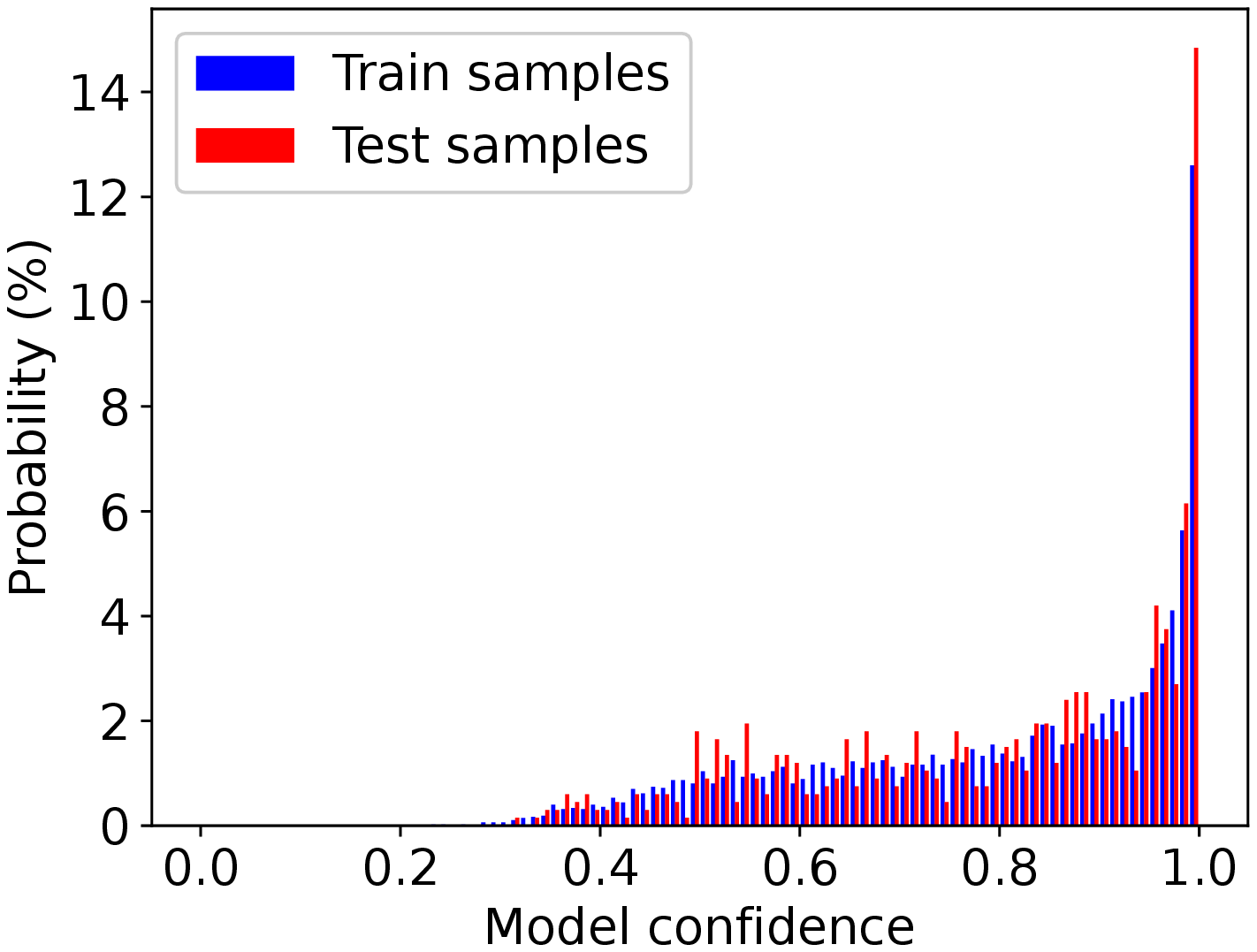}}
   & 
\subfloat[Incorrectly classified samples (EL-10)]
{\includegraphics[width=0.30\linewidth]{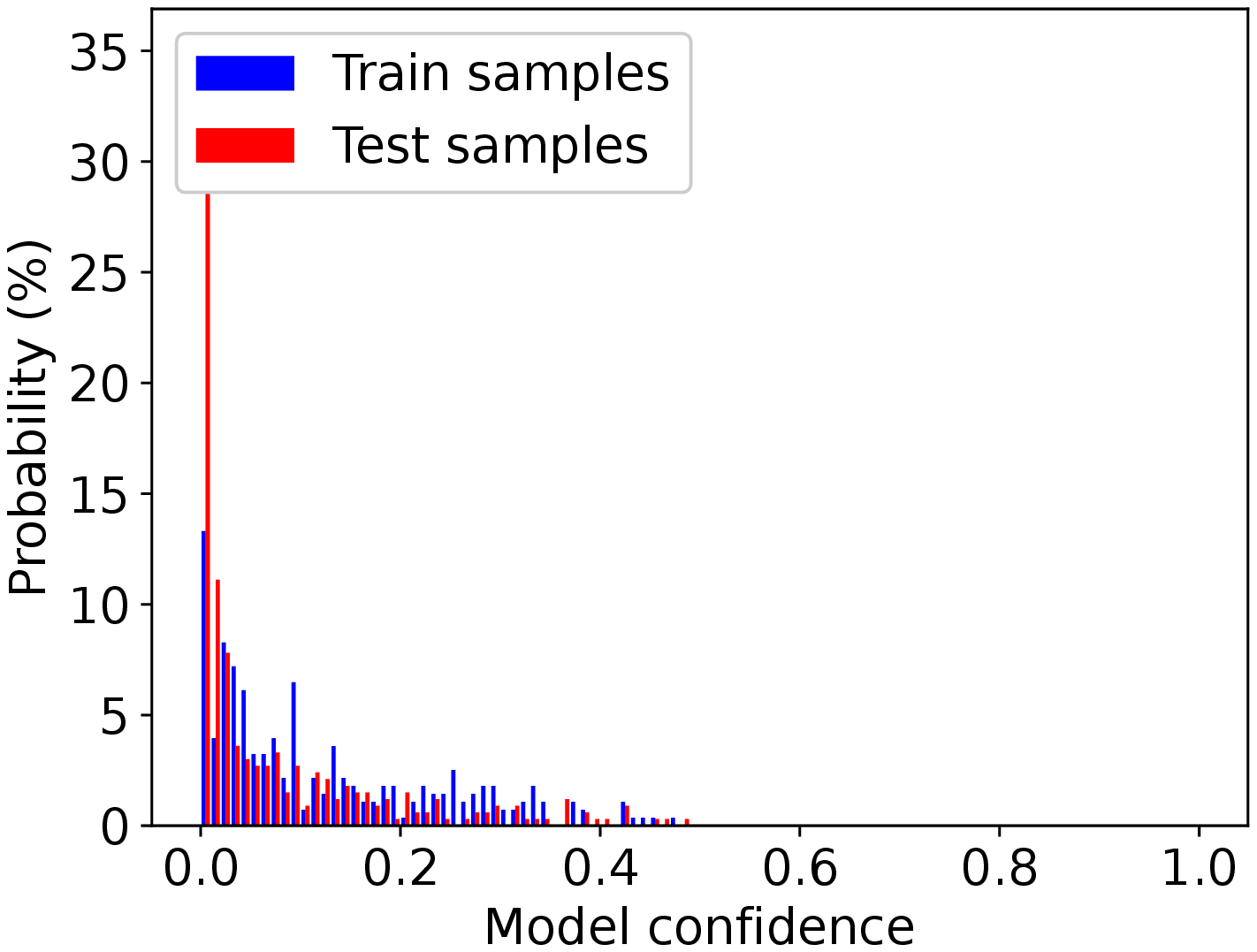}}\\

\subfloat[All samples (EL-10)]
{\includegraphics[width=0.30\linewidth]{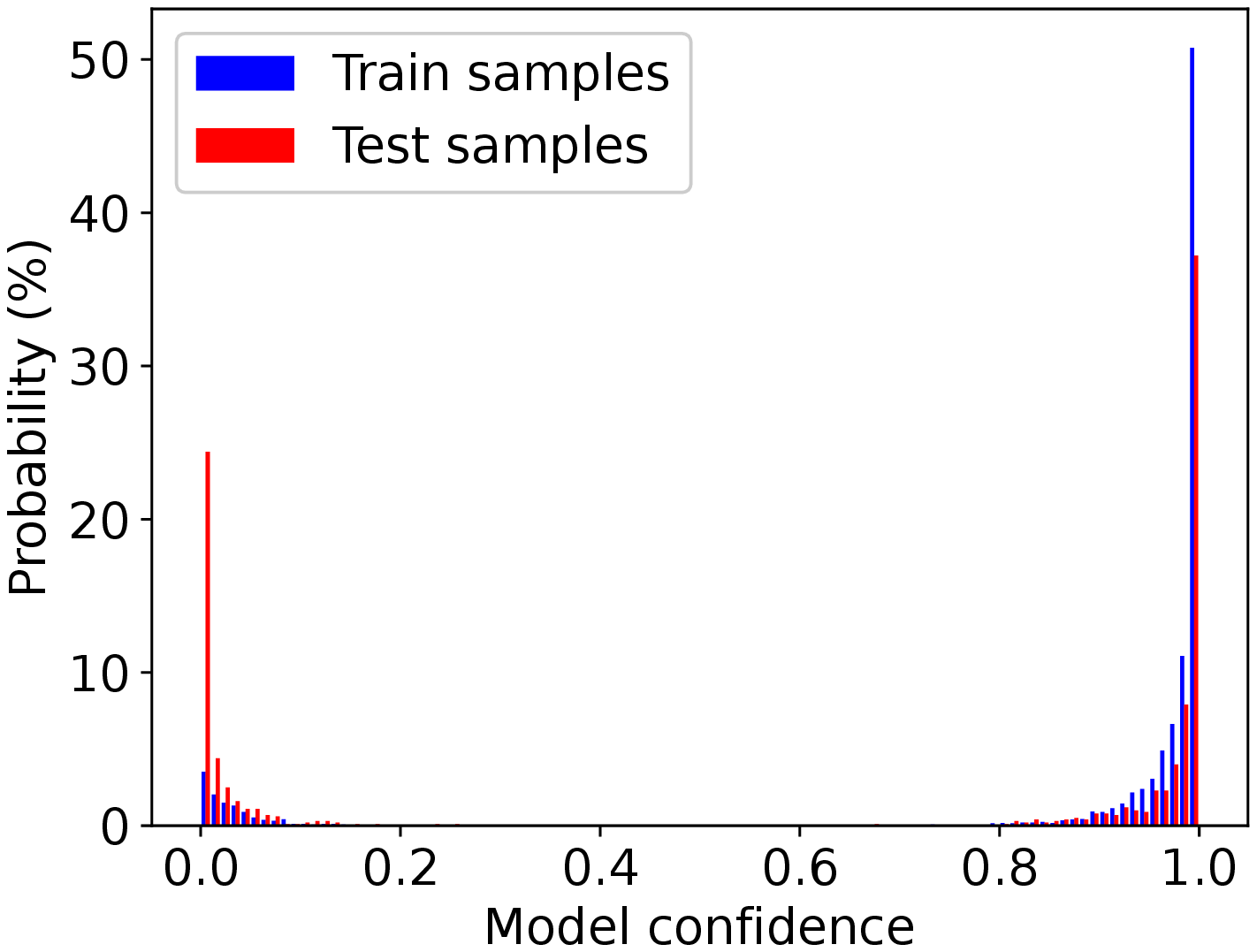}} 
   & 
\subfloat[Correctly classified samples (EL-10)]
{\includegraphics[width=0.30\linewidth]{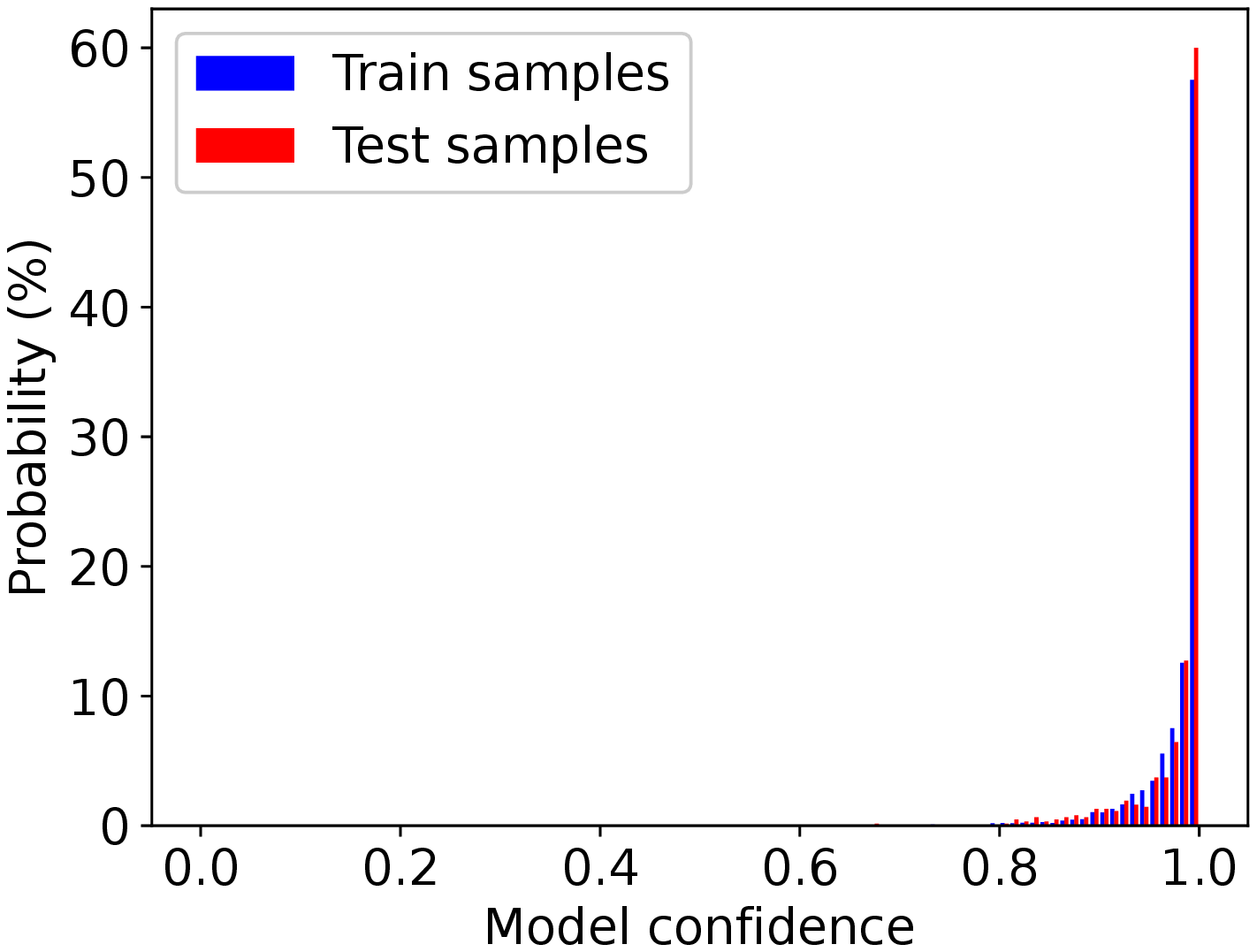}}
   & 
\subfloat[Incorrectly classified samples (EL-10)]
{\includegraphics[width=0.30\linewidth]{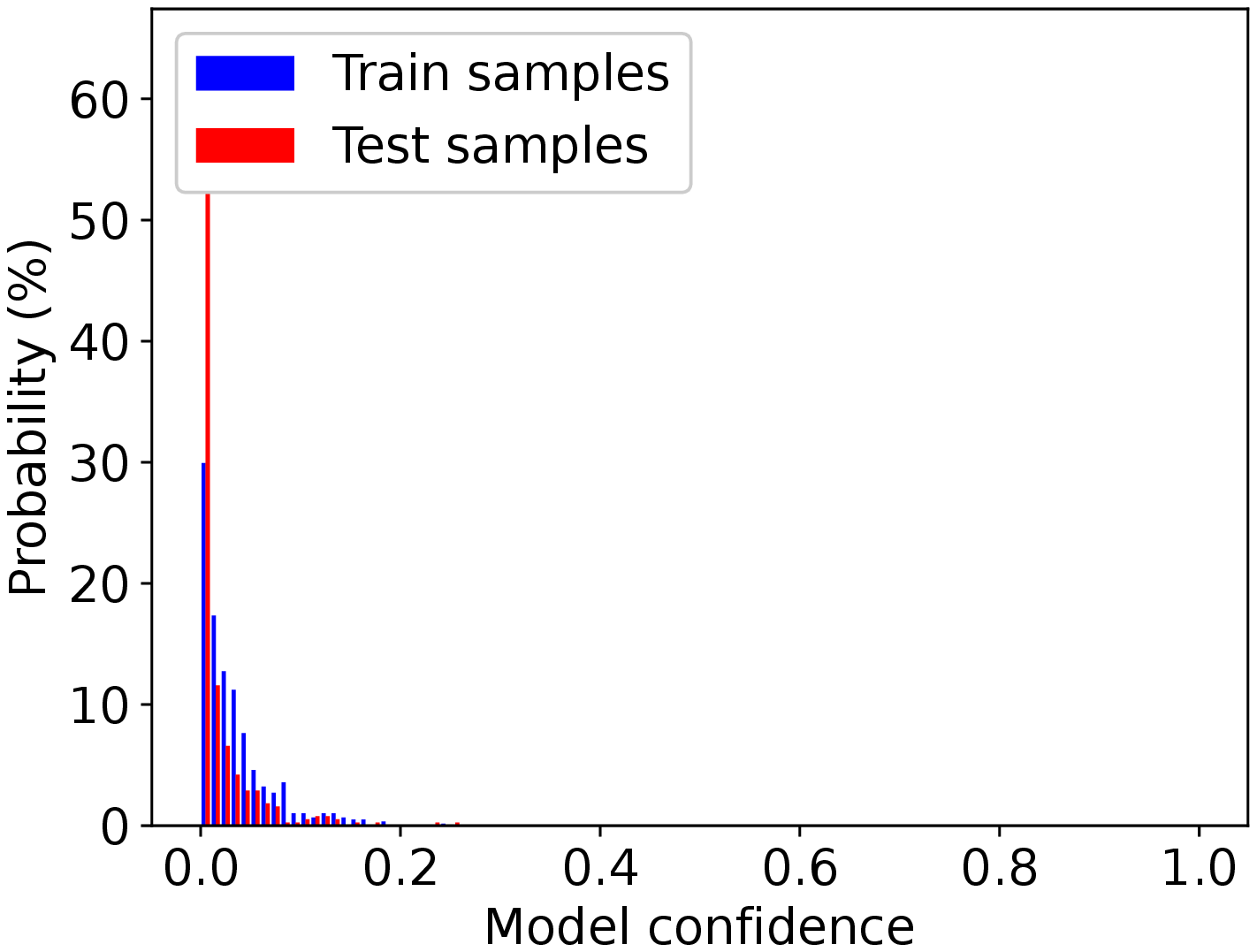}}\\
\end{tabular}
\end{tabularx}
\caption{ResNet20 on CIFAR10 (Class \#3): Confidence (i.e. maximum of the softmax output) distribution of a single model (top row),  an ensemble of 10 models using \textit{ensemble averaging} (the second row), an ensemble of 10 models using \textit{first agreed confidence} approach (the third row), and an ensemble of 10 models using \textit{maximum agreed confidence} approach (the fourth row). Jensen–Shannon divergence of the two distributions are as follows: a) .2276, b) .1484, c) .2515, d) .3037, e) .1682, f) .3408, g) .2940, h) .1432, i) .3076, j) .2595, k) .0846, and l) .2753.}
\label{fig-dist}
\end{figure*}

Ensemble learning is only helpful when base learners disagree on some samples \cite{kuncheva2003measures, sagi2018ensemble}. Otherwise, ensembling does not improve accuracy. Furthermore, when deep ensemble is used, the confidence values of multiple models are averaged to obtain the final prediction. Consequently, when ensembling improves accuracy, it averages the prediction confidence of highly confident predictions (mostly from models which correctly classified the sample) and less confident predictions (mostly from models which misclassified the sample). As a result, confidence distribution shift is inevitable for both train and test samples. This phenomenon can be observed as the distribution of Figure~\ref{fig-dist}(a) shifts to that of \ref{fig-dist}(d). This can be better observed by separating correctly classified samples which have significantly higher prediction confidence (Figure~\ref{fig-dist}(e)) and misclassified samples which have lower prediction confidence (Figure~\ref{fig-dist}(f)). One can clearly observe that both distributions shift more towards the center from Figure~\ref{fig-dist}(b) and (c) to Figure~\ref{fig-dist}(e) and (f).
However, the change in the distribution of train and test samples does not necessarily cause a more effective membership inference if the change has a similar effect on the confidence distribution of both train and test samples. In the next subsection, we analyze the potential factors that affect the distribution change and how they can change confidence distribution of member and non-member sets differently.

\subsection{Effect of Ensembling on Individual Samples}
\label{section-factors}

We use $y_i$ to denote the confidence value of the $i^{th}$ model  in an ensemble of $n$ models. Hence, for a given sample $x$, the output of the ensemble is:
\begin{equation}
\label{eq-ensemble-averaging}
    y_{el}(x) = \frac{ \sum^{n} y_i(x)} {n} = \frac{ \sum^{c} y^{c}_i(x) + \sum^{m} y^{m}_i(x) } {n}.
\end{equation}
Given a single sample, we can further divide models in the ensemble into two groups: 1) models that correctly classified the sample denoted by $y^{c}_i$ and 2) models that misclassified the sample by $y^{m}_i$. For a given sample $x$, $c$ models correctly classify it and $m$ models misclassify, where $c+m=n$. Note that the value of $c$ and $m$ depends on the sample\footnote{ By an abuse of notation, we use $c$ ($m$) to refer to (in)correctly classifying models and also as a superscript for the model output of (in)correctly classifying models, that is, $y^{c}_i$ ($y^{m}_i$).}.

\begin{figure*}
\centering
\centering
\begin{tabular}{ccc}
\subfloat[Correct agreement distribution]{\includegraphics[width=0.3\linewidth]{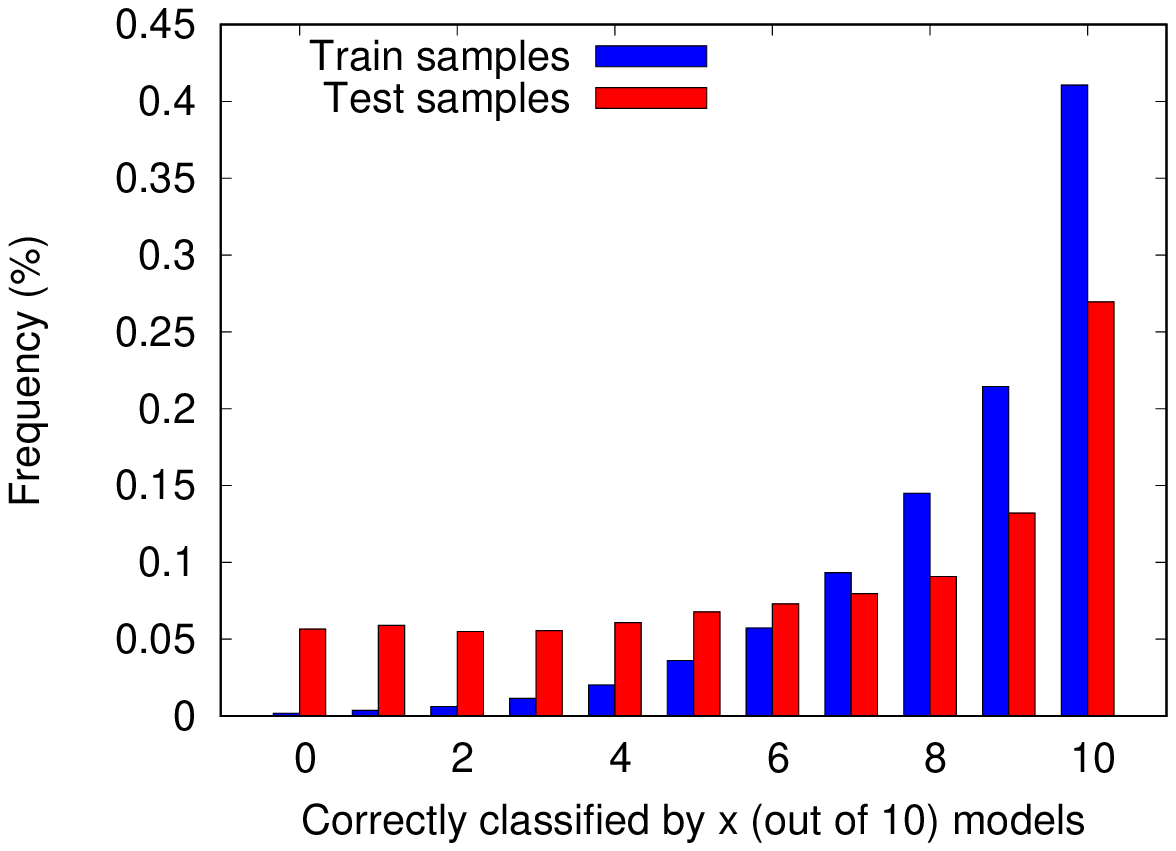}} 
& \subfloat[Distribution difference at each agreement level]{\includegraphics[width=0.3\linewidth]{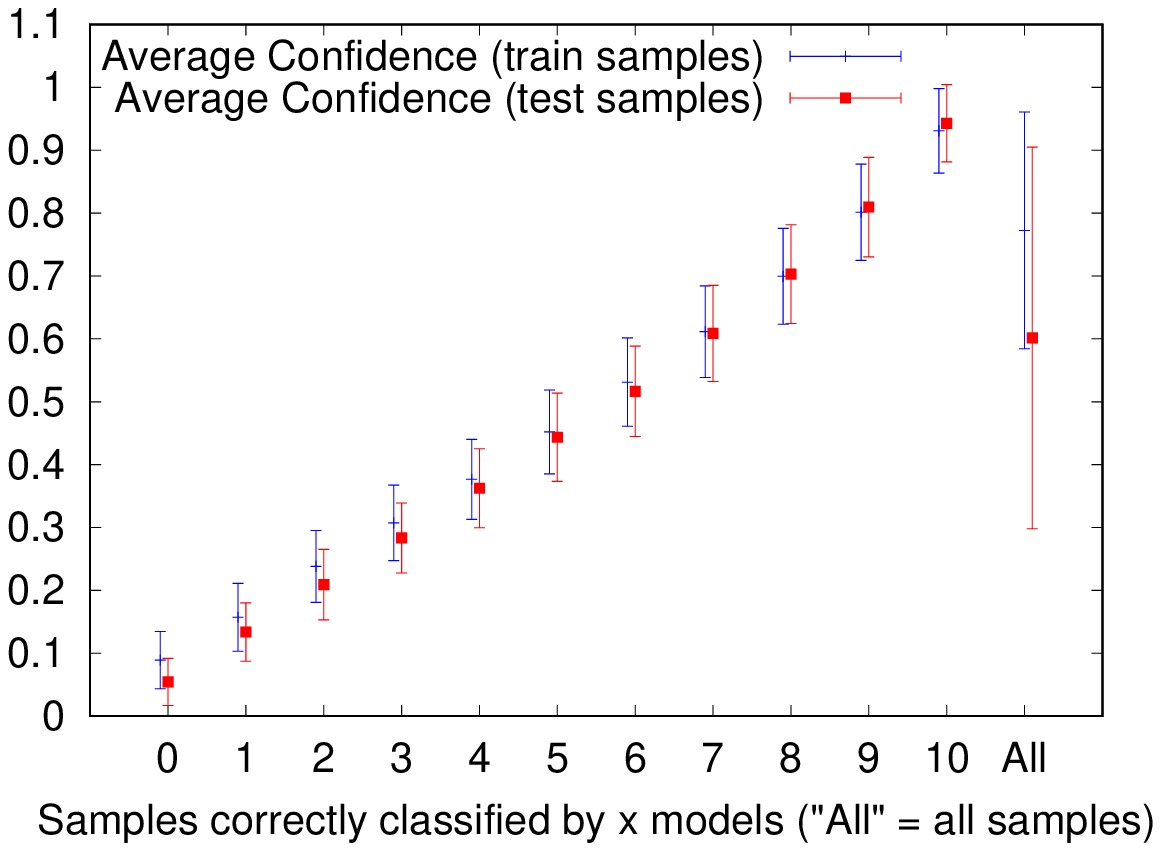}}
& \subfloat[The effect of three factors on MI attack]{\includegraphics[width=0.3\linewidth]{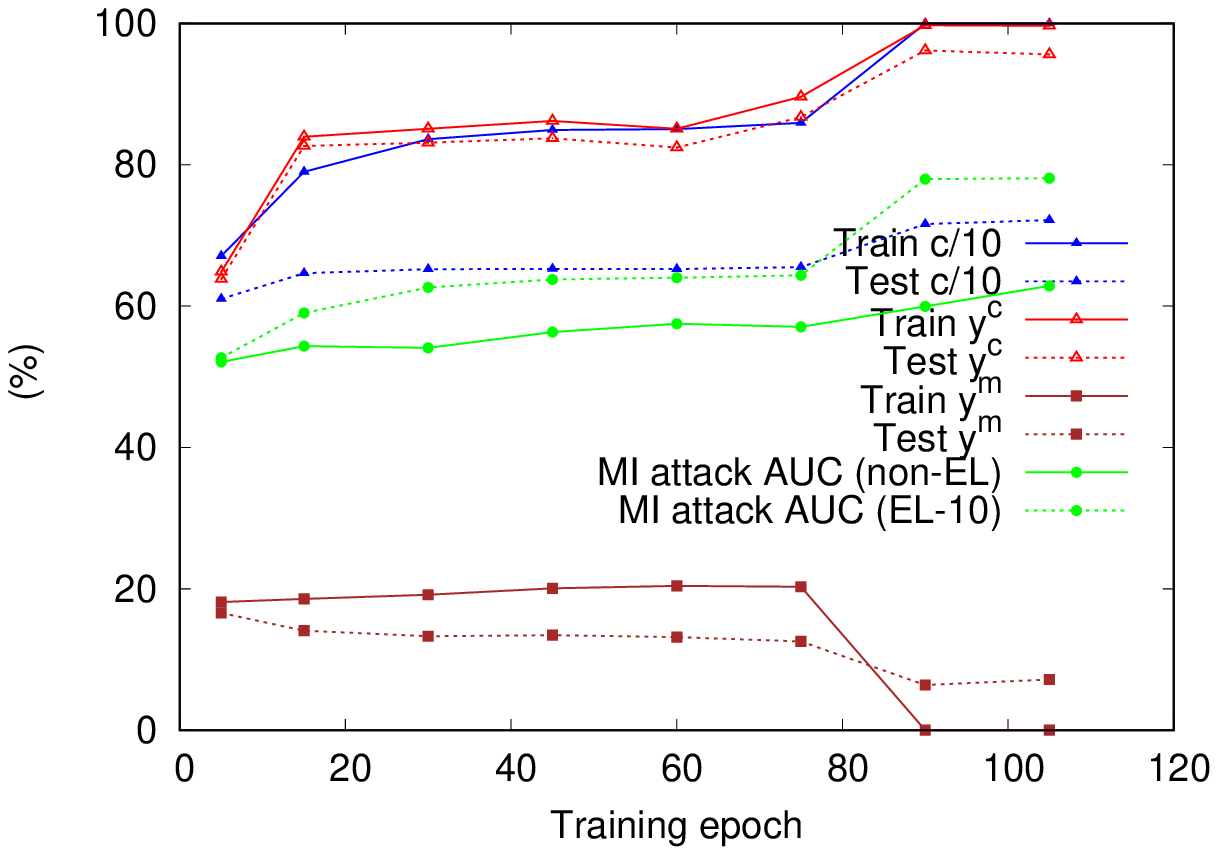}}\\
\end{tabular}
\caption{AlexNet model trained on CIFAR10. \textbf{Left:} The distribution of the number of times a sample is correctly classified by 10 models used in the ensemble. The models often make less classification mistake on train samples than test samples. \textbf{Center:} By separating samples based on how many times they have been correctly classified, we can observe that the confidence output of these samples are negligible between train and test sets. Only when all samples are compared the distribution difference is significant and that is the direct effect of the third factor, namely the level of correct agreement. \textbf{Right:} The effect of the three factors on the MI attack. The values of $y^c$ and $y^m$ are confidence values in percentage. $c/10$ is the percentage of models that correctly classify a sample. As the gap between the level of correct agreement of train and test widens (the blue lines), the MI attack on ensembles becomes more effective than a single model (green lines). }
\label{fig-factors}
\end{figure*}

Based on the Eq. (\ref{eq-ensemble-averaging}), three major factors affect the final confidence value ($y_{el}$) of a sample: $y^c$, $y^{m}$, and $c$. As a result, if these values are different for train and test samples, the ensembling causes different shift in the distributions, and consequently, membership inference attack will be more effective. These factors are as follows:
\begin{enumerate}
    \item Confidence value of correctly classifying models ($y^c$): Since the majority of samples are supposed to be correctly classified by a practical model, any distinguishable confidence difference between train and test samples can lead to a very effective membership inference attack. However, as shown in Figure~\ref{fig-dist}(b), we can observe that there is no significant difference between train and test samples.
    
    \item Confidence value of misclassifying models ($y^{m}$): Unlike correctly classified samples, there is a marginal difference between confidence distribution of train and test samples of misclassified samples (see Figure~\ref{fig-dist}(c)). This may be exploited by membership inference attack to partially distinguish between train and test samples.
    
    \item Level of correct agreement ($c$) among models: As shown in Figure~\ref{fig-factors}(a), the number of models that correctly classify a sample ($c$) is greater for train samples than test samples. Since prediction confidence of correctly classified samples are higher than misclassified samples on average, i.e., $y^c>y^m$, and $c$ is smaller for test samples, the ensemble confidence of test samples ($y_{el}$) becomes lower than train samples. As a result, this factor can largely contribute to the effectiveness of membership inference attacks on deep ensembles. 

\end{enumerate}

We note that the first two factors are not unique to ensembles and can be exploited by an attacker in a single model scenario as well. As a result, these two factors have been extensively studied in \cite{rezaei2020towards} in a single model scenario across various image datasets and well-trained models. They have shown that for deep models the first factor ($y^c$) is almost indistinguishable between train and test set and only the second factor ($y^m$) is marginally distinguishable. However, this marginal difference does not have a considerable impact on the different distribution shift in train and test sets. 

On the other hand, \textbf{the level of agreement} has a big impact on different distribution shifts of train and non-train samples. To better demonstrate this, we can analyze the distribution difference in each level of agreement separately. As shown in Figure~\ref{fig-factors}(b), the average confidence of train and test samples are very close and indistinguishable when each level of agreement is drawn separately. If the effect of the first two factors were considerable, the two confidence values for each level of agreement would have been more distinguishable. Note that the average confidence between train and test sets is more distinguishable for the first two points in x-axis (where the majority of models misclassify a sample). However, these distributions only constitute a tiny portion of the training dataset, as shown with the first two blue bins in Figure~\ref{fig-factors}(a). However, when all samples are combined, we can vividly observe that the average confidence gap between train and test sets considerably widens, as shown in the last point in x-axis in Figure~\ref{fig-factors}(b). This clearly demonstrates that the major factor in different distribution shift between train and test sets is the level of agreement ($c$).

Note that, unlike the first two factors, the third factor ($c$) only improves the effectiveness of membership inference attacks in ensemble scenarios because it does not exist in a single model. In other words, if a defence strategy eliminates the effect of the average level of correct agreement (i.e., it ensures that $c$ is close between train and test samples), the membership inference attack is still possible on the ensemble, but only to the degree that it is possible on a single model\footnote{Although this can be understood by analyzing the Eq. (\ref{eq-ensemble-averaging}), it is difficult to demonstrate empirically. The reason is that these three factors are not independent, and, hence, our attempts to significantly change the third factor without changing the other two factors have been unsuccessful.}. As shown in Figure~\ref{fig-factors}(c), as the gap between $y^c$ of train and test sets (red lines) and the gap between $y^m$ of train and test sets (brown lines) increases, the attack on both single model (non-ensemble) and also the ensemble (EL-10) increases. However, only when the average level of correct agreement gap between train and test (blue lines) widens, the membership inference attack on ensembles becomes more effective than on non-ensembles.

Another important observation from Figure~\ref{fig-factors}(c) is that the minimum level of agreement gap between train and test occurs when models are relatively underfitted (i.e., the blue lines in first few epochs). This phenomena has also been partially observed in \cite{fort2019deep} (Figure 2(c)). The main reason is that underfitted models often only learn the most common and generalizable features and, thus, they often agree on the features and predictions. As they move from underfitted region to overfitted region, their generalization gaps widen (blue lines in Figure~\ref{fig-factors}(c)). As a result, they tend to correctly classify train samples more often than test samples. Consequently, they agree on train samples more than test samples and, hence, average gap of correct agreement between train and test set widens. Hence, the wider generalization gap of base learners is, the more effective membership inference attack would be on deep ensembles.

\subsection{Fusion Approaches to Avoid Diverging Distribution Shifts}

As shown in the Section \ref{section-factors}, the main factor for the large diverging distribution shifts of train and test samples in deep ensembles is the level of agreement. This is an inherent consequence of averaging the confidence of multiple base models. There are several ways to avoid outputting the average of base models. Confidence masking approaches can achieve this goal by manipulating the confidence values. The simplest form is to only output the class label (or the top k classes) \cite{shokri2017membership}, or to add a random noise to the confidence value \cite{jia2019memguard}. The drawback of these approaches is that they make the confidence values unreliable which is critical for applications where confidence estimation is necessary.

To address this issue, we propose three methods to mitigate diverging distribution shift and output a valid confidence value. One simple approach is to output the confidence values of a single model among base models. However, if the model is chosen randomly, this approach does not provide the accuracy improvement of ensembling. To avoid this problem, we first use ensemble averaging over confidence outputs of base models, similar to deep ensembles, to obtain the \textit{ensemble predicted label}. Then we output the confidence values of the first model that predicts the label as the ensemble predicted label. We call this approach \textit{first agreed confidence}. Because the predicted label of the ensemble is essentially the same as deep ensembles, the accuracy of the first agreed label is exactly the same as deep ensembles. As shown in the third row of Figure~\ref{fig-dist}, the distribution of the confidence output is similar to a single model (the first row of Figure~\ref{fig-dist}), as expected. Therefore, this approach can easily mitigate the privacy cost of deep ensembles and achieve the same accuracy.

Interestingly, there are simple approaches that not only output a valid confidence values and achieve similar accuracy, but also significantly improve the privacy. Instead of outputting the confidence value of the first model that predicts the ensemble predicted label, we output the confidence values of the most confident model among base models that predicts the ensemble predicted label. We call this approach \textit{maximum agreed confidence}. Similarly, this approach also achieves the same accuracy as deep ensembles because the prediction label is the same. Maximum agreed confidence approach has multiple advantages: 1) it omits the effect of the level of agreement on the output, 2) it outputs a confidence value that reflects one of the base models and hence it is still reliable for the purpose confidence estimation, and 3) it shifts the confidence values of both member and non-member to the extreme ends (either 0 or 1) and, hence, the distributions of member and non-members become even less distinguishable. The last advantage is clearly shown in the last row of Figure~\ref{fig-dist}. In Section \ref{sec-defenses}, we demonstrate that this approach improves both accuracy and privacy at the same time.

Another similar approach is to output the confidence of the most confident model among all models, instead of the model that predicts the ensemble predicted label. We call this approach \textit{maximum confidence}. The accuracy of maximum confidence might be slightly lower than deep ensembles because the most confident model may occasionally be the one that misclassifies the input sample although the majority of base models do not. However, as we show in Section \ref{sec-defenses}, this approach mitigates membership inference attacks slightly better than maximum agreed confidence because it outputs high confidence for some incorrectly classified samples, which mostly belong to the nonmember samples \cite{rezaei2020towards}. 


Each of the three approaches is advantageous in different scenarios. The first agreed confidence approach outputs a confidence of a single model and, as shown in Figure \ref{fig-dist}, it is similar to a single model scenario. Hence, it is beneficial for scenarios where the confidence estimation is needed to be similar to a single model. The maximum confidence and maximum agreed confidence approaches change the overall confidence distribution. Although the confidence values still come from the output of one of the base models, it is not known if it is as useful as a single model scenario when confidence estimation is concerned. This requires further investigation. Regardless of confidence estimation, the maximum agreed confidence achieves the highest accuracy, as of deep ensembles, while improving privacy. The maximum confidence, however, achieves highest privacy while improving accuracy. Therefore, depending on whether the objective is to maximize accuracy or to maximize privacy, one can use maximum agreed confidence or maximum confidence.

\subsection{Why Does it Outperform Gap Attack Significantly?}
\label{sec-why-gap}

Recently, several studies report a simple baseline attack called \textit{gap attack} \cite{choo2020label}, also known as \textit{naive attack} \cite{rezaei2020towards, leino2020stolen} that achieves similar performance as the confidence-based attacks in most scenarios. The gap attack predicts a sample as member if it is correctly classified by the target model, and as non-member otherwise. In other words, gap attack essentially reflects the generalization gap \cite{rezaei2020towards}. In \cite{rezaei2020towards}, authors extensively analyzed this phenomena in deep models and argued current MI attacks that barely outperform gap attacks are ineffective in practice because they only reflect the generalization gap and cannot infer the membership status of each individual sample accurately. 

\begin{figure*}
\centering
\includegraphics[width = 0.9\linewidth]{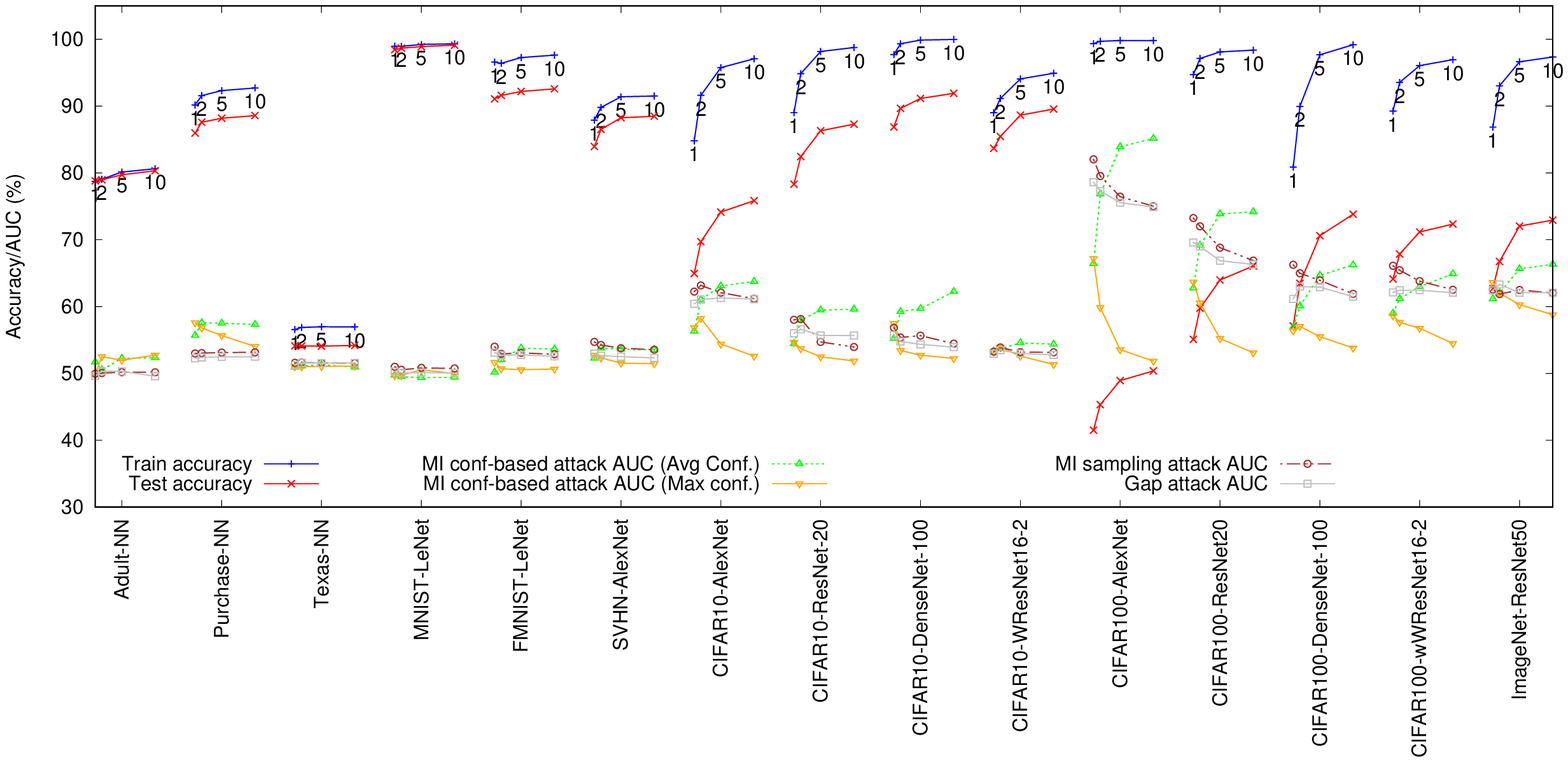}
\caption{Membership inference attack results across all datasets/models. Each curve indicates an ensemble of 1 (non-EL), 2, 5, and 10 models from left to right. Green curves indicate original deep ensembles where confidence values are averaged. Yellow curves indicate an ensemble of same models using maximum agreed confidence. Note that train and test accuracy of both approaches are the same. MI sampling and Gap attack is conducted on original deep ensembles.}
\label{fig-all-datasets}
\end{figure*}

Figure~\ref{fig-all-datasets} shows that the effectiveness of membership inference attacks increases and outperforms the gap attack as deep ensembles are deployed. This raises significant privacy concern since the gap attack is often suggested as a baseline that is also hard to outperform in non-ensemble setting \cite{rezaei2020towards, leino2020stolen, choo2020label}. Note that gap attack can be viewed as a metric directly reflecting the generalization gap rather than a reliable membership inference. As suggested in \cite{rezaei2020towards}, we can separate correctly classified samples and misclassified samples to understand why membership inference attacks can barely outperform gap attack. As shown in Figure~\ref{fig-dist}(b) and (c), the distributions of train and test samples are similar when separated into correctly classified and misclassified sets. The reason why the distribution of all samples (Figure~\ref{fig-dist}(a)) looks more distinguishable when correctly classified samples and misclassified samples are combined is that there are often more misclassified samples in the test set than the train set. This is the information that gap attack exploits which essentially reflects the generalization gap. In order for a membership inference attack to considerably outperform the gap attack, the distribution of correctly classified samples and misclassified samples should leak membership status information, which is not often the case as it is shown in Figure~\ref{fig-dist}(b) and (c), and \cite{rezaei2020towards}.

When ensembling is used, the distribution of confidence values changes dramatically, as explained in Section \ref{section-dist-change}. By comparing the confidence distribution of correctly classified samples in an ensemble (Figure~\ref{fig-dist}(e)) with a non-ensemble scenario (Figure~\ref{fig-dist}(b)), the distribution is clearly more distinguishable in ensemble case. This is of significant privacy concern because, as discussed in \cite{rezaei2020towards}, majority of samples in practice belong to the correctly classified set. Similar trend is also observable in misclassified samples (Figure~\ref{fig-dist}(f)). Hence, the confidence values, that barely leak more information than generalization gap itself in a single model scenario, now considerably leak more membership information than just the generalization gap. That is the reason why membership inference attacks are significantly more effective in deep ensembles in comparison to the gap attack.

\section{Experiments Results}
\label{sec-eval}

\subsection{Experimental Setup}
\label{sec-setup}
We explore a wide range of datasets that are often used in deep ensemble literature or membership inference literature: Adult\footnote{http://archive.ics.uci.edu/ml/datasets/Adult}, Texas, Purchase \cite{shokri2017membership}, MNIST \cite{lecun1998gradient}, FMNIST \cite{fmnist}, SVHN \cite{netzer2011reading}, CIFAR10 \cite{krizhevsky2009learning}, CIFAR100 \cite{krizhevsky2009learning}, and ImageNet \cite{russakovsky2015imagenet}. For non-image datasets (Adult, Purchase, and Texas), we use a fully connected neural network consisting of a hidden layer of size 128 and a Softmax layer. All other training parameters for these datasets are set as suggested in \cite{shokri2017membership}. For image datasets, we use a wide range of convolutional neural networks depending on the input dimension and the difficulty of the task. We use the model implementations adopted in \cite{nasr2019comprehensive, rezaei2020towards}\footnote{https://github.com/bearpaw/pytorch-classification}. We train 10 models for each dataset with random initialization and construct an ensemble of 2, 5 and 10 models, respectively.


Attack models for Shokri and Watson attacks are NNs with three hidden layers of size 128, 128, and 64, respectively. In this section, we consider a scenario that is most advantageous to the attacker where $80\%$ of the training dataset is given to the attacker and the goal is to infer the membership of the remaining samples, similar to \cite{rezaei2020towards}. This can be construed as an upper-bound for the confidence-based attacks that does not use difficulty calibration. We explore Shokri and Watson attacks in the next sections. For sampling attack, we perturb each sample 50 times and count the number of time the prediction label has changes, as in \cite{rahimian2020sampling}. For ImageNet, we attack a set of samples including 50 member and nonmember images per class. We explore a random set of ten hyper-parameters, including the one proposed in \cite{rahimian2020sampling}, for the noise perturbation and report the highest attack performance. Here, we only report AUC of membership inference attacks. In practice, the attacker needs to train shadow models to estimate the best threshold value which may result in less accurate attack. All other training parameters are set as suggested in \cite{rezaei2020towards}. 
See Appendix \ref{appendix-all-defense} for more results. The results of the weighted averaging, and snapshot ensembles and diversified ensemble networks are shown in Appendix \ref{appendix-all-epochs-stacking}, and Appendix \ref{appendix-all-epochs-advanced-ensemble}, respectively.

Figure~\ref{fig-all-datasets} shows the results on all datasets. For some datasets, such as Adult, Texas, and MNIST, deep ensemble approach barely changes the accuracy or privacy. That is because the disagreement across models is insignificant in these datasets. For all other datasets, deep ensemble approach improves the accuracy (blue/red curves) as well as the effectiveness of confidence-based membership inference attacks (green curves). As mentioned in Section \ref{section-factors}, the most salient factor in membership inference effectiveness on deep ensembles is the accuracy gap between train and test set. Figure~\ref{fig-all-datasets} clearly shows that whenever this generalization gap is large for non-ensemble case, the attack improvement is significant using ensembling. It is worth noting that the ensembling can often reduce the generalization gap and the effectiveness of the gap attack (e.g., CIFAR10-DenseNet-100, CIFAR100-AlexNet, or CIFAR100-ResNet20). However, due to the reasons explained in Section \ref{sec-why-gap}, the membership inference effectiveness still increases.

Interestingly, the effectiveness of sampling attack, unlike confidence-based attacks, often decreases on deep ensembles (brown curves). The main reason is that deeps ensembles are more robust than a single model, as shown in \cite{yang2021certified, liang2020towards}. Therefore, perturbing target samples to obtain information about its membership status is less effective in deep ensembles. Another observation is that maximum agreed confidence ensembling (yellow curves) can considerably mitigate the effectiveness of confidence-based membership inference.

\begin{figure*}
\def\tabularxcolumn#1{m{#1}}
\begin{tabularx}{\linewidth}{@{}cXX@{}}
\begin{tabular}{ccc}
\subfloat[Adult (NN)]{\includegraphics[width=0.29\linewidth]{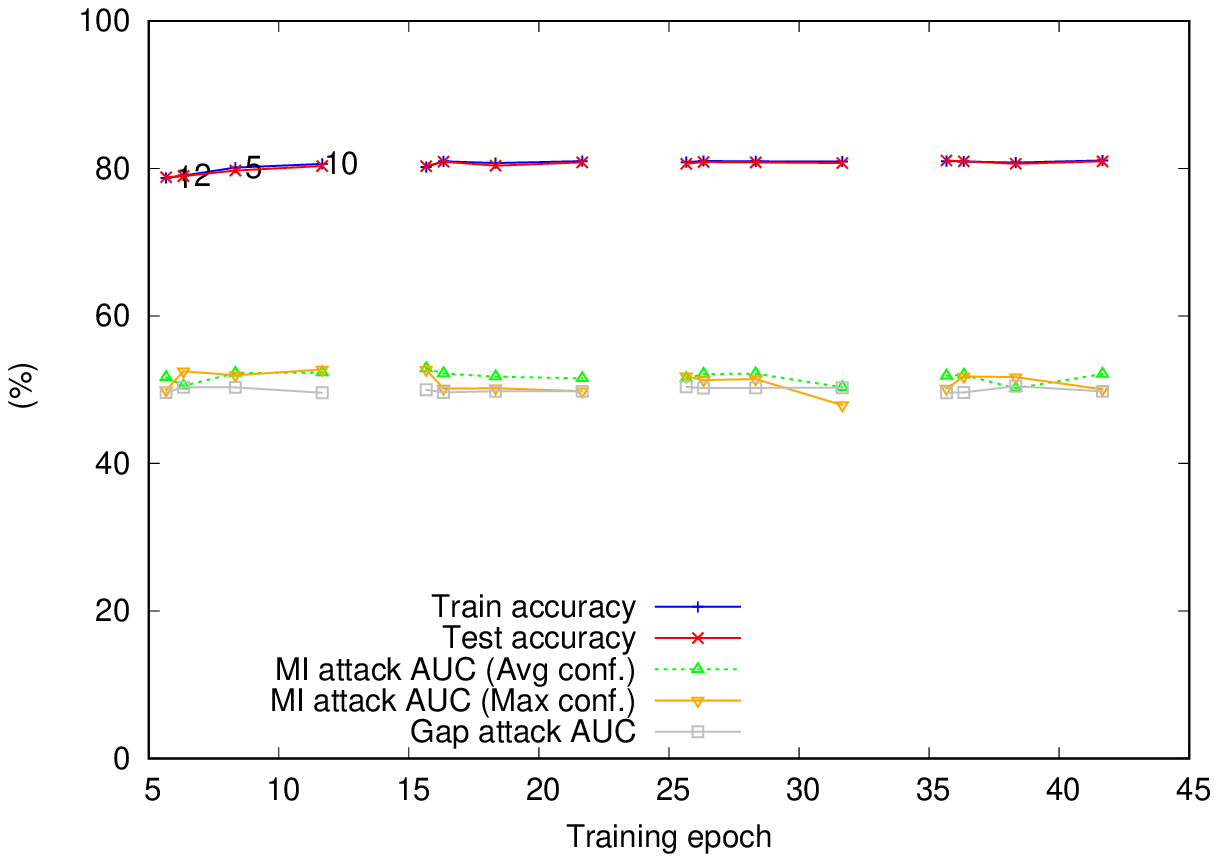}} 
  & \subfloat[Texas (NN)]{\includegraphics[width=0.29\linewidth]{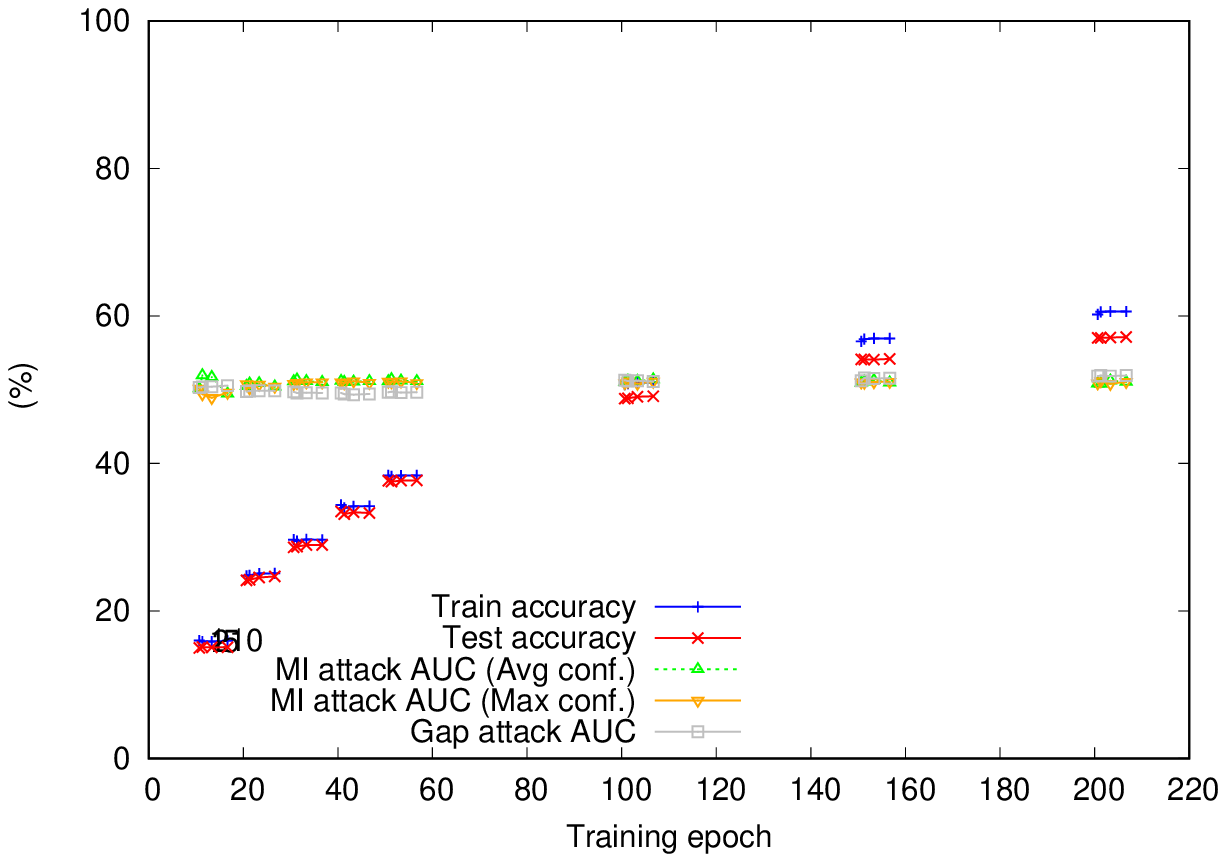}}
  & \subfloat[MNIST (LeNet)]{\includegraphics[width=0.29\linewidth]{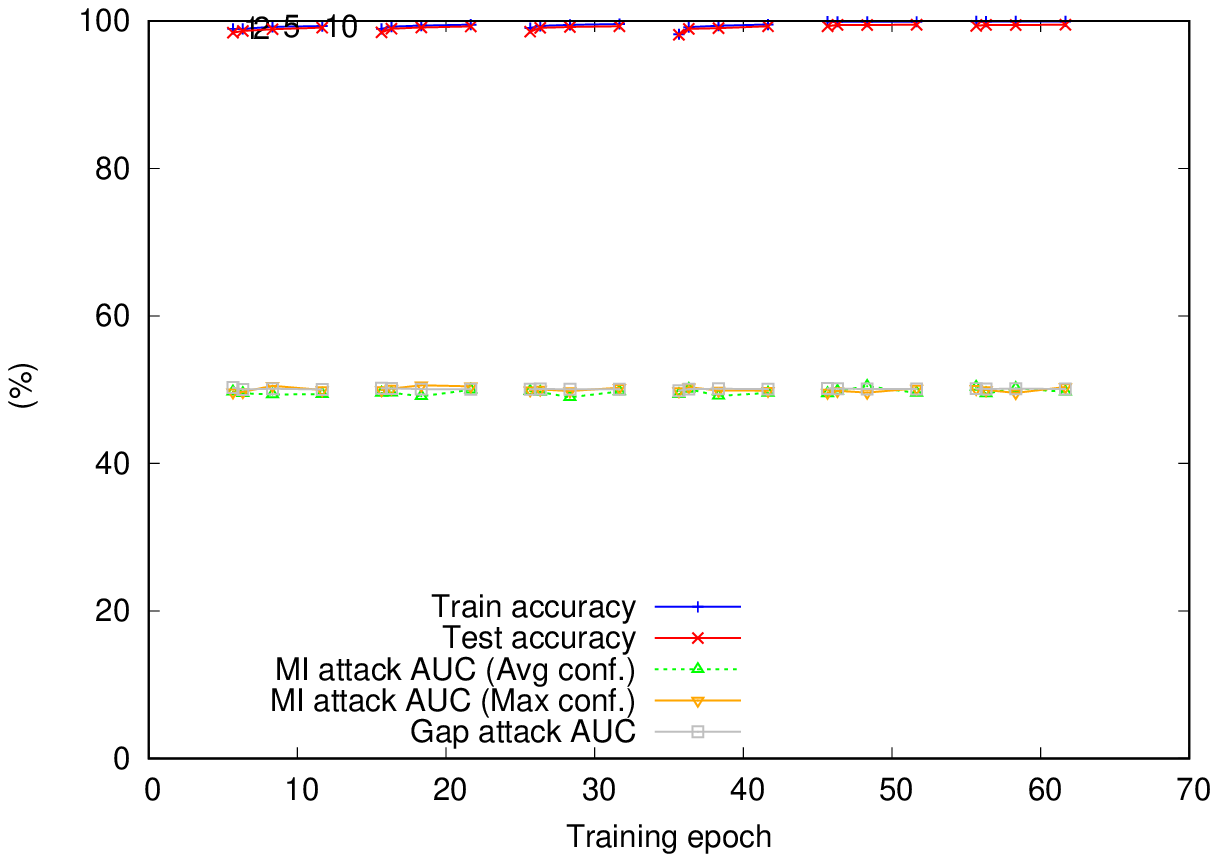}} \\
\subfloat[FMNIST (LeNet)]{\includegraphics[width=0.29\linewidth]{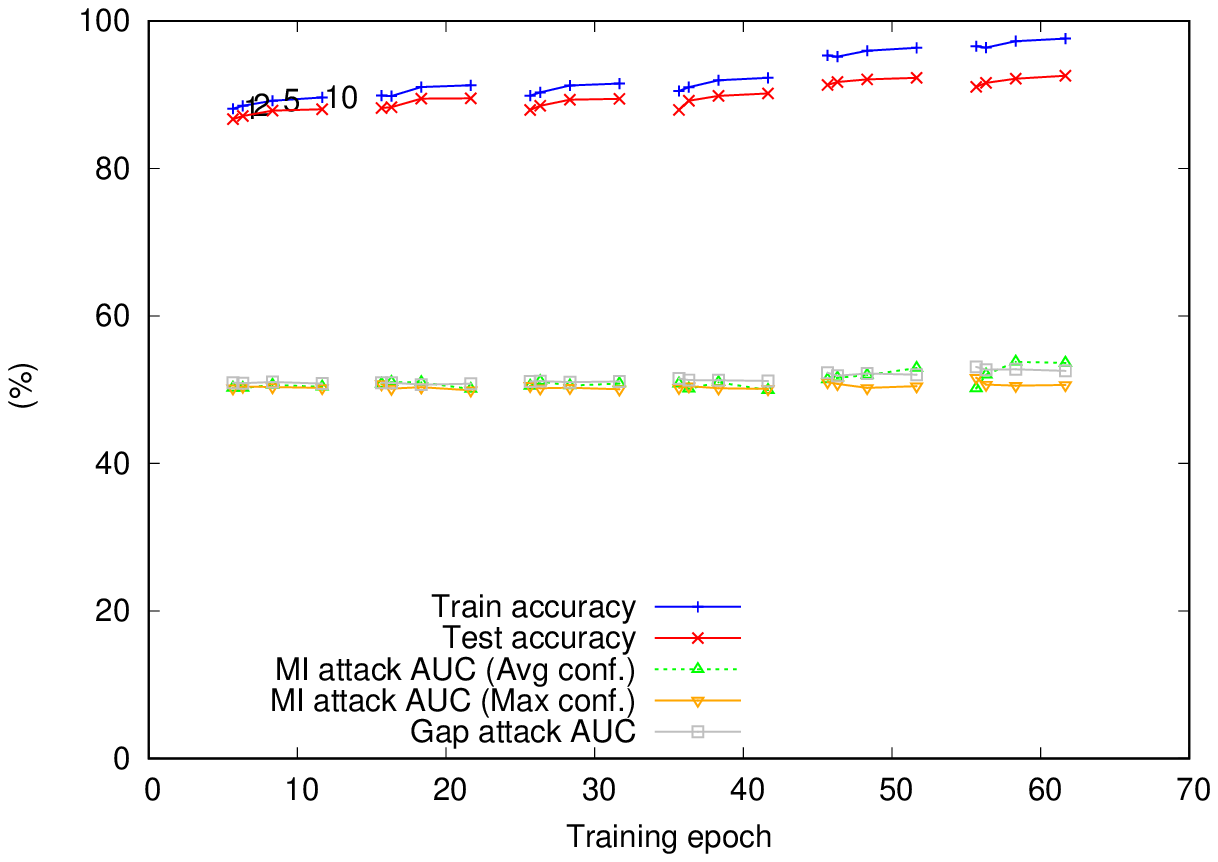}} 
  & \subfloat[SVHN (AlexNet)]{\includegraphics[width=0.29\linewidth]{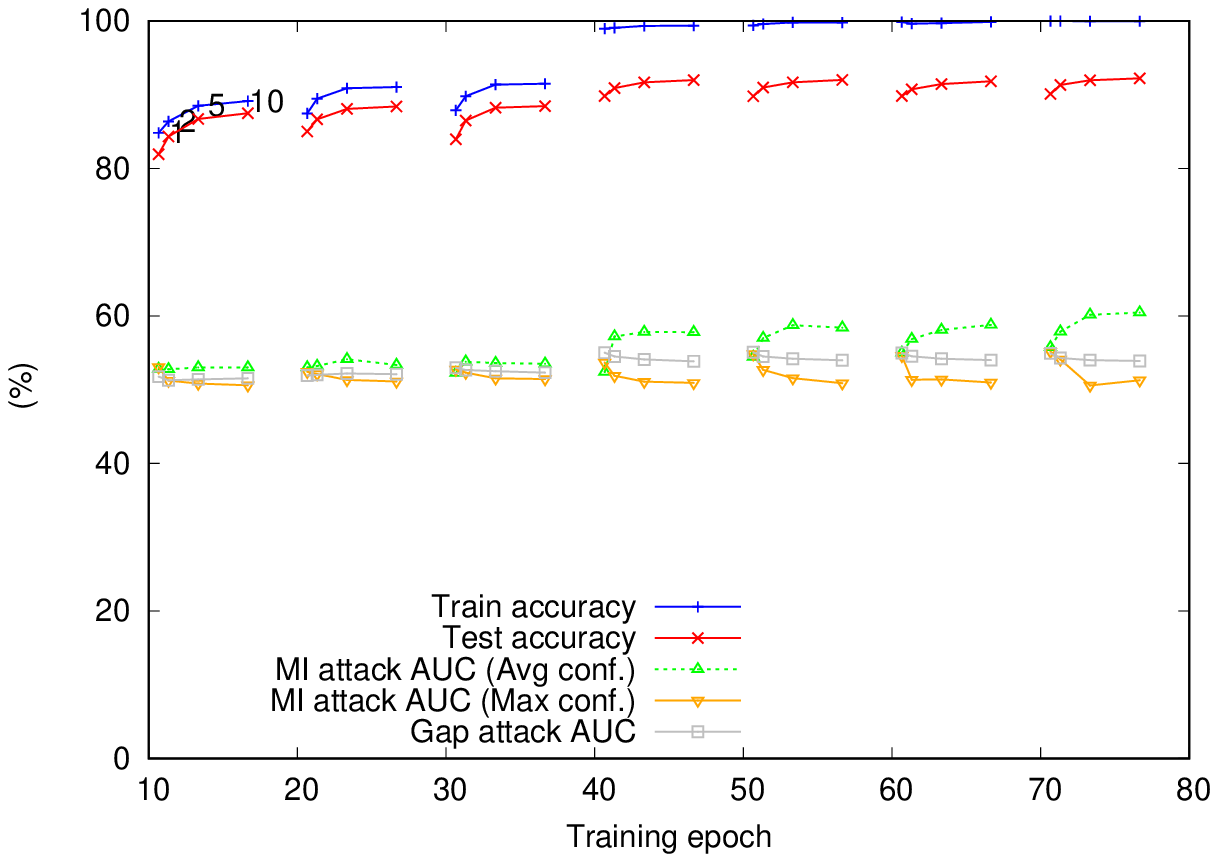}}
  & \subfloat[SVHN (ResNet20)]{\includegraphics[width=0.29\linewidth]{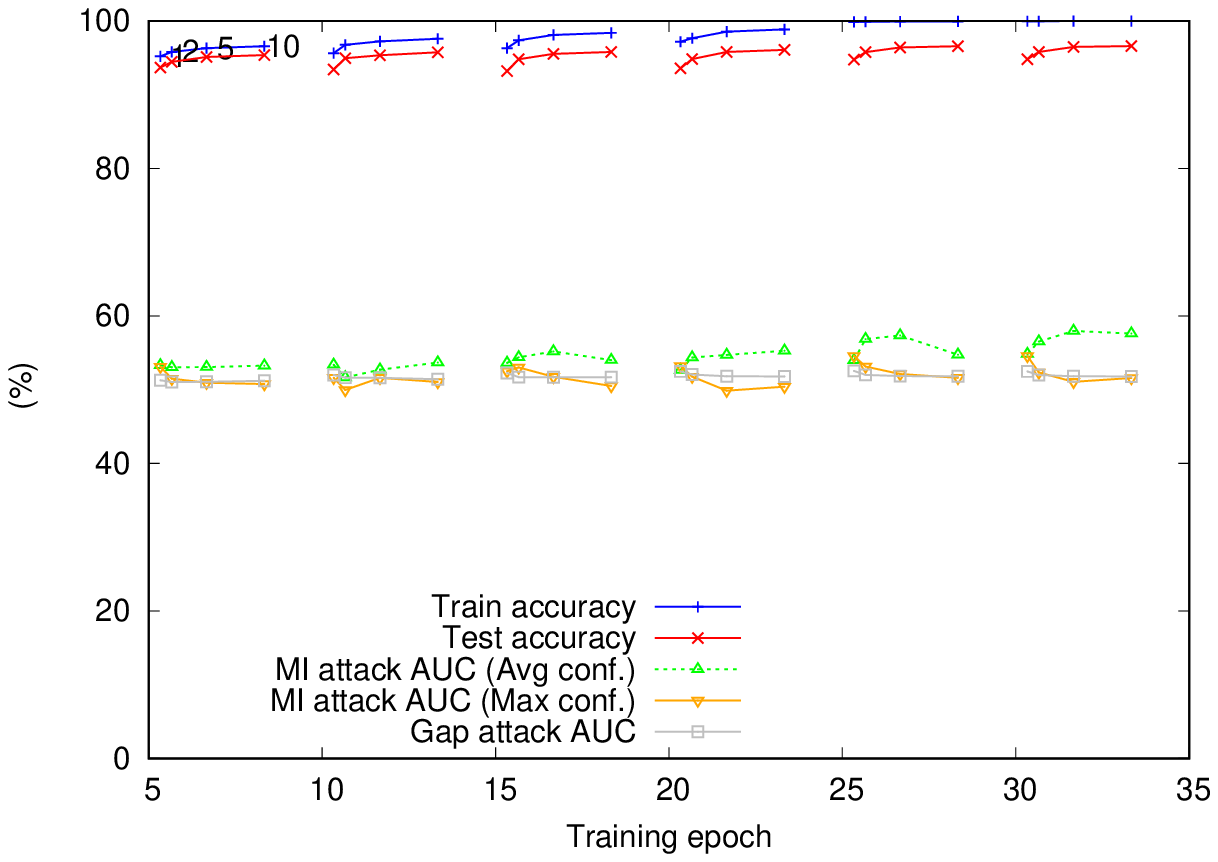}} \\
\subfloat[CIFAR10 (AlexNet)]{\includegraphics[width=0.29\linewidth]{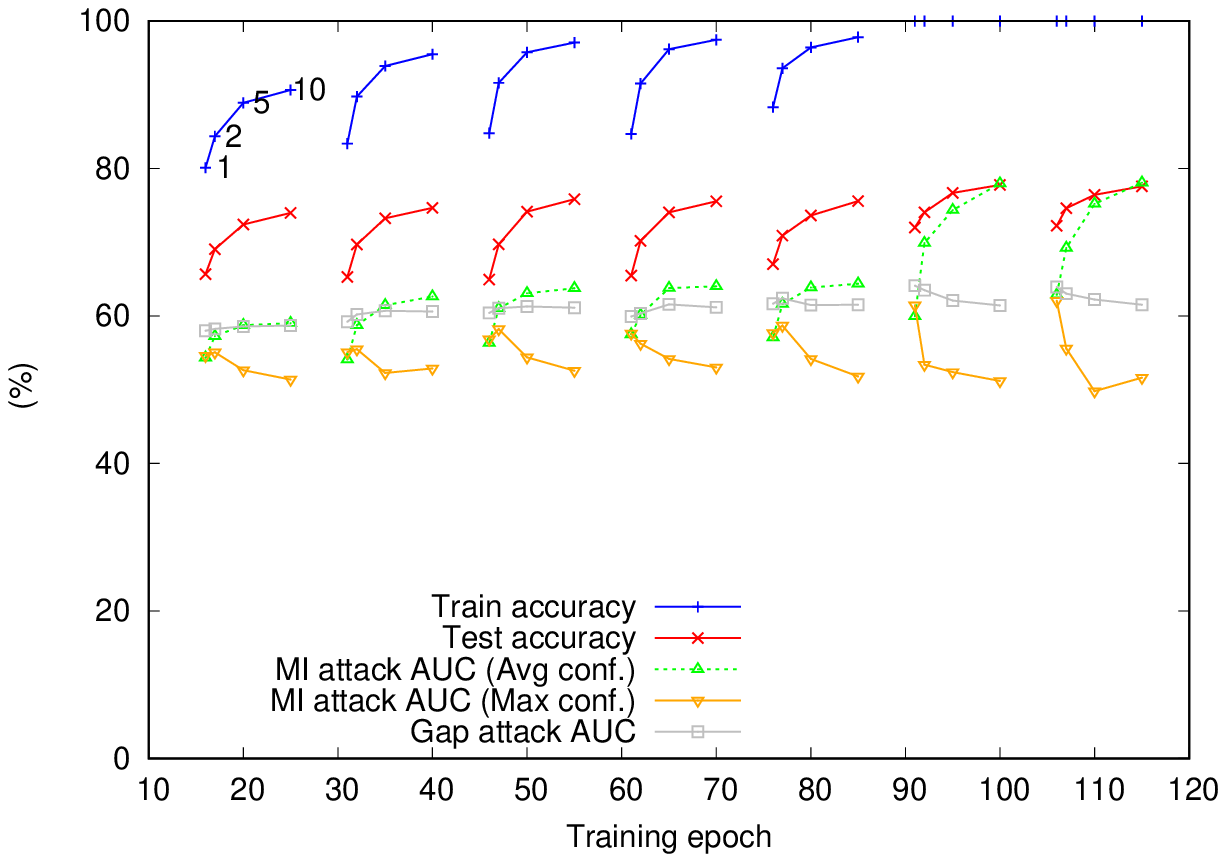}} 
  & \subfloat[CIFAR10 (ResNet20)]{\includegraphics[width=0.29\linewidth]{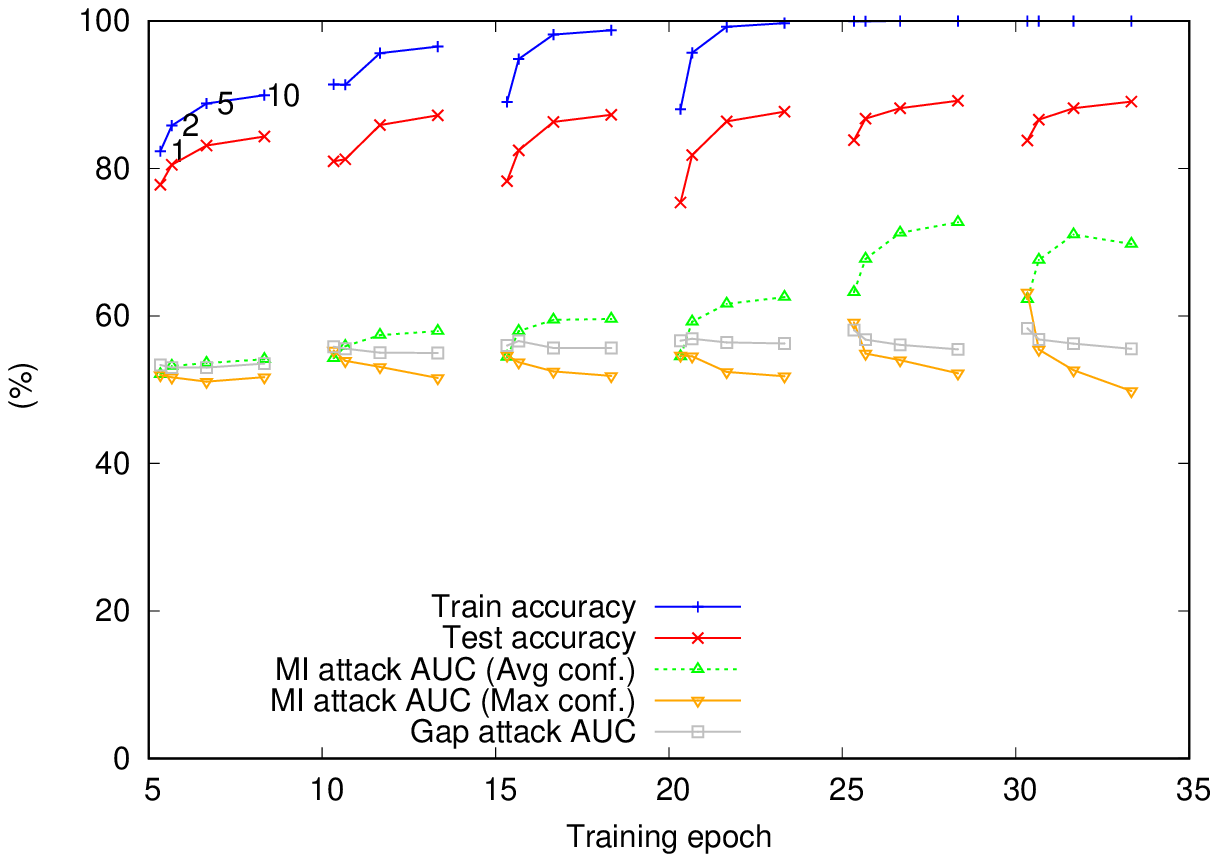}}
  & \subfloat[CIFAR10 (DenseNet100)]{\includegraphics[width=0.29\linewidth]{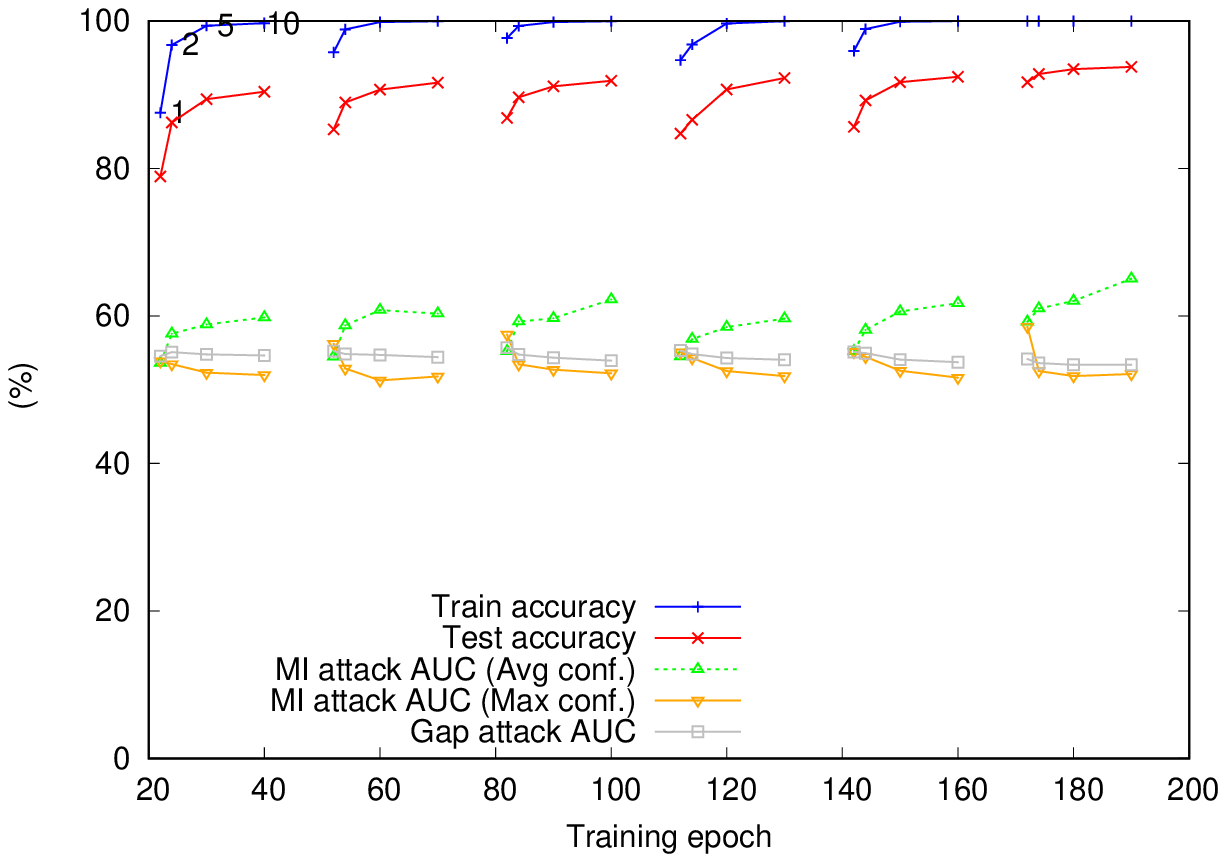}} \\
\subfloat[CIFAR10 (WResNet16-2)]{\includegraphics[width=0.29\linewidth]{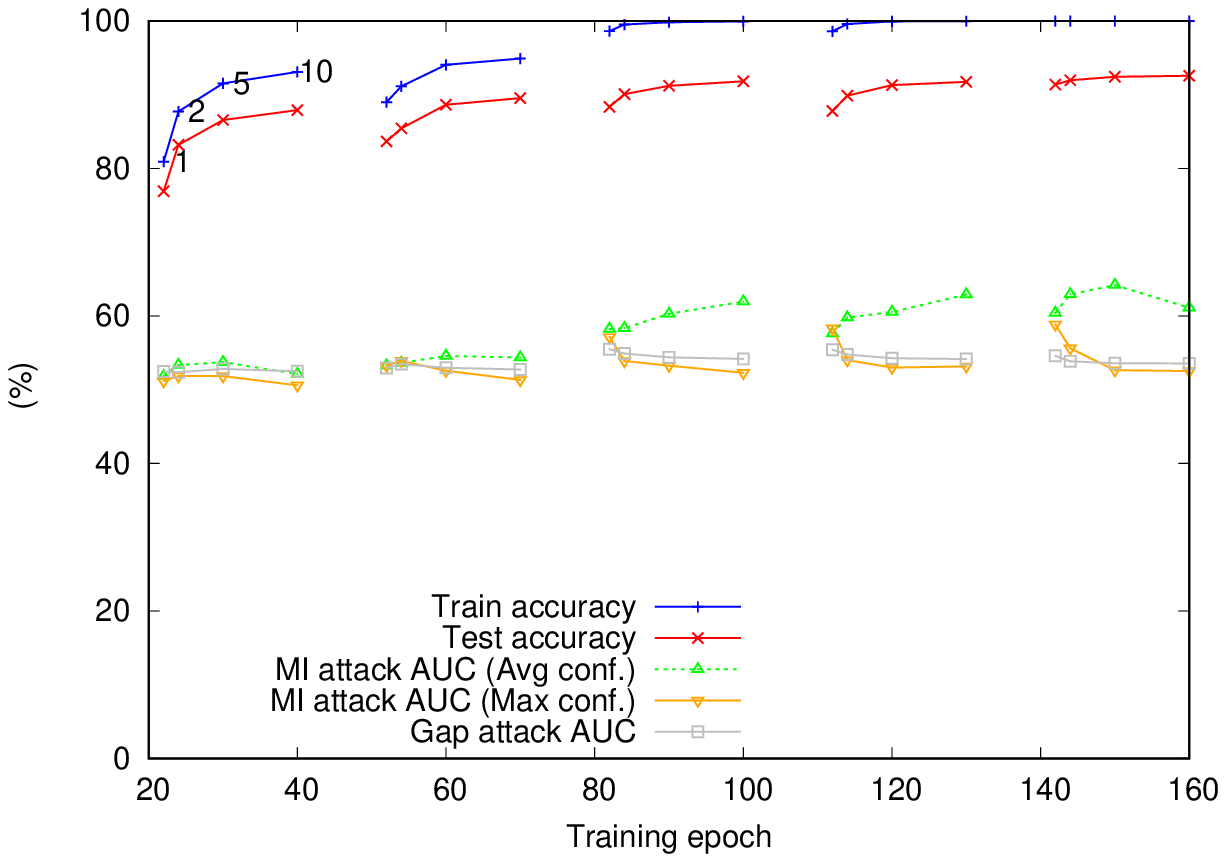}} 
  & \subfloat[CIFAR100 (AlexNet)]{\includegraphics[width=0.29\linewidth]{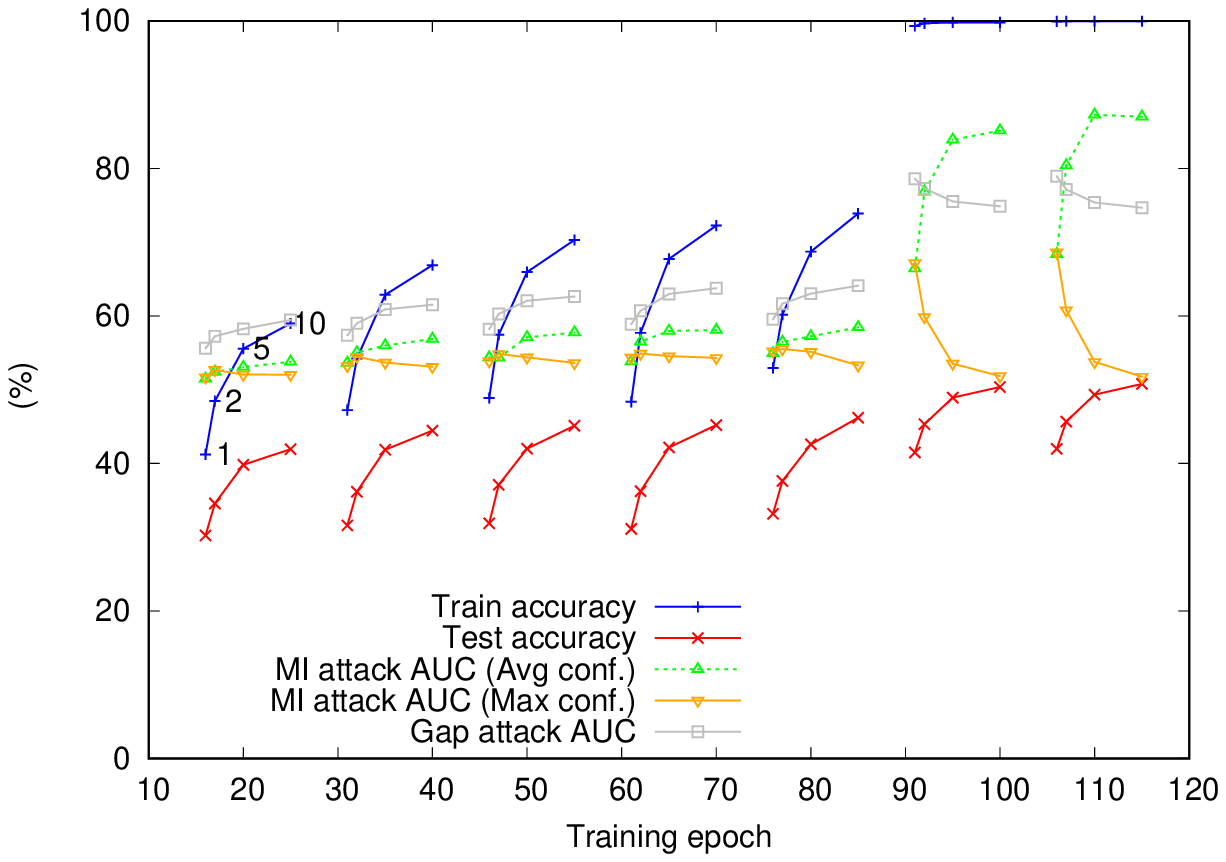}}
  & \subfloat[CIFAR100 (ResNet20)]{\includegraphics[width=0.29\linewidth]{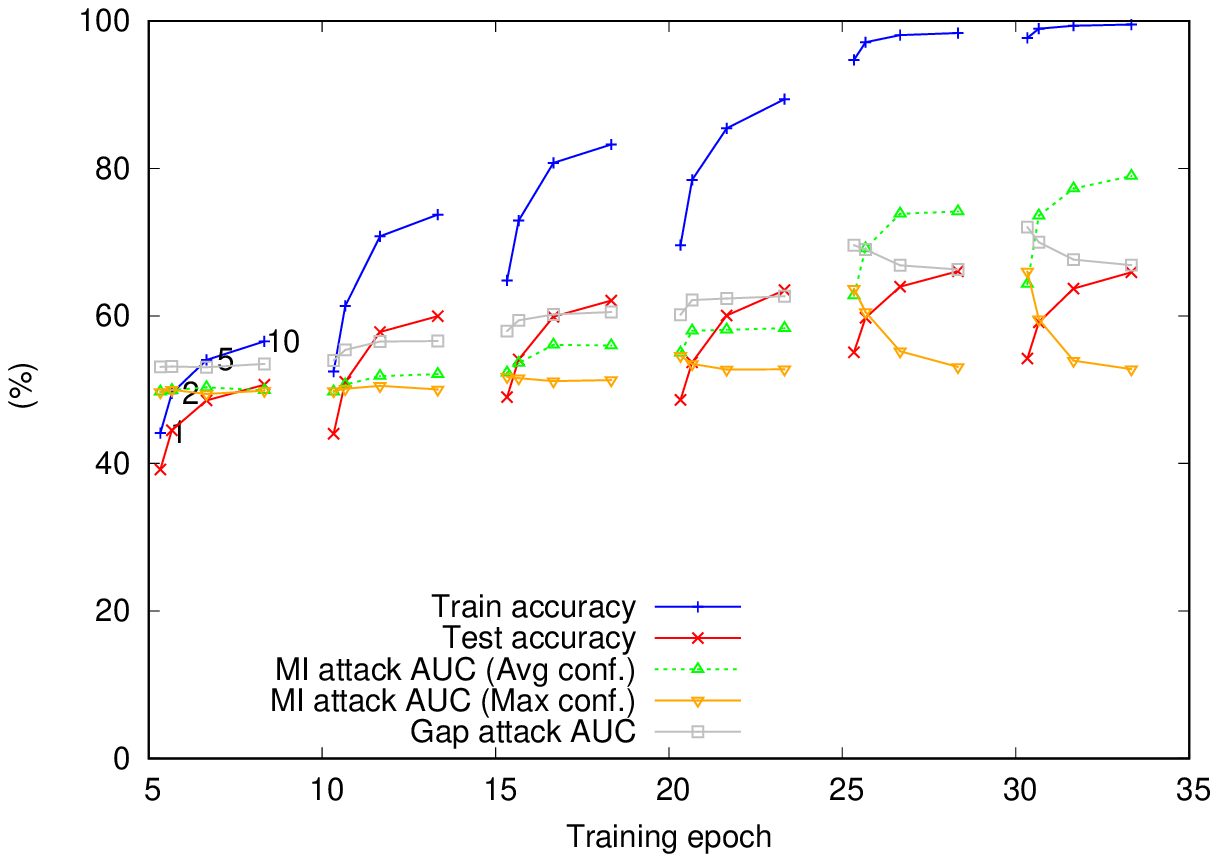}} \\
\subfloat[CIFAR100 (DenseNet100)]{\includegraphics[width=0.29\linewidth]{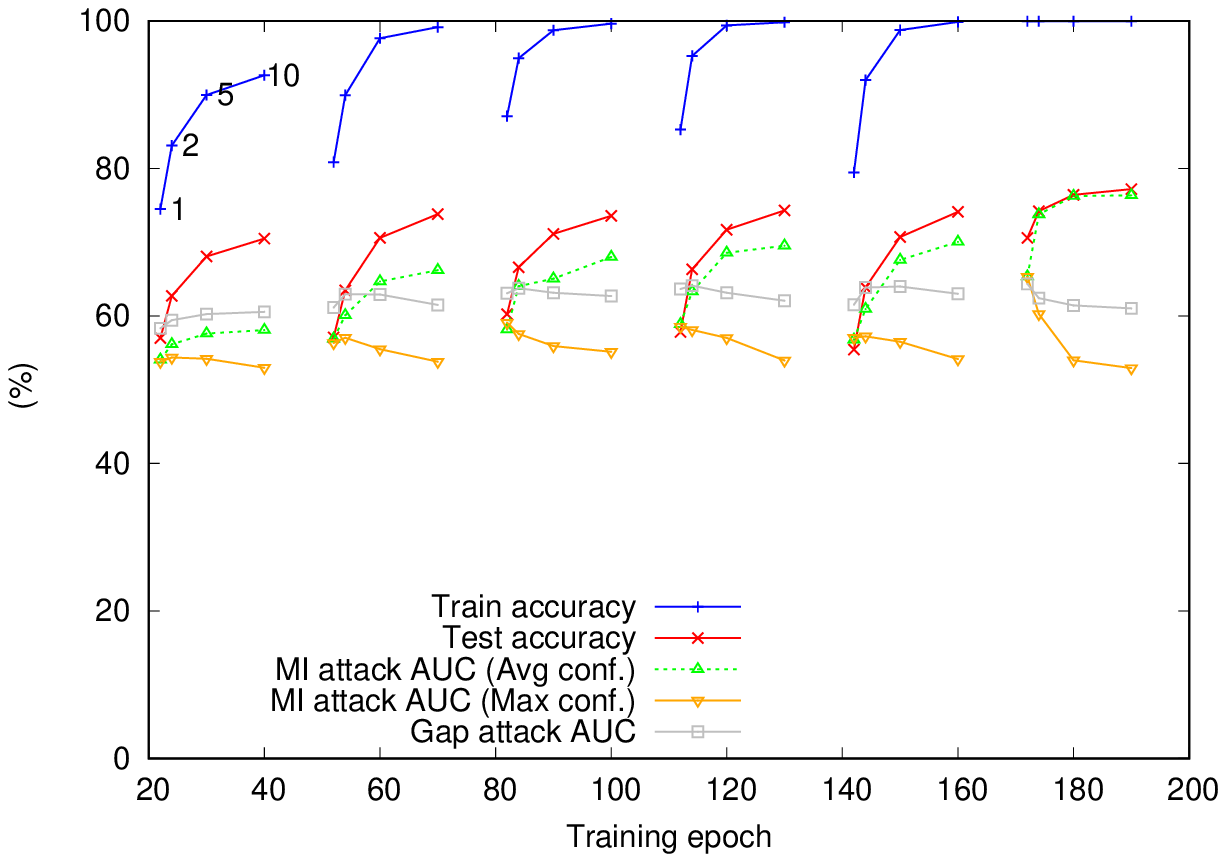}} 
  & \subfloat[CIFAR100 (WResNet16-2)]{\includegraphics[width=0.29\linewidth]{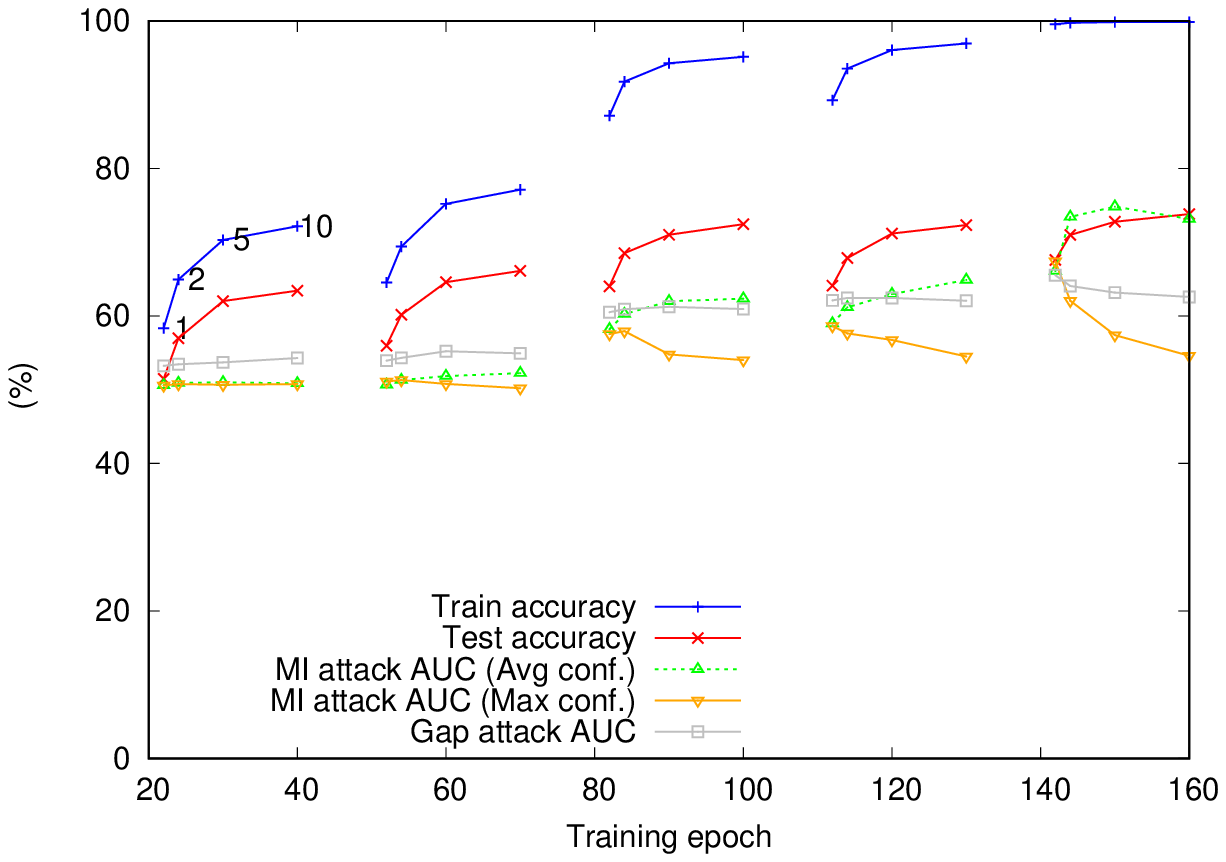}}
  & \subfloat[ImageNet (ResNet50)]{\includegraphics[width=0.29\linewidth]{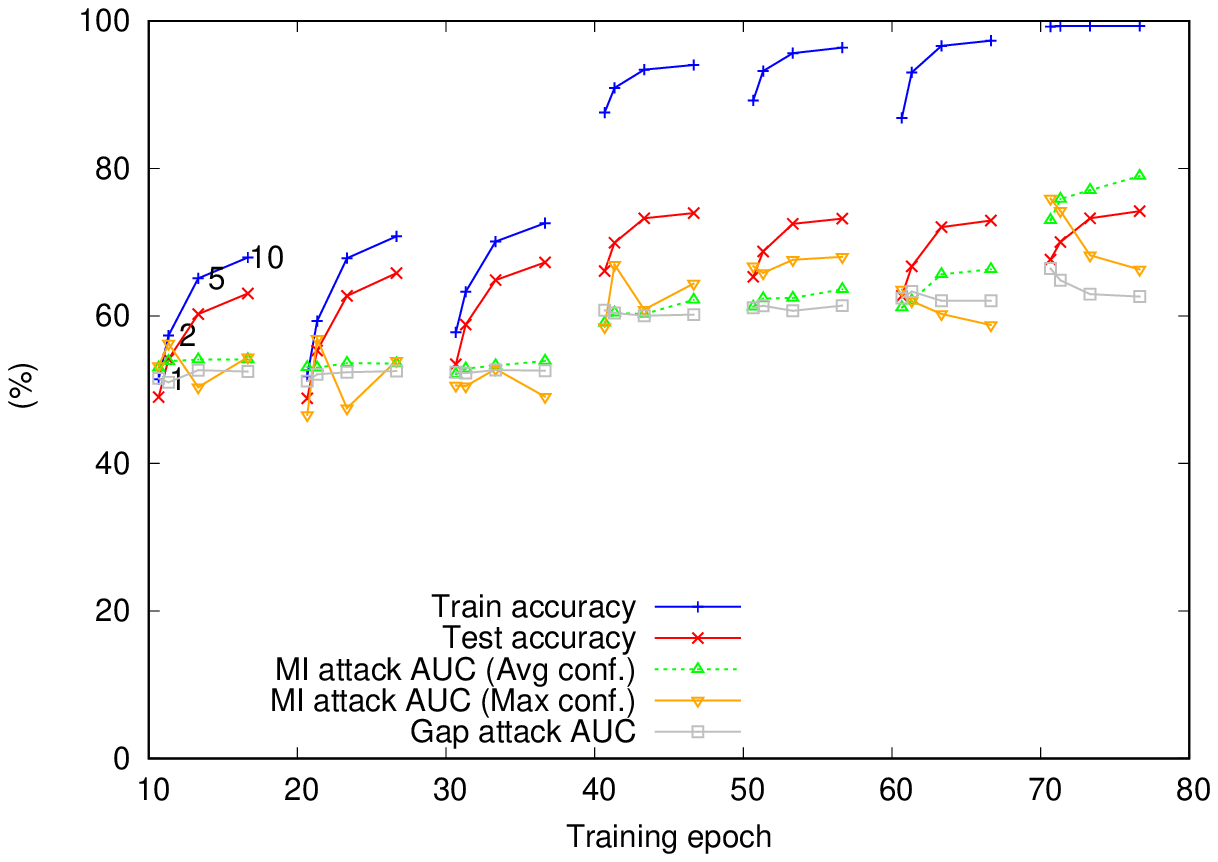}} \\
\end{tabular}
\end{tabularx}
\caption{Target models' accuracy and MI attacks' AUC across all datasets and models. }\label{fig-all-epochs}
\end{figure*}

Figure~\ref{fig-all-epochs} demonstrates the improvement of accuracy and MI attack over various training epochs. For datasets that ensembling outperforms a single model, using an ensemble of underfitted models is less prone to MI attack. However, it leads to lower accuracy. Due to the computational cost of running sampling attack, we only perform the sampling attack on certain epochs in Figure \ref{fig-all-datasets}.

\subsection{Defense mechanisms}
\label{sec-defenses}

\begin{figure*}
\def\tabularxcolumn#1{m{#1}}
\begin{tabularx}{\linewidth}{@{}cXX@{}}
\begin{tabular}{ccc}
\subfloat[CIFAR10-AlexNet: Shokri's attack]
{\includegraphics[width=0.30\linewidth]{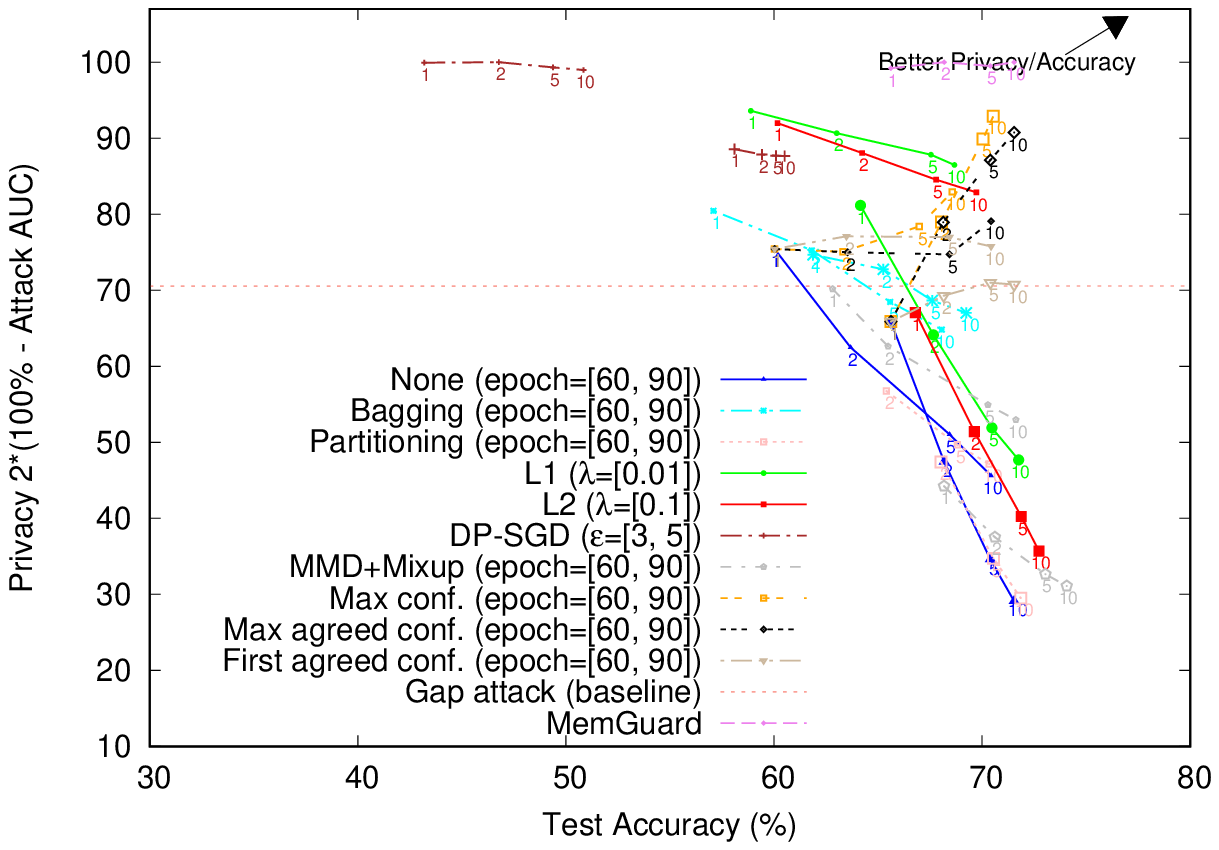}} 
   & 
\subfloat[CIFAR10-AlexNet: Watson's MI attack]
{\includegraphics[width=0.30\linewidth]{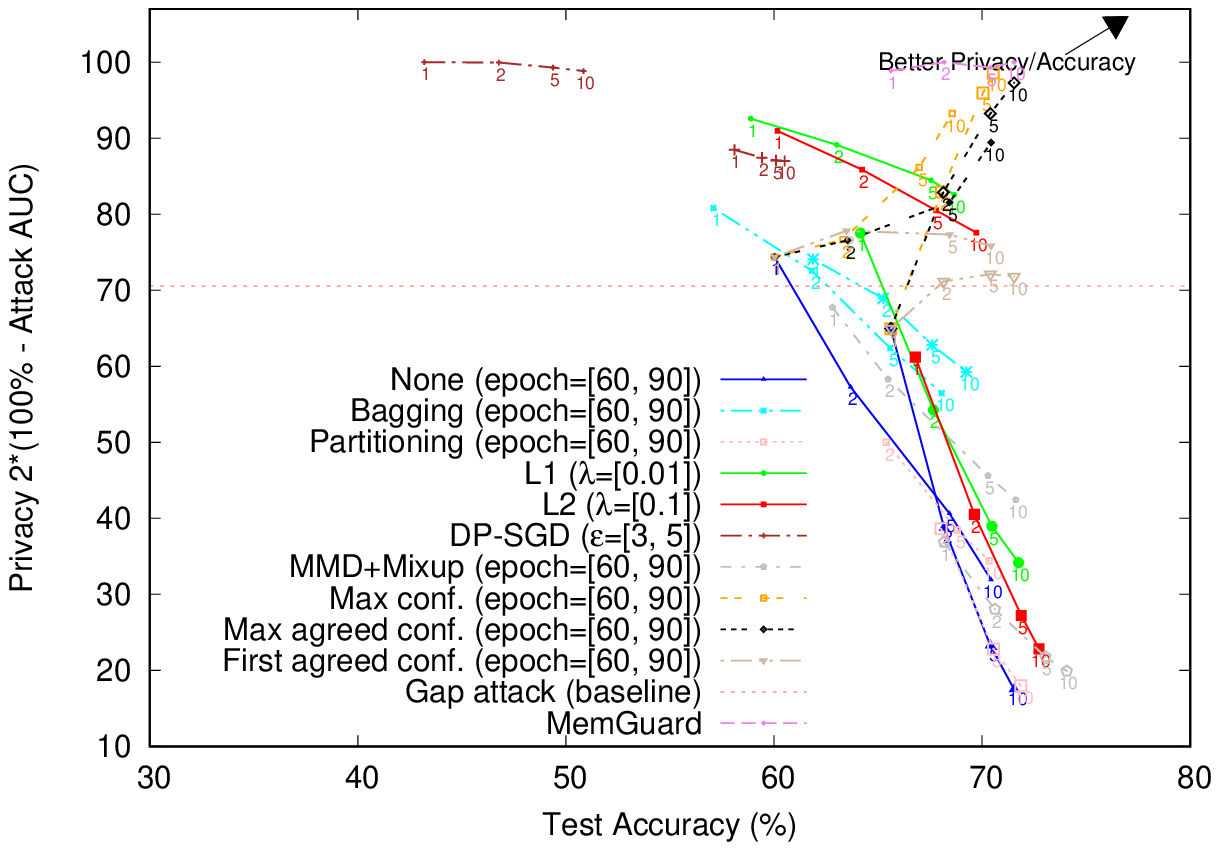}}
   &
\subfloat[CIFAR10-AlexNet: Sampling MI attack]
{\includegraphics[width=0.30\linewidth]{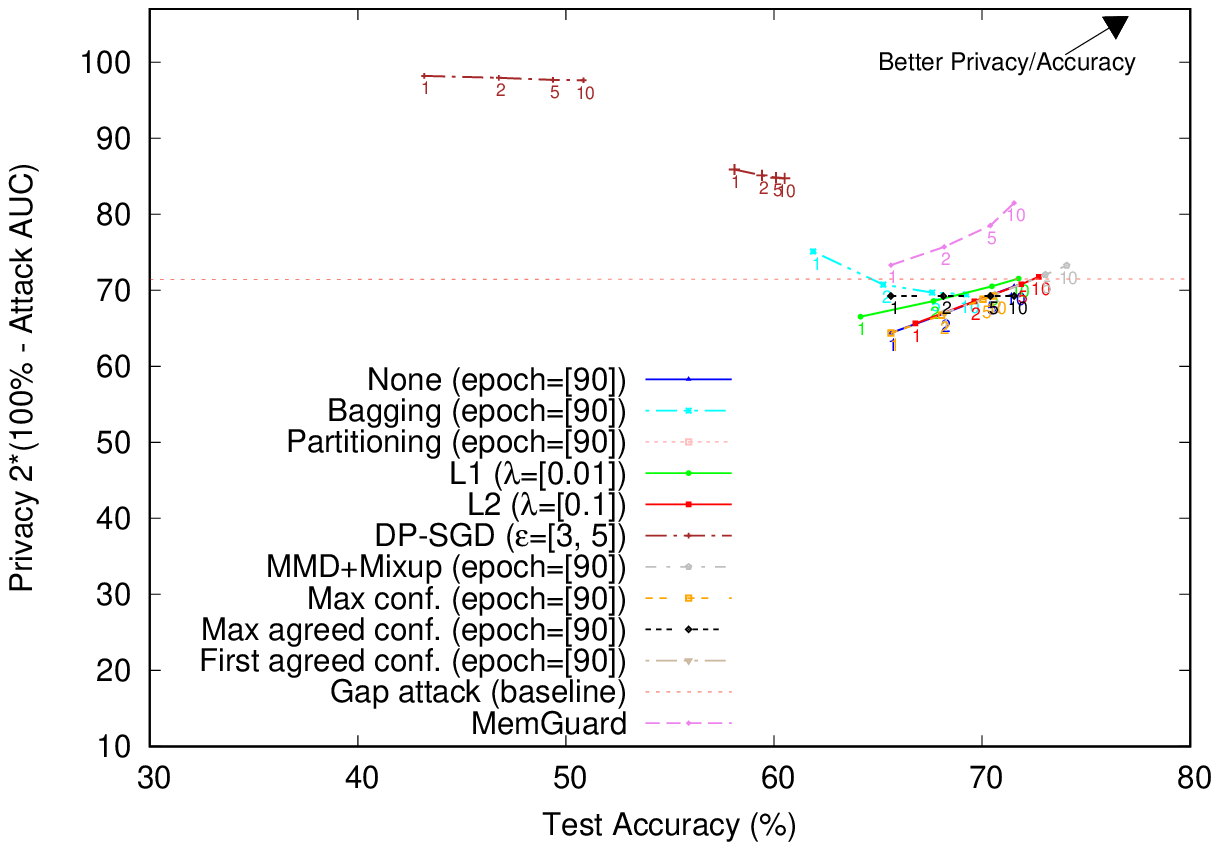}}\\

\subfloat[CIFAR100-AlexNet: Shokri's attack]
{\includegraphics[width=0.30\linewidth]{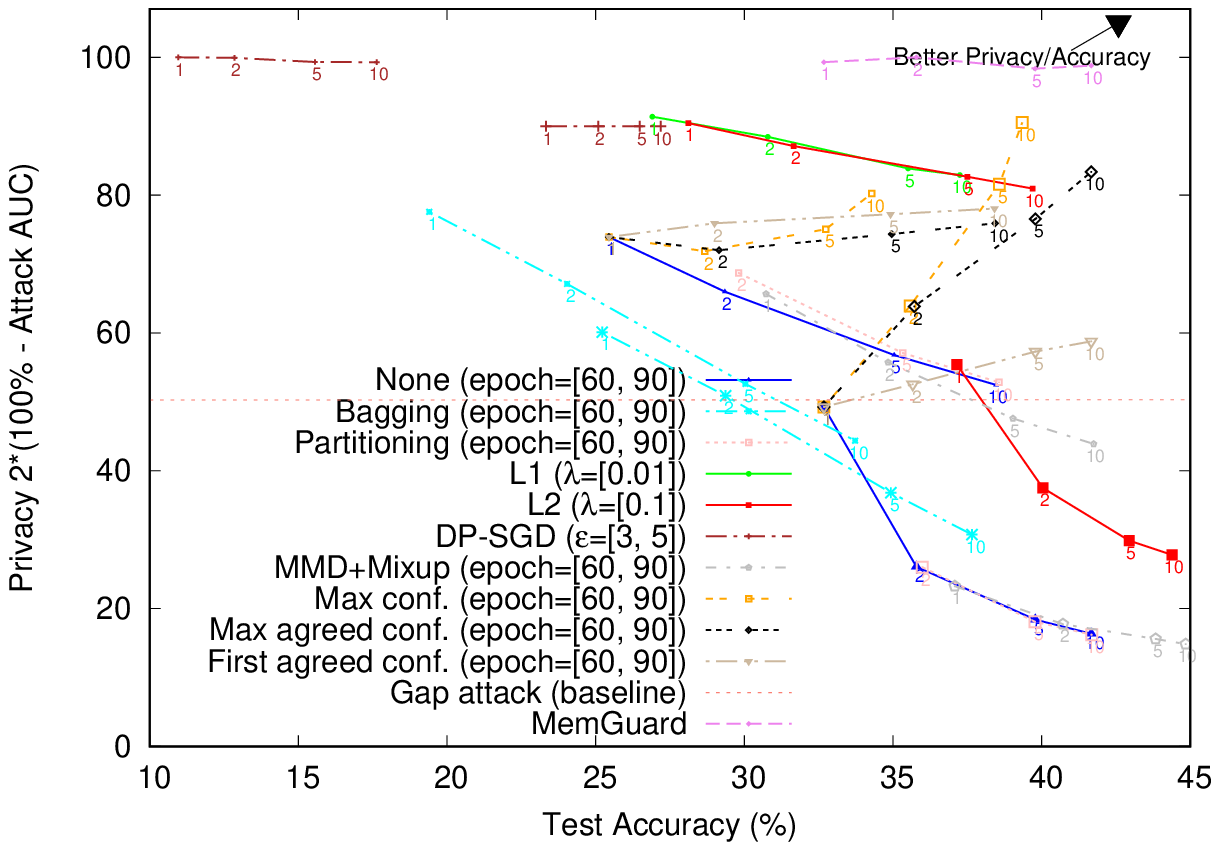}} 
   & 
\subfloat[CIFAR100-AlexNet: Watson's MI attack]
{\includegraphics[width=0.30\linewidth]{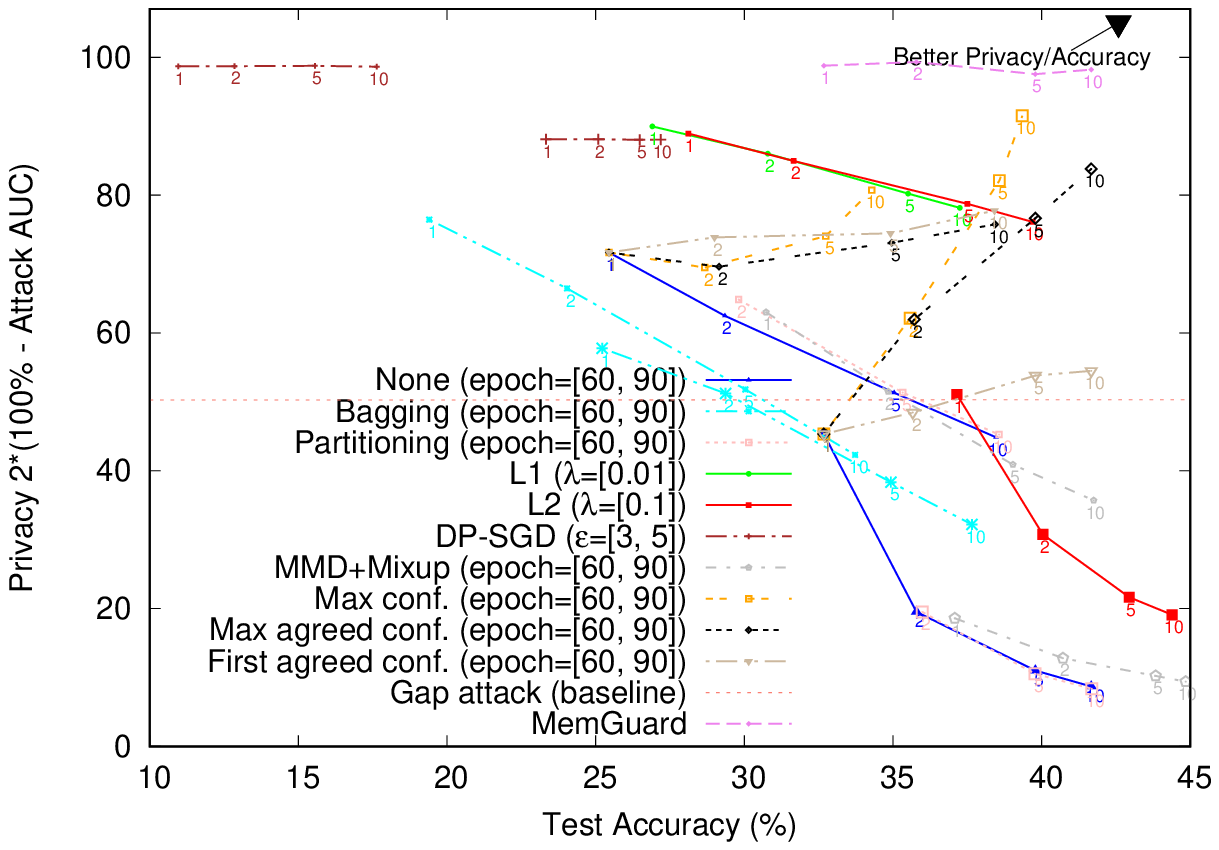}}
   & 
\subfloat[CIFAR100-AlexNet: Sampling MI attack]
{\includegraphics[width=0.30\linewidth]{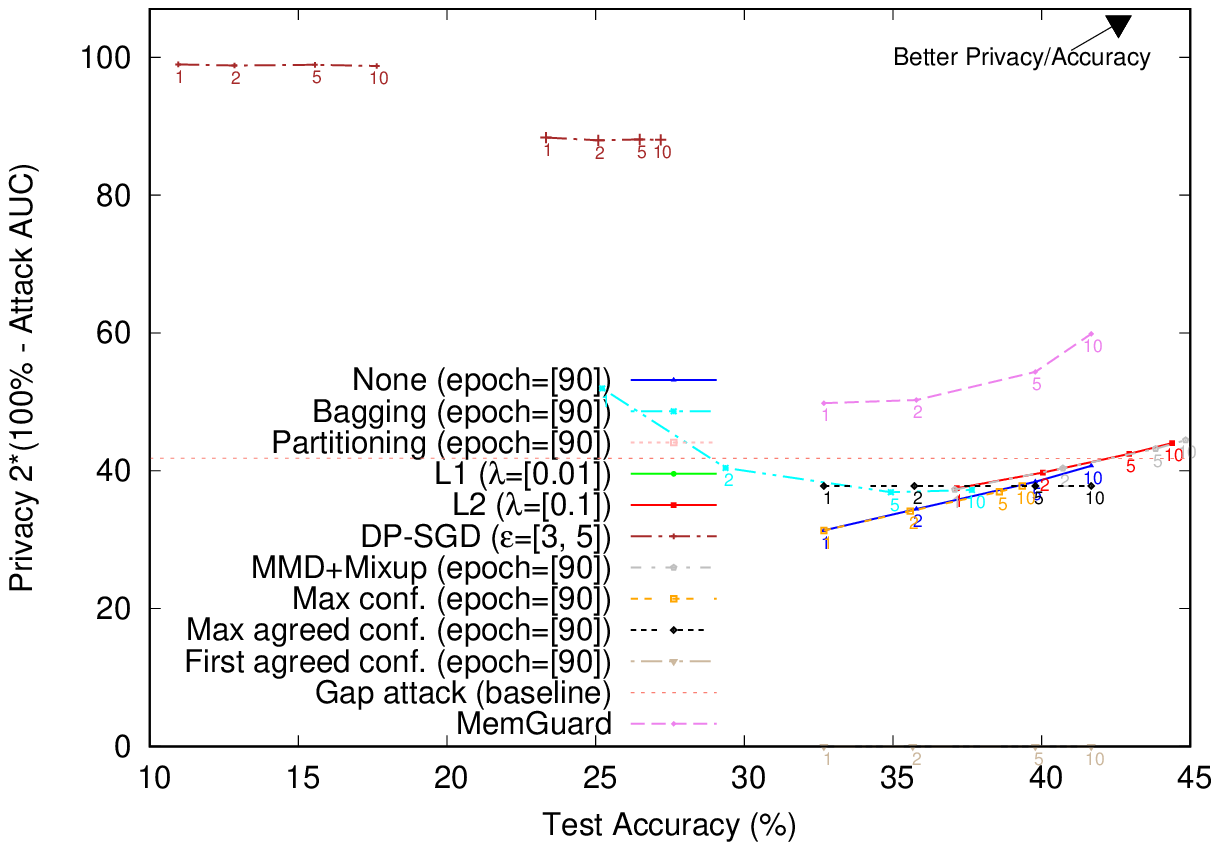}}\\

\end{tabular}
\end{tabularx}
\caption{Effect of defense mechanisms on an AlexNet model trained on CIFAR10 and CIFAR100. The size of each point indicates the relative value of its parameter. The horizontal pink line indicates the performance of gap attack on a deep ensemble of 10 models.}\label{fig-trade-off}
\end{figure*}

As discussed in previous sections, the less the generalization gap of base models are, the less effective the membership inference would be on the deep ensemble. Therefore, any standard regularization technique can potentially work as a defense mechanism. In this paper, we study L1 and L2 regularization, DP-SGD\footnote{We use Opacus implementation: https://github.com/pytorch/opacus.} ($\epsilon \approx [3,5]$ and $\delta=10^{-5}$), and MMD+Mixup \cite{li2021membership}.  For defense mechanisms, we use L1 and L2 regularization with 0.01 and 0.1 as a weight of the loss function, respectively. With only a single mode, we achieve ~42\% and 60\% accuracy using DP-SGD. The reason model accuracy is slightly lower than the literature \cite{tramer2020differentially} is due to the fact that we train victim models with half of the training data. The other half is reserved for MI attacker.Furthermore, one can simply terminate training early to keep the model's weights in less overfitted region. Moreover, some ensembling techniques, such as bagging, and partitioning, limits the access of models to all training samples, which can potentially reduce membership inference effectiveness on deep ensembles. 
Moreover, see Appendix \ref{appendix-all-epochs-stacking} for weighted averaging ensembles. We also evaluate two state-of-the-art deep ensembling approaches, namely snapshot ensemble and diversified ensemble network, in Appendix \ref{appendix-all-epochs-advanced-ensemble}.

In the section, we show the effectiveness of Shokri, Watson, and sampling attacks on all defense mechanisms. We divide the training set of each dataset into two disjoint sets: victim's training data, $D_{tr}$, and attacker's (shadow) training data, $D_s$. We use $D_{tr}$ to train base models for the victim's ensemble and $D_s$ to train 10 shadow models for Shokri and Watson attacks. For the Shokri attack, we train 10 shadow models with the same architecture and training hyper-parameters as victim models. Note that it has shown that Shokri attack's accuracy barely changes even if the architecture and training hyper-parameters of shadow models do not match the victim model \cite{salem2018ml}. For the Watson attack, we use the Shokri's attack output as the base membership inference score and we calibrate it using 10 other shadow models trained on $D_s$, as explained in \cite{watson2021importance}. The sampling attack configuration is similar to Section \ref{sec-setup}. Due to the lack of space, we only show the accuracy/privacy trade-off of AlexNet model trained on CIFAR10 and CIFAR100. See Appendix \ref{appendix-all-defense} for more results.

\textbf{Confidence-based attacks:} Figure~\ref{fig-trade-off} shows the effect of defense mechanisms on ensemble learning. The pink dashed line indicates the performance of gap attack on a deep ensemble of 10 models with no regularization. Hence, any point below this line means privacy leakage greater than a trivial baseline (Gap attack). Although Watson attack is generally more effective due to its difficulty calibration, both Shokri and Watson attacks change similarly when used against deep ensembles. We can observe a consistent trade-off between ensemble accuracy and privacy that resembles Pareto optimal points. The only exceptions are our proposed approaches, namely maximum agreed confidence, maximum confidence, and first agreed confidence. 
The difference between maximum agreed confidence and maximum confidence is marginal. The former achieves slightly better accuracy, while the latter achieves slightly better privacy (in terms of confidence-based MI attack).

Interestingly, none of the approaches that applies modification during the training of the base models could break the trade-off, including bagging, partitioning, L1/L2 regularizations, DP-SGD, and MMD+Mixup. Note that privacy degradation rate for these approaches is clearly not constant. An ensemble of heavily regularized models or under-fitted models barely causes more privacy leakage (e.g., L2 regularization at epoch 60). On the other hand, an ensemble of overfitted models (e.g., non-regularized models trained for 90 epochs) results in large privacy leakage. It is worth mentioning that some approaches are sometimes outperform the original deep ensemble both in terms of accuracy and privacy despite being bounded to the trade-off. For instance, an ensemble of $n$ deep models with L2 regularization (red curves) can often outperform the deep ensemble of $n$ models (blue curves) both in terms of accuracy and privacy. However, it still manifests the trade-off in a sense that increasing the number of L2 regularized models in an ensemble increases the accuracy while decreasing the privacy (in terms of confidence-based MI attack).

\begin{table*}[h]
\scriptsize
  \caption{Comparison of different defence mechanisms with respect to true positive in low false positive regime and average per-sample distortion to the confidence output. This table only includes Watson attack.}
  \label{tbl-pf}
  \centering
  \begin{tabular}{lllllllllllllll}
    \toprule
    Dataset & - & \multicolumn{4}{c}{TPR @ 0.001 @ FPR} & \multicolumn{4}{c}{TPR @ 0.1 @ FPR} & \multicolumn{4}{c}{Average per-sample confidence distortion} \\
    \cmidrule(r){3-6}
    \cmidrule(r){7-10}
    \cmidrule(r){11-14}
    -    & Ensemble size: & 1 & 2 & 5 & 10 & 1 & 2 & 5 & 10 & 1 & 2 & 5 & 10 \\
    \midrule
    & None (deep ensemble) & 0.08\% & 0.08\% & 1.16\% & 2.20\% & 0.78\% & 2.42\% & 4.88\% & 6.56\% & 0.0 & 0.0 & 0.0 & 0.0 \\
    & Bagging & 0.08\% & 0.19\% & 0.48\% & 0.52\% & 0.78\% & 0.71\% & 1.34\% & 1.72\% & 0.0 & 0.18 & 0.16 & 0.15\\
    & Partitioning & 0.08\% & 0.09\% & 1.09\% & 2.21\% & 0.78\% & 1.65\% & 3.51\% & 6.77\% & 0.0 & 0.07 & 0.05 & 0.03 \\
    & L1 (0.01) & 0.00\% & 0.00\% & 0.00\% & 0.00\% & 0.21\% & 0.50\% & 2.24\% & 3.02\% & 0.15 & 0.13 & 0.11 & 0.10 \\
    & L2 (0.1) &  0.00\% & 0.21\% & 0.70\% & 0.70\% & 0.55\% & 1.50\% & 2.15\% & 3.10\% & 0.09 & 0.07 & 0.05 & 0.04 \\
    CIFAR10 & MMD+Mixup & 0.00\% & 0.56\% & 0.44\% & 1.30\% & 1.90\% & 3.04\% & 3.36\% & 3.15\% & 0.19 & 0.18 & 0.16 & 0.15 \\
    & DP-SGD & 0.00\% & 0.00\% & 0.00\% & 0.00\% & 0.05\% & 0.04\% & 0.04\% & 0.09\% & 0.49 & 0.48 & 0.46 & 0.46 \\
    & MemGuard & 0.00\% & 0.00\% & 0.00\% & 0.00\% & 0.00\% & 0.0\% & 0.00\% & 0.00\% & 0.37 & 0.38 & 0.37 & 0.37 \\
    & Max conf. & 0.08\% & 0.10\% & 0.0\% & 0.0\% & 0.78\% & 0.32\% & 0.24\% & 0.18\% & 0.0 & 0.04 & 0.06 & 0.07 \\
    & Max agreed conf. & 0.08\% & 0.10\% & 0.00\% & 0.00\% & 0.78\% & 0.32\% & 0.22\% & 0.19\% & 0.0 & 0.04 & 0.06 & 0.06 \\
    & First agreed conf. & 0.08\% & 0.08\% & 0.00\% & 0.00\% & 0.78\% & 0.58\% & 0.58\% & 0.29\% & 0.0 & 0.04 & 0.05 & 0.06 \\

    \midrule
    & None (deep ensemble) & 0.10\% & 0.14\% & 0.21\% & 0.21\% & 0.54\% & 1.52\% & 3.67\% & 5.10\% & 0.0 & 0.0 & 0.0 & 0.0 \\
    & Bagging & 0.10\% & 0.12\% & 0.06\% & 0.04\% & 0.54\% & 1.04\% & 2.02\% & 4.52\% & 0.0 & 0.38 & 0.33 & 0.31 \\
    & Partitioning & 0.10\% & 0.08\% & 0.20\% & 0.20\% & 0.54\% & 1.21\% & 5.76\% & 5.84\% & 0.0 & 0.17 & 0.13 & 0.10 \\
    & L1 (0.01) & 0.00\% & 0.00\% & 0.00\% & 0.00\% & 0.08\% & 0.08\% & 0.08\% & 0.08\% & 0.95 & 0.95 & 0.93 & 0.93 \\
    & L2 (0.1) & 0.04\% & 0.04\% & 0.54\% & 0.52\% & 0.43\% & 1.43\% & 2.64\% & 4.47\% & 0.23 & 0.21 & 0.17 & 0.15 \\
    CIFAR100 & MMD+Mixup & 0.13\% & 0.10\% & 0.10\% & 0.09\% & 2.21\% & 4.47\% & 7.02\% & 6.97\% & 0.28 & 0.26 & 0.23 & 0.21 \\
    & DP-SGD & 0.00\% & 0.00\% & 0.00\% & 0.00\% & 0.10\% & 0.08\% & 0.15\% & 0.12\% & 0.87 & 0.85 & 0.83 & 0.82\\
    & MemGuard & 0.00\% & 0.00\% & 0.00\% & 0.00\% & 0.00\% & 0.00\% & 0.00\% & 0.00\% & 0.40 & 0.39 & 0.38 & 0.38 \\
    & Max conf. & 0.10\% & 0.02\% & 0.00\% & 0.00\% & 0.54\% & 0.23\% & 0.15\% & 0.09\% & 0.0 & 0.10 & 0.15 & 0.17 \\
    & Max agreed conf. & 0.10\% & 0.02\% & 0.00\% & 0.03\% & 0.54\% & 0.23\% & 0.15\% & 0.13\% & 0.0 & 0.10 & 0.15 & 0.16 \\
    & First agreed conf. & 0.10\% & 0.01\% & 0.00\% & 0.01\% & 0.54\% & 0.25\% & 0.11\% & 0.20\% & 0.0 & 0.10 & 0.13 & 0.14 \\

    \bottomrule
  \end{tabular}
\end{table*}

\textbf{Sampling attack:}
The sampling attack is different from confidence-based membership inference attacks because it does not follow the accuracy-privacy trade-off, as shown in the last row of Figure~\ref{fig-trade-off}. In fact, the effectiveness of the sampling attack decreases in most ensemble learning approaches as more base models are added to the ensemble. As shown in \cite{rahimian2020sampling}, the most effective defense for sampling attack is DP-SGD. However, as the number of base models increases in a deep ensemble, the performance of sampling attack degrades to a point where it is often worse than the trivial gap attack. Since the gap attack is a shown to be ineffective in practice \cite{rezaei2020towards}, we consider any ensembling approach itself as a defense mechanism for the sampling attack. Hence, a deep ensemble with maximum confidence can effectively improve both accuracy and privacy.



Although AUC has been overwhelmingly used to report the performance of MI attacks, \cite{carlini2021membership} first argued that a more reliable metric is true positive at low false positive rate. In Table \ref{tbl-pf} we present this metric for Watson attack. We do not report Shokri and sampling attacks here because their true positive was almost zero for low false positive rate, as also shown in \cite{carlini2021membership}.

As mentioned earlier, confidence-masking defenses heavily distort the confidence output of models causing an issue for applications that rely on real confidence values, such as uncertainty estimation. In Table \ref{tbl-pf}, we present how much distortion each defence mechanism imposes to the original confidence values. Here, for each data sample, we compute the L1 distance between the original deep ensemble confidence values and the defence mechanism's confidence output. Then, we normalize the values to be between 0 and 1 and then take the average. As shown in Table \ref{tbl-pf}, MMD+Mixup, DP-SGD, and MemGuard are heavily distorting the confidence outputs while our proposed approaches impose smaller distortion.


\section{Discussion}

We note some limitations of our empirical analysis and opportunities for future work. First, privacy is a multi-faceted concept and can be defined or quantified in several ways. In this work, we quantified privacy leakage in terms of the effectiveness of membership inference attacks. One can quantify privacy in terms of other relevant attacks such as model inversion \cite{fredrikson2015model, he2019model}, property inference \cite{ateniese2015hacking, ganju2018property}, and model stealing attack \cite{tramer2016stealing} as well as formally-provable measures such as differential privacy \cite{dwork2008differential}.

Second, ensemble learning is an umbrella term covering a wide variety of methods to combine multiple base learners. Although the ensemble learning is the most widely-used approach in deep learning, an arbitrary method of training and combining base learners can still be construed as ensemble learning while improving privacy. We mainly focused to deep ensembles because of its prevalent use for ensembling deep models. Moreover, we conduct experiment over several other ensembling approaches and the conclusion remain the same. A more exhaustive experimental evaluation may discover new results which is out of scope of this paper.

Third, this paper focuses on black-box attack scenario and our solution to accuracy-privacy trade-off in deep ensembles relies on changing the fusing part of ensemble learning. In other words, base models in the ensemble are intact. Consequently, if all base models are publicly available in a white-box setting, an attacker can still average the outputs and bypass the maximum agree confidence mechanism. 
One solution can focus on training models sequentially (boosting) and applying some non-trivial criteria during training of each model to force the distribution of correct agreement to be close for train and non-train samples. How to achieve this is not trivial and needs further research.

\section{Conclusion}
In this paper, we investigate membership inference attacks in deep ensemble learning and demonstrate that there exists a trade-off between accuracy and privacy. 
We show that the most influential factor that causes more effective membership inferences attack against deep ensembles is the level of agreement between base models. We illustrate the effect of several classical regularization techniques, including L1/L2 regularization and DP-SGD, to mitigate membership inference attacks and conclude that none of them can break the trade-off and improve the accuracy and privacy, simultaneously. Finally, we propose a simple yet highly effective solution that only changes ensemble's fusion post-training.

\section*{Acknowledgment}
The work was partially supported by NSF through grants  USDA-020-67021-32855, IIS-1838207, CNS 1901218, and OIA-2134901.


\bibliographystyle{IEEEtran}


\appendices

\section{}

\begin{figure}
\centering
\includegraphics[width = 0.7\linewidth]{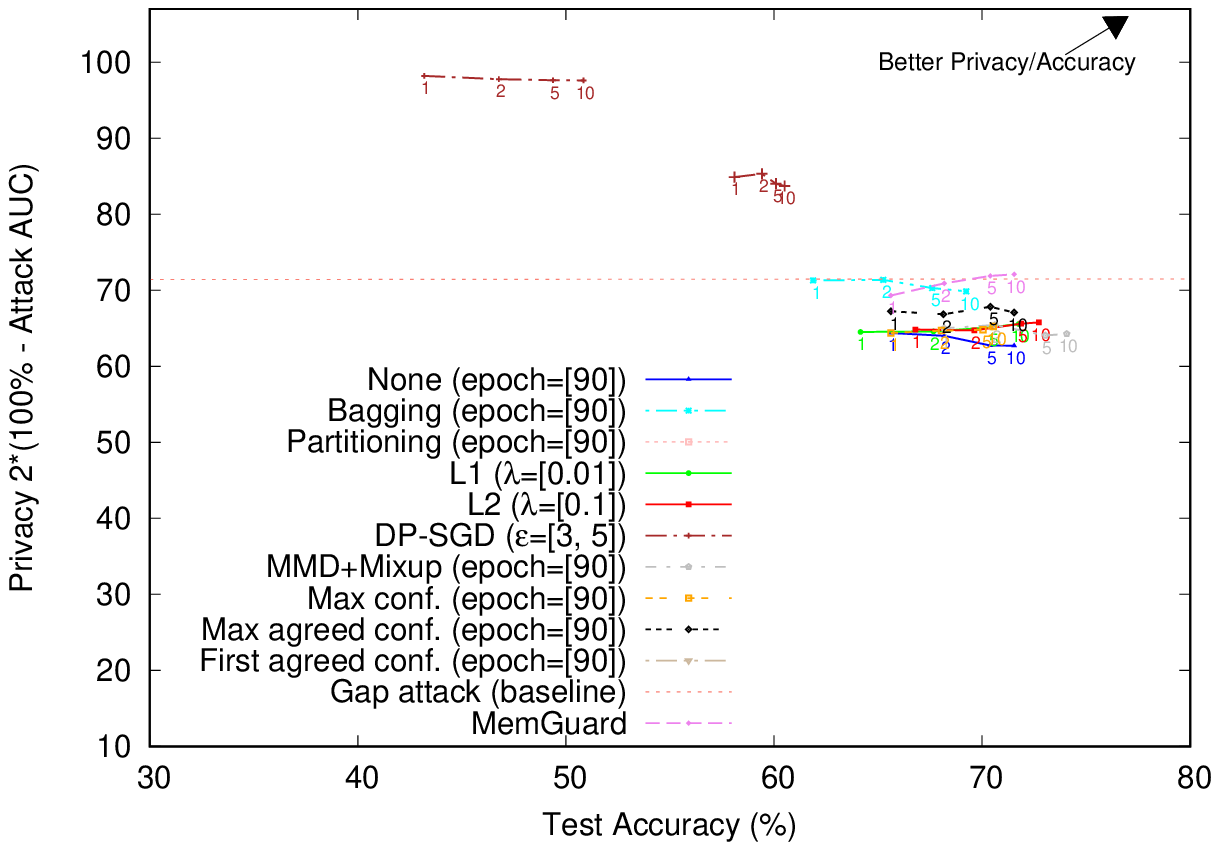}
\caption{label only attack on CIFAR10-AlexNet model.}
\label{fig-all-label-only1}
\vspace{-.1in}
\end{figure}

\begin{figure}
\centering
\includegraphics[width = 0.7\linewidth]{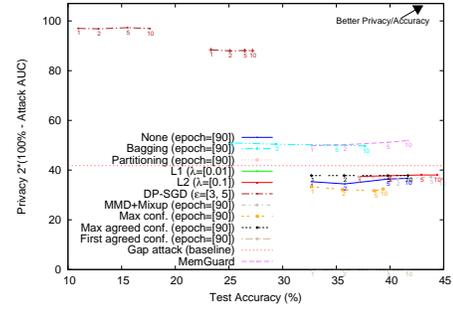}
\caption{label only attack on CIFAR100-AlexNet model.}
\label{fig-all-label-only2}
\vspace{-.1in}
\end{figure}

\subsection{Label Only Attack}
\label{appendix-label-only}
In this section, we analyze the effect of label-only MI attack proposed in \cite{choquette2021label}. The idea is to use an adversarial example generator to find the distance to the decision boundary as membership inference metric. They use “HopSkipJump” \cite{chen2020hopskipjumpattack} from Cleverhans\footnote{https://github.com/cleverhans-lab/cleverhans} to craft adversarial examples. We use the same algorithm, implementation and hyper-parameters. The results are shown in Figure \ref{fig-all-label-only1} and \ref{fig-all-label-only2}. We have not seen significant difference in terms of MI attack AUC when ensembling is used. However, we observe that by increasing the number of models in an ensemble, the HopSkipJump used in label only attack becomes less effective in finding adversarial samples in the specified number of iterations. Due to the limited time and computational budget, we have not explored stronger approaches. Using more iterations and a stronger adversarial attack may leads to a different result.

\subsection{Defense Mechanism}
\label{appendix-all-defense}
\begin{figure*}
\def\tabularxcolumn#1{m{#1}}
\begin{tabularx}{\linewidth}{@{}cXX@{}}
\begin{tabular}{ccc}
\subfloat[CIFAR10-ResNet20: Shokri's attack]
{\includegraphics[width=0.30\linewidth]{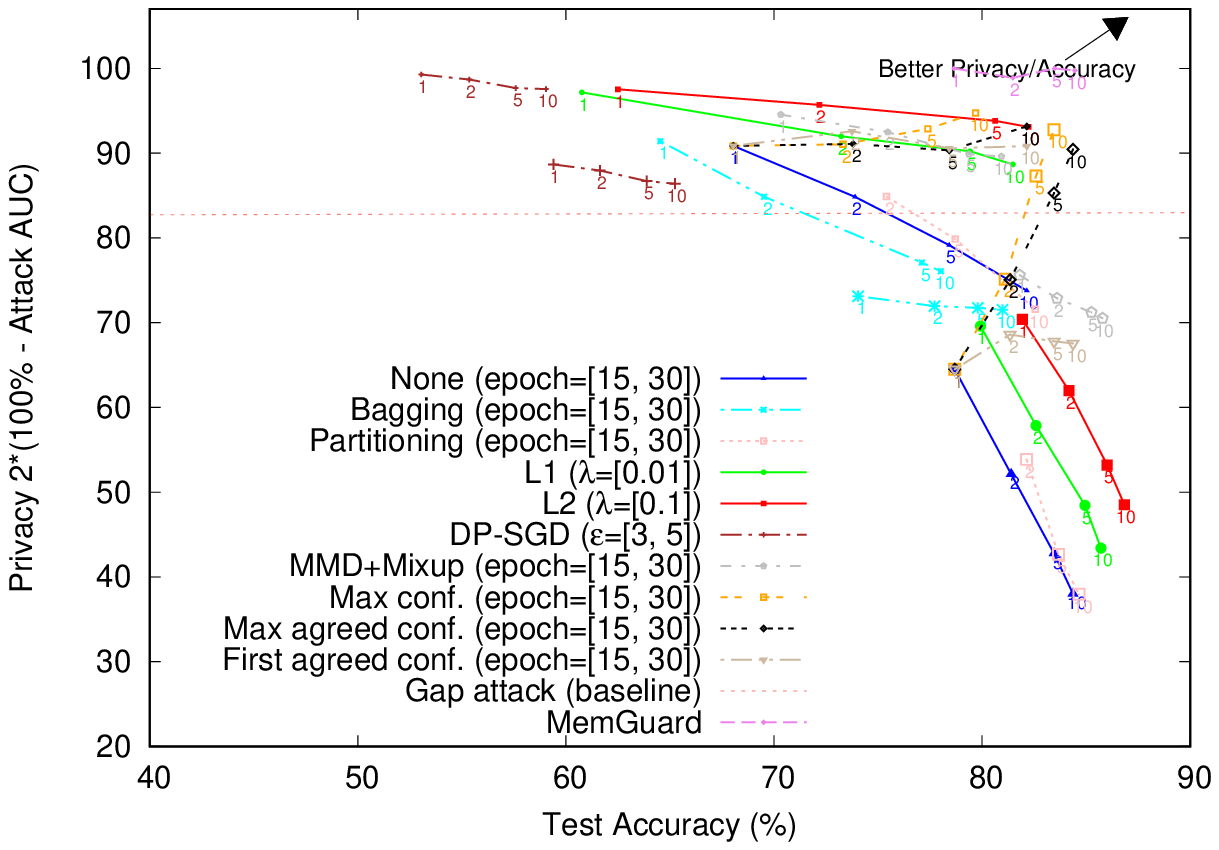}} 
  & 
\subfloat[CIFAR10-ResNet20: Watson's MI attack]
{\includegraphics[width=0.30\linewidth]{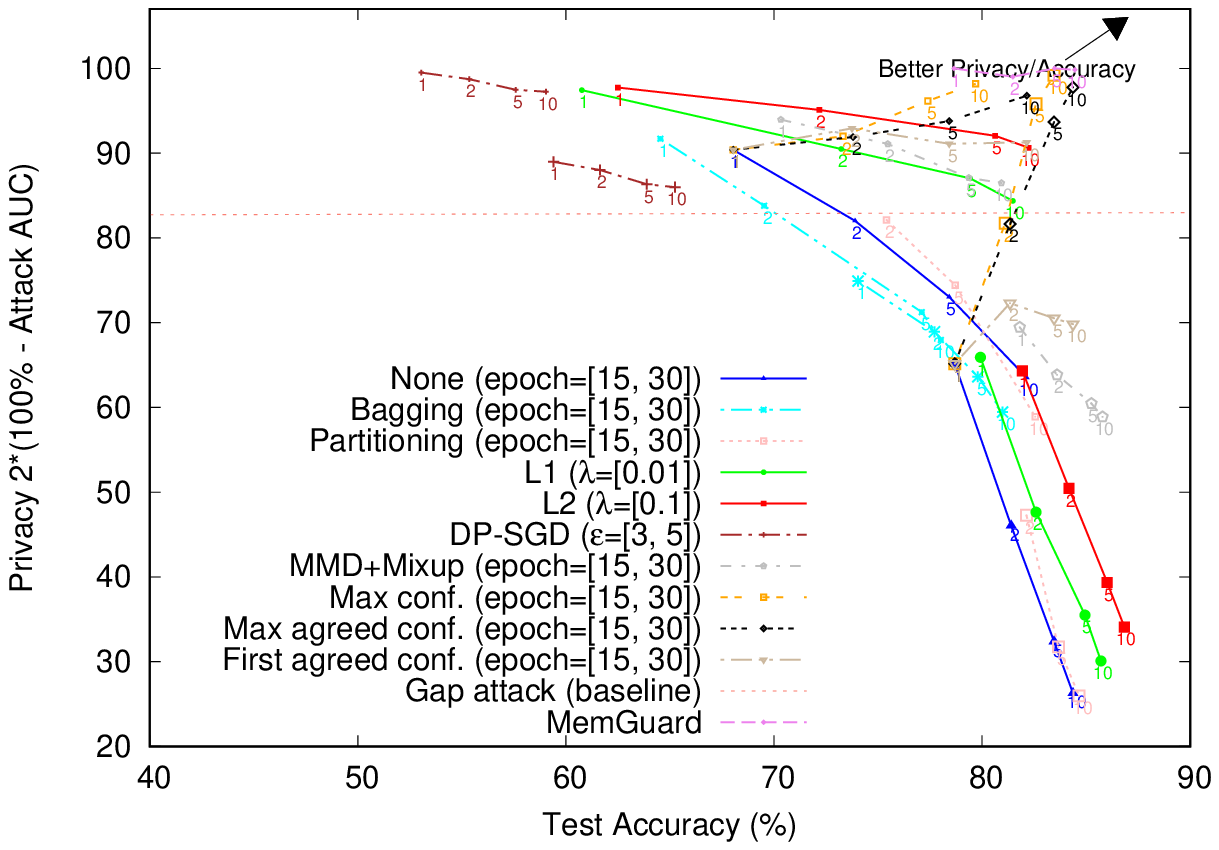}}
  &
\subfloat[CIFAR10-ResNet20: Sampling MI attack]
{\includegraphics[width=0.30\linewidth]{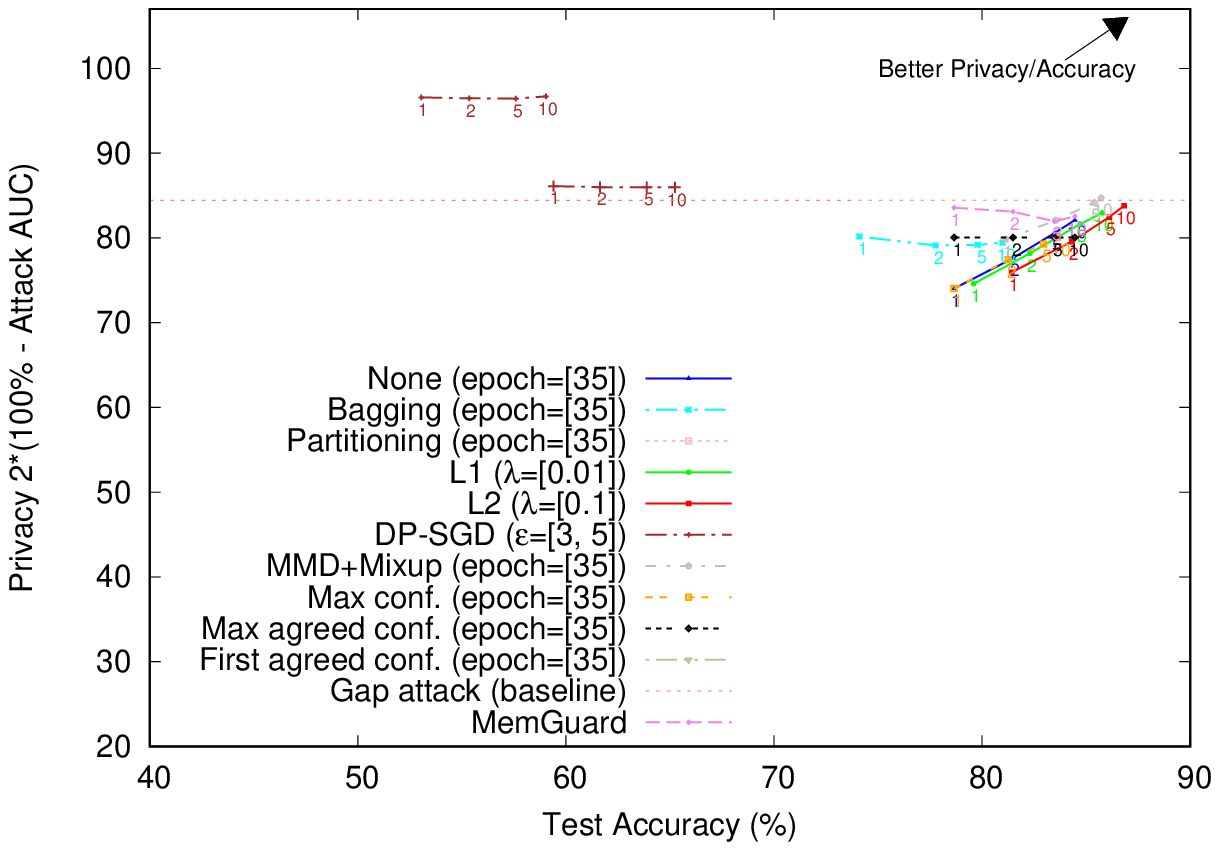}}\\

\subfloat[CIFAR100-ResNet20: Shokri's attack]
{\includegraphics[width=0.30\linewidth]{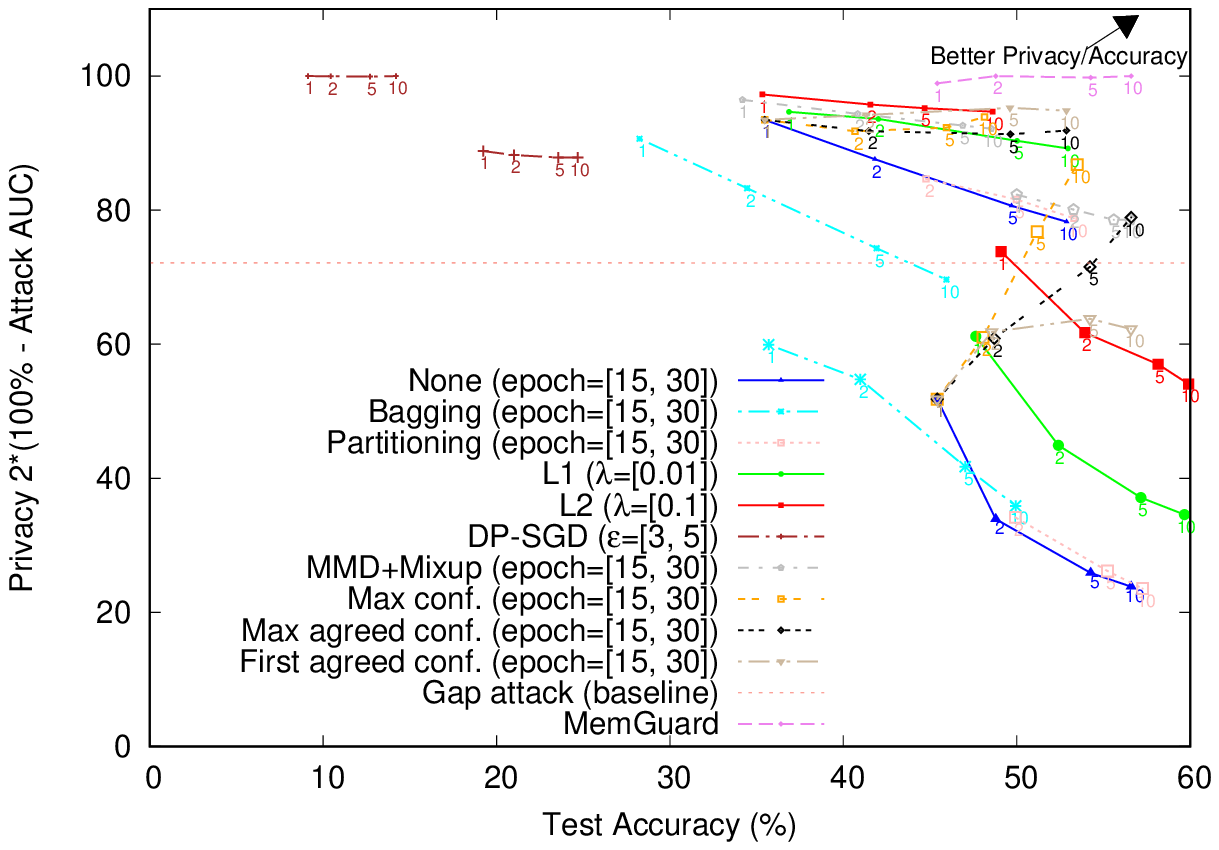}} 
  & 
\subfloat[CIFAR100-ResNet20: Watson's MI attack]
{\includegraphics[width=0.30\linewidth]{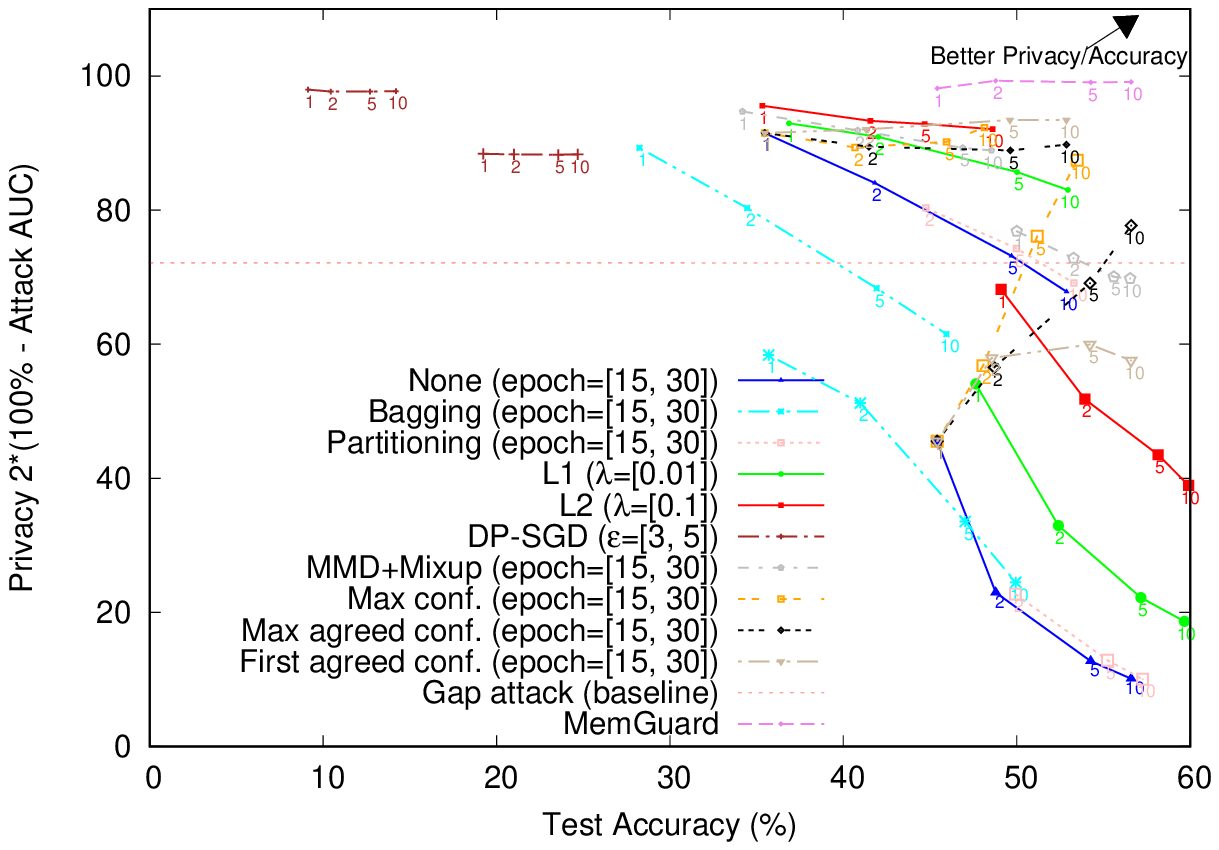}}
  & 
\subfloat[CIFAR100-ResNet20: Sampling MI attack]
{\includegraphics[width=0.30\linewidth]{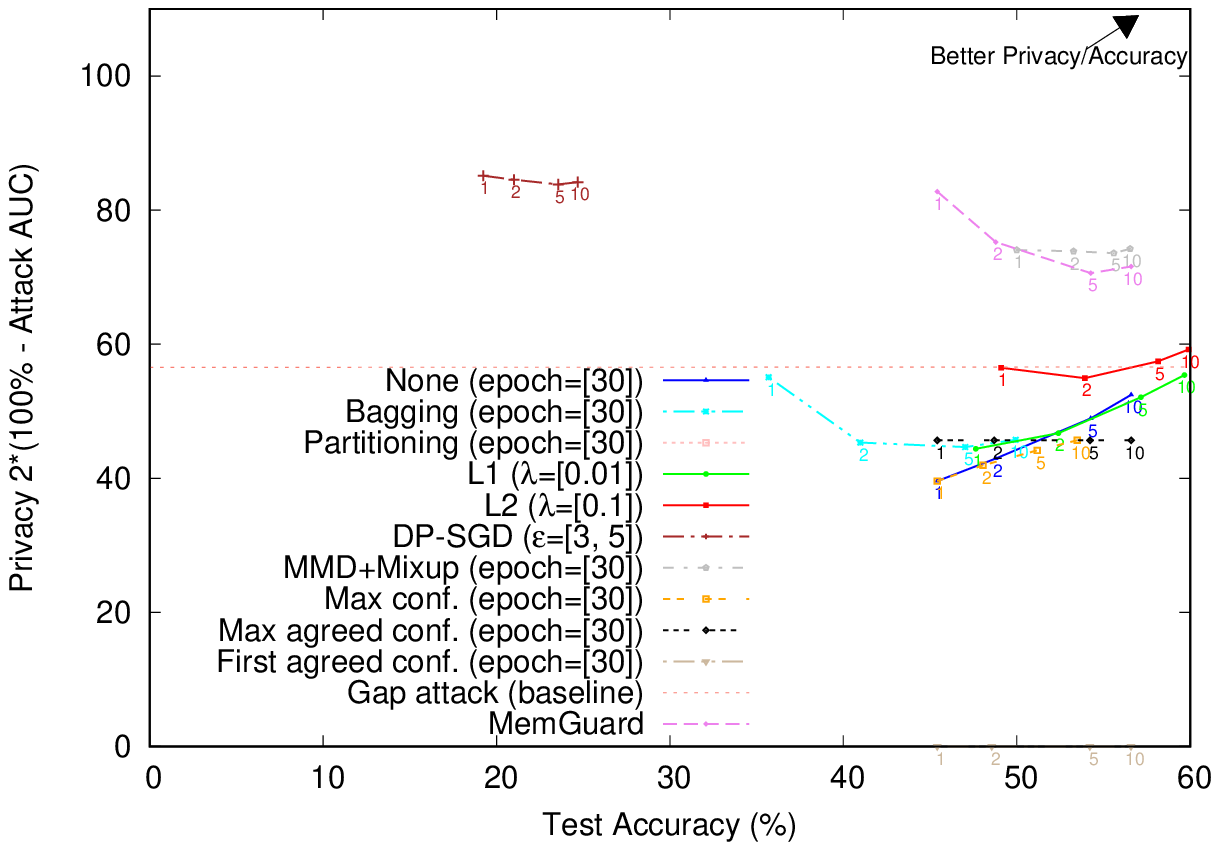}}\\

\subfloat[SVHN-AlexNet: Shokri's MI attack]
{\includegraphics[width=0.30\linewidth]{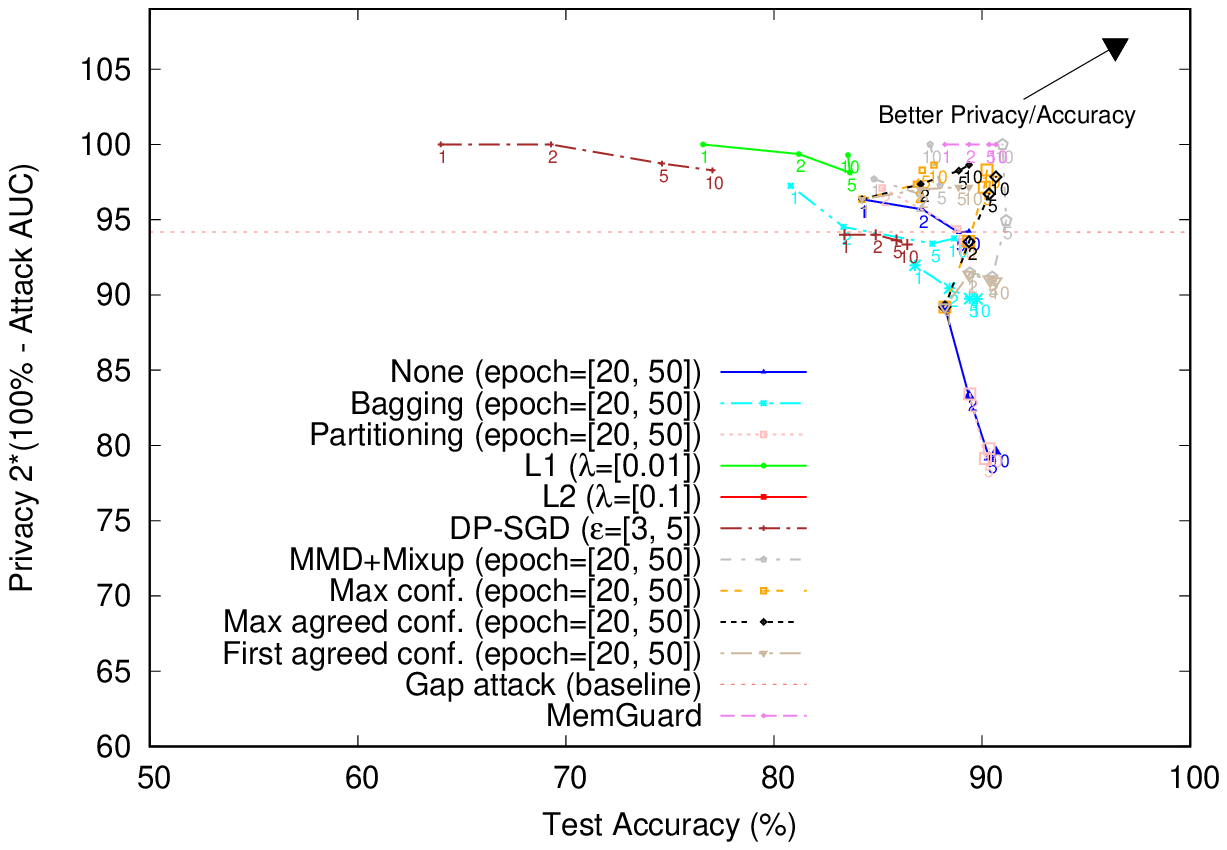}} 
  & 
\subfloat[SVHN-AlexNet: Watson's MI attack]
{\includegraphics[width=0.30\linewidth]{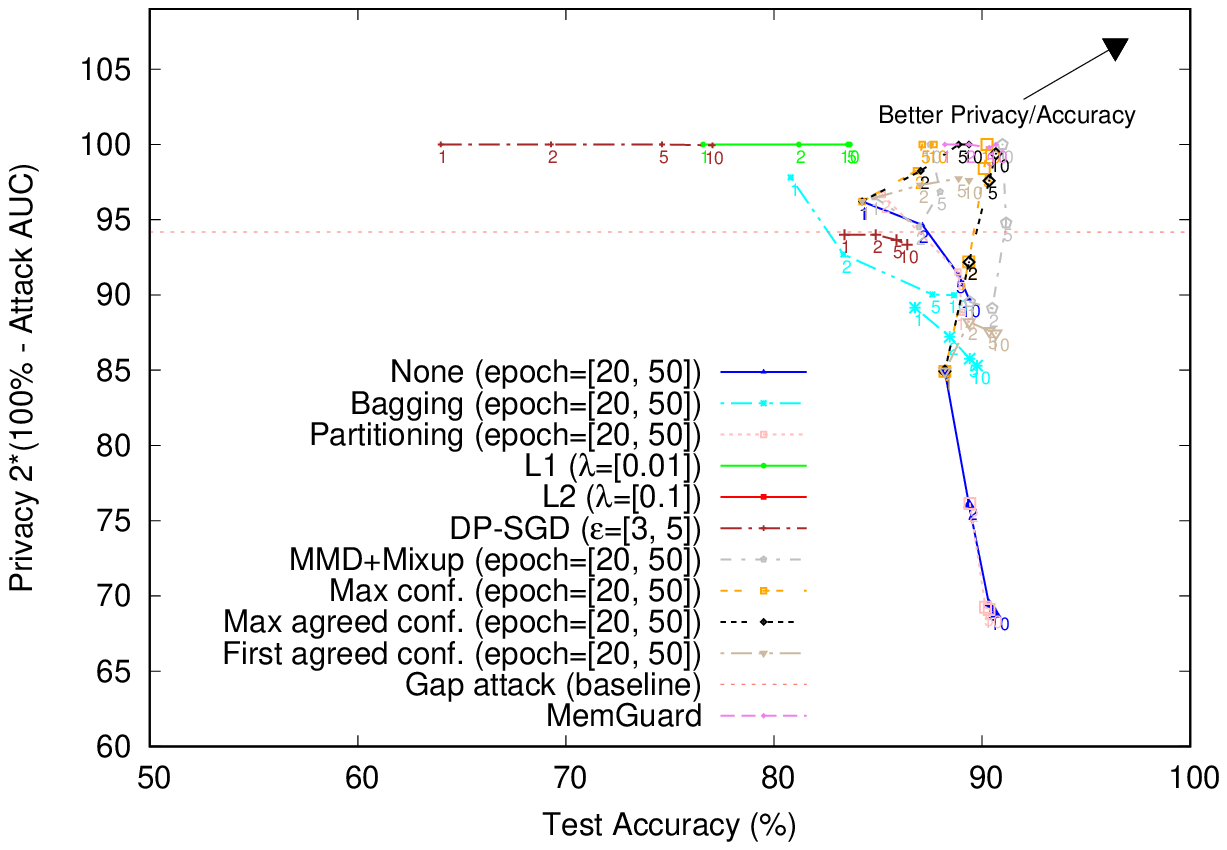}}
  &
\subfloat[SVHN-AlexNet: Sampling MI attack]
{\includegraphics[width=0.30\linewidth]{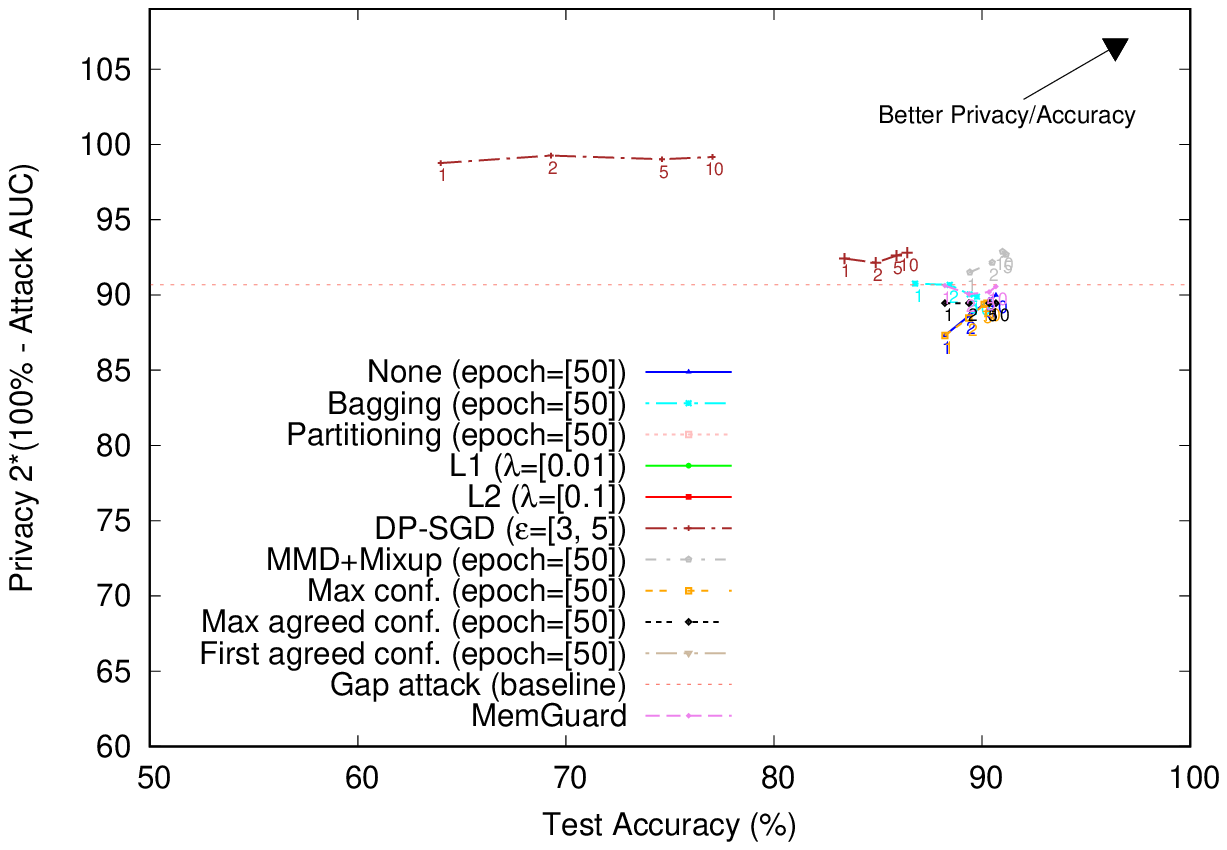}}\\

\subfloat[SVHN-ResNet20: Shokri's attack]
{\includegraphics[width=0.30\linewidth]{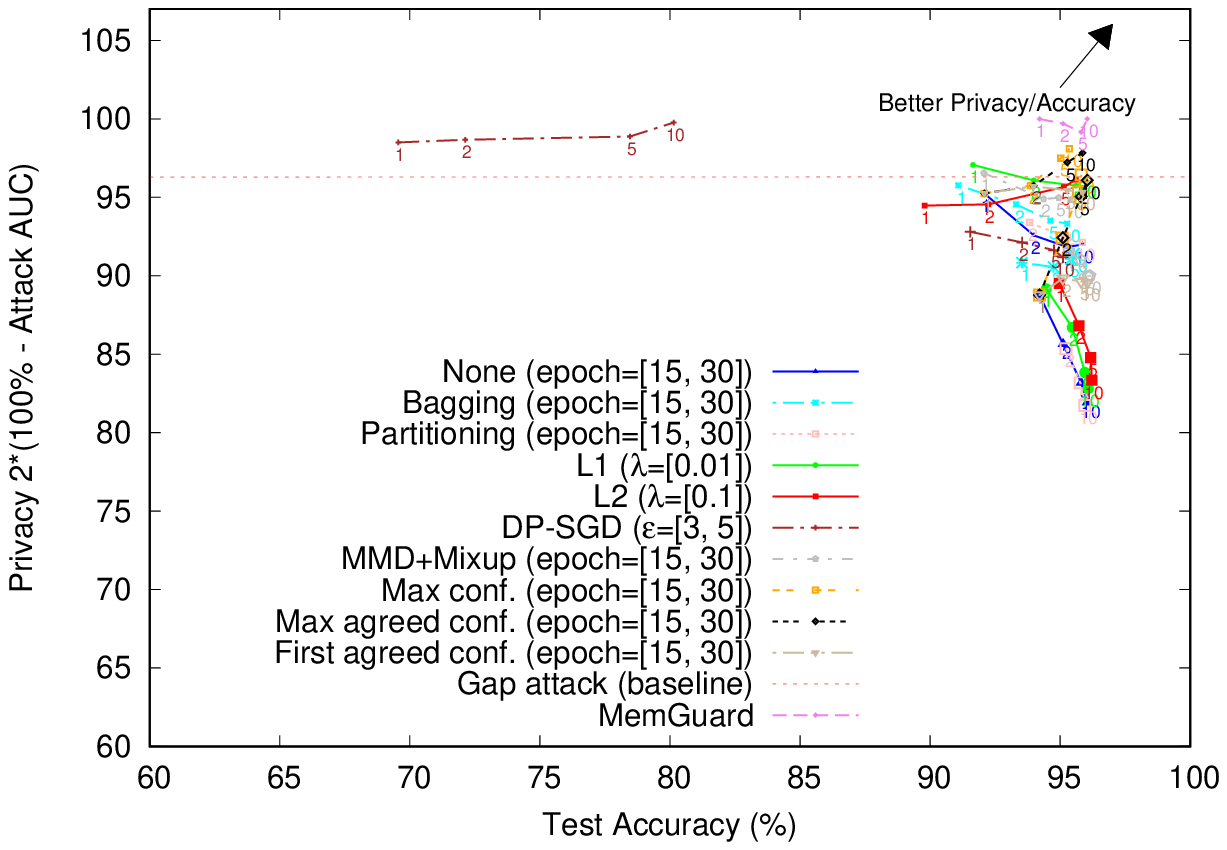}} 
  & 
\subfloat[SVHN-ResNet20: Watson's MI attack]
{\includegraphics[width=0.30\linewidth]{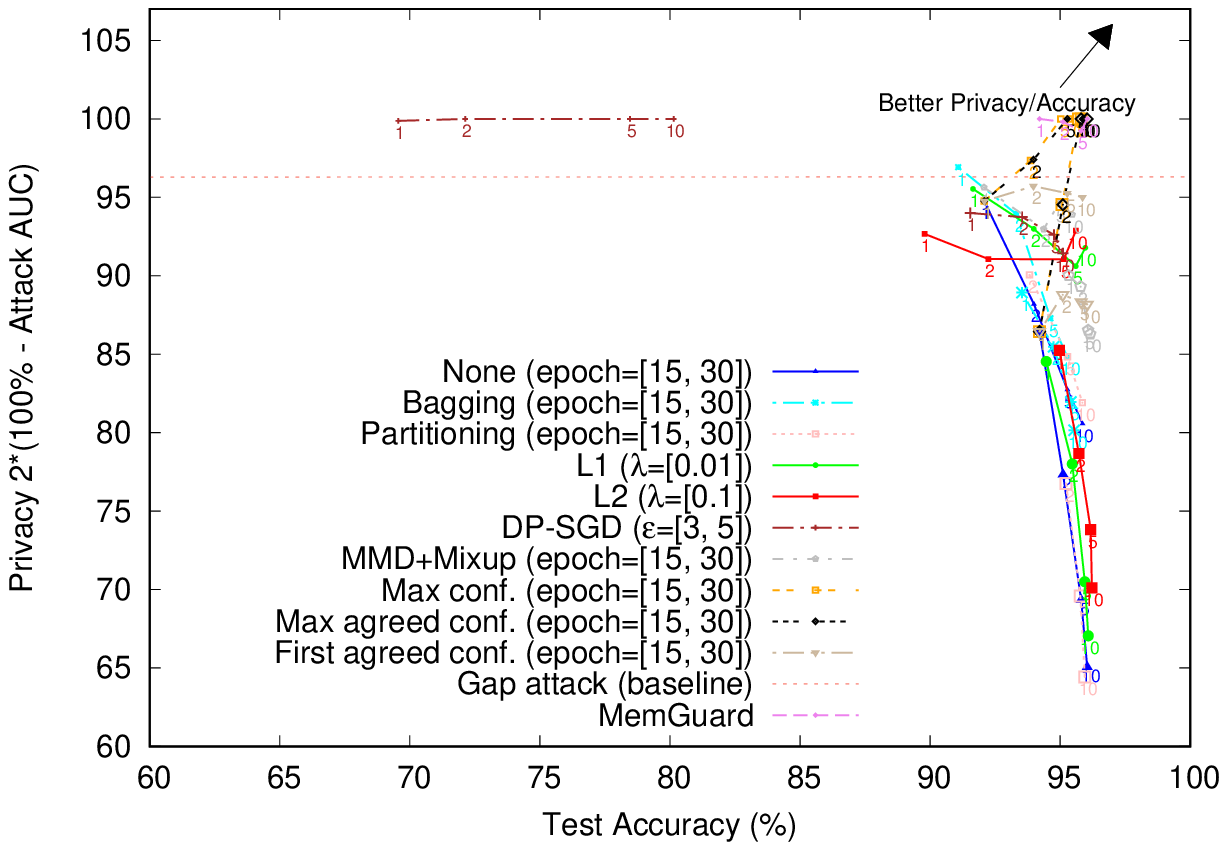}}
  & 
\subfloat[SVHN-ResNet20: Sampling MI attack]
{\includegraphics[width=0.30\linewidth]{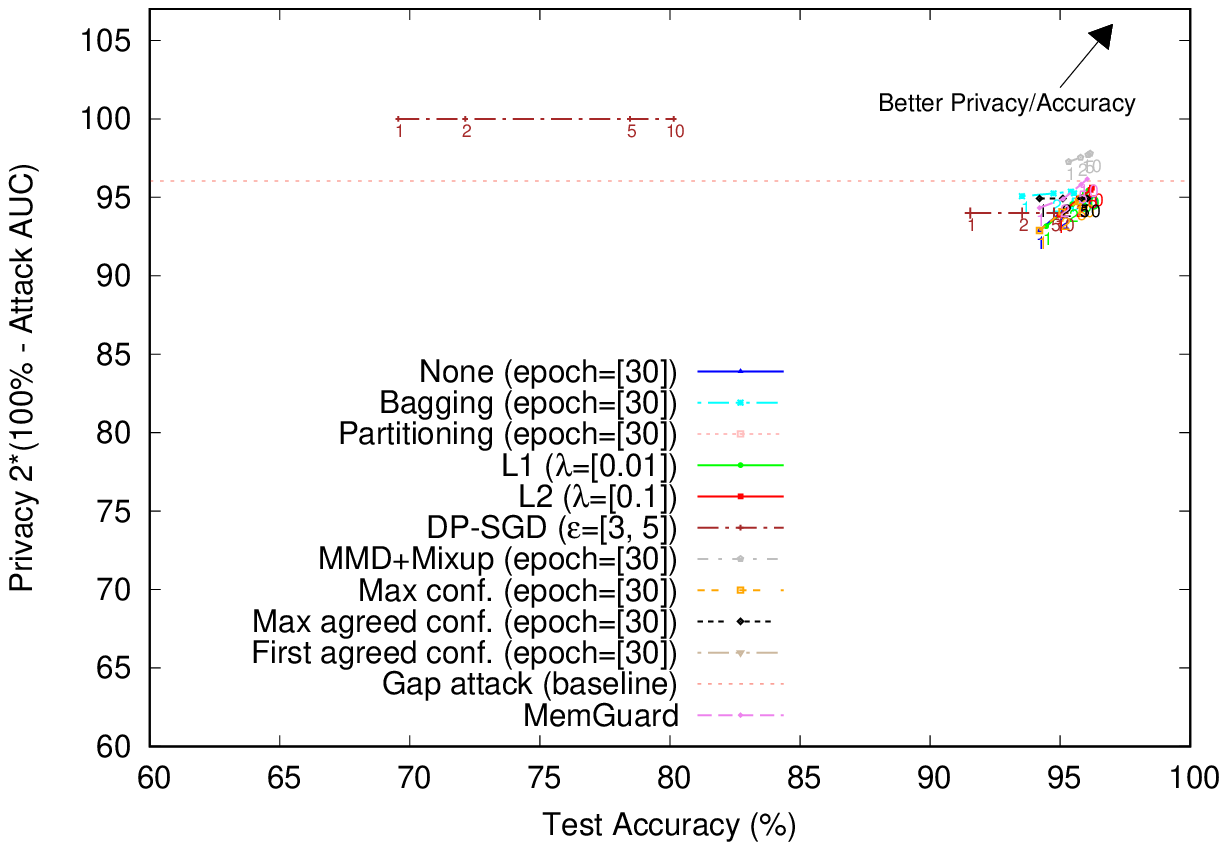}}\\

\end{tabular}
\end{tabularx}
\caption{Effect of defense mechanisms.}\label{fig-all-defenses}
\end{figure*}

The effect of all defense mechanisms are shown in Figure \ref{fig-all-defenses}. Most defense mechanisms become less effective when deep ensemble is used. As shown Section \ref{sec-eval}, maximum agreed confidence and maximum confidence achieve the best accuracy and privacy.

\subsection{More Advanced Ensembling Approaches}
\label{appendix-all-epochs-advanced-ensemble}
\begin{figure*}
\def\tabularxcolumn#1{m{#1}}
\begin{tabularx}{\linewidth}{@{}cXX@{}}
\begin{tabular}{ccc}
\subfloat[CIFAR10 (AlexNet)]{\includegraphics[width=0.29\linewidth]{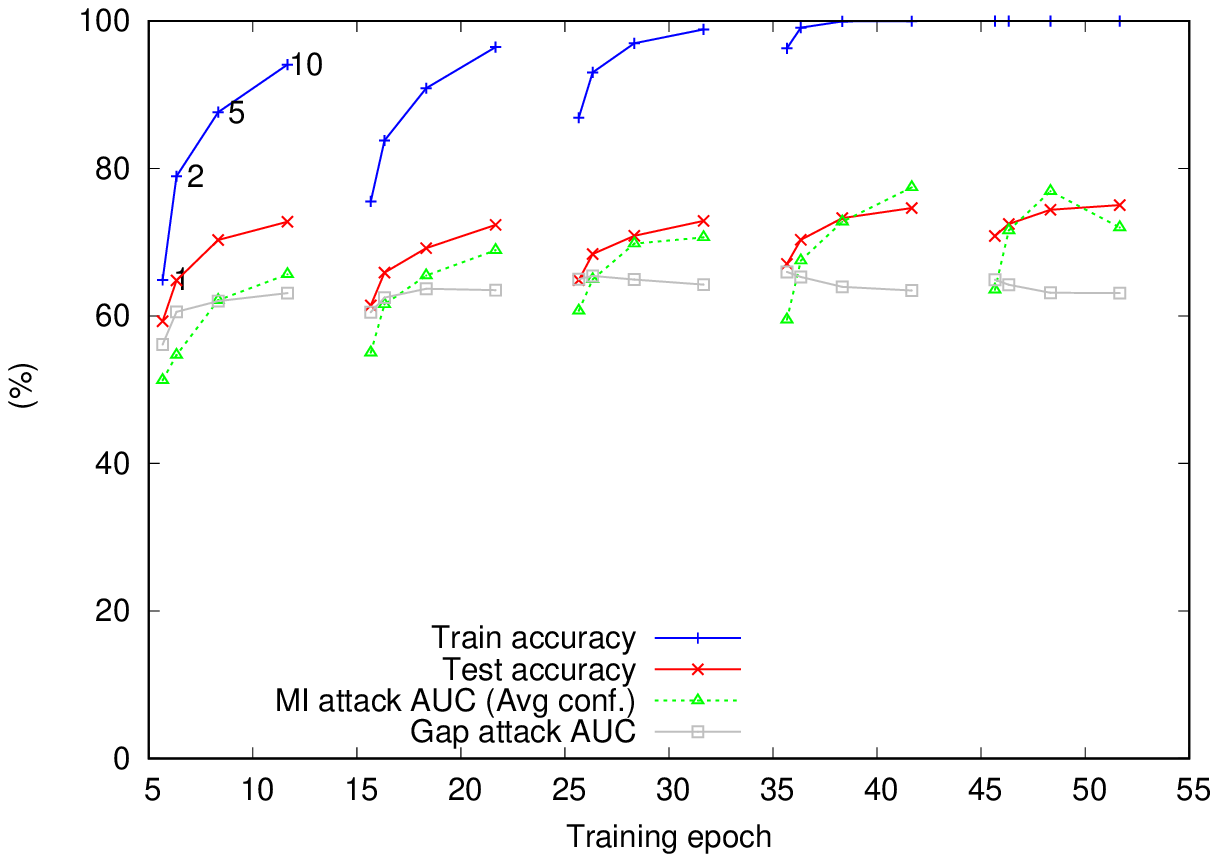}} 
  & \subfloat[CIFAR10 (ResNet20)]{\includegraphics[width=0.29\linewidth]{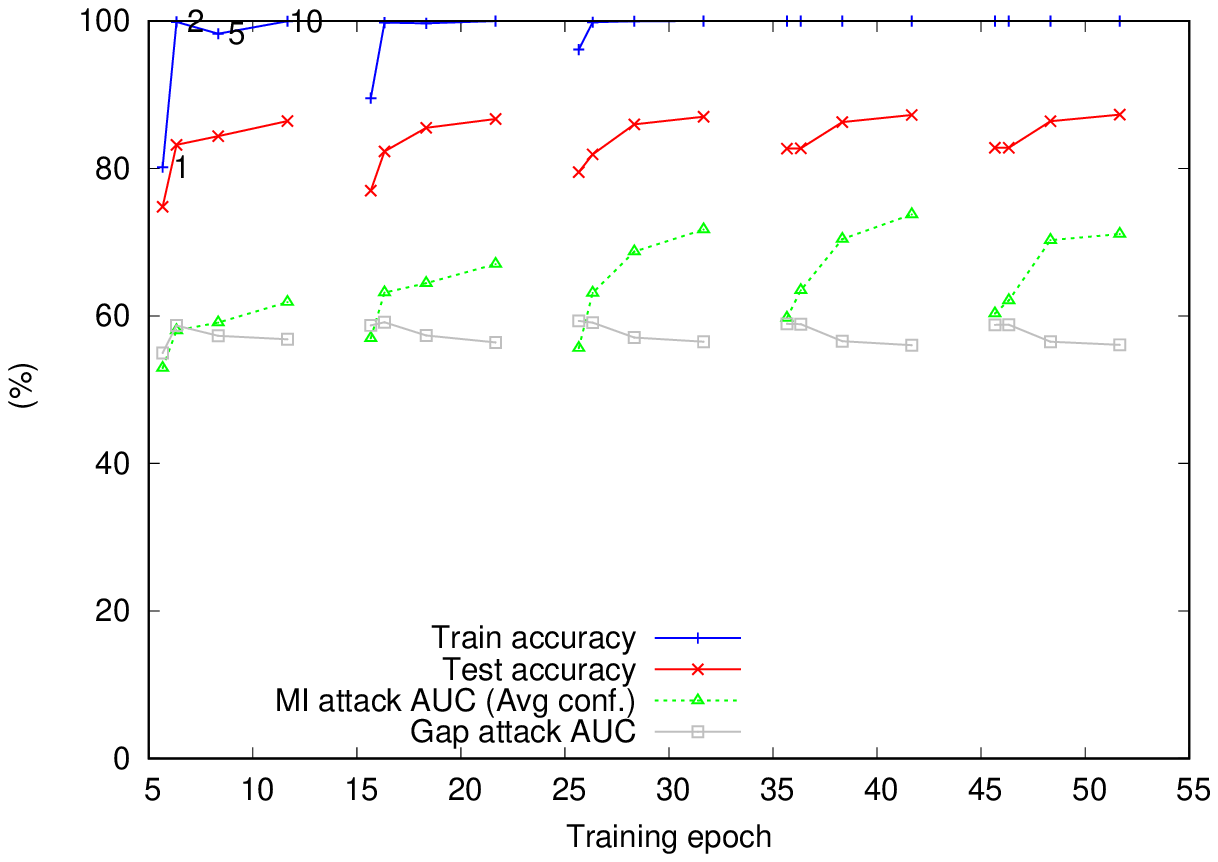}}
  & \subfloat[CIFAR10 (DenseNet100)]{\includegraphics[width=0.29\linewidth]{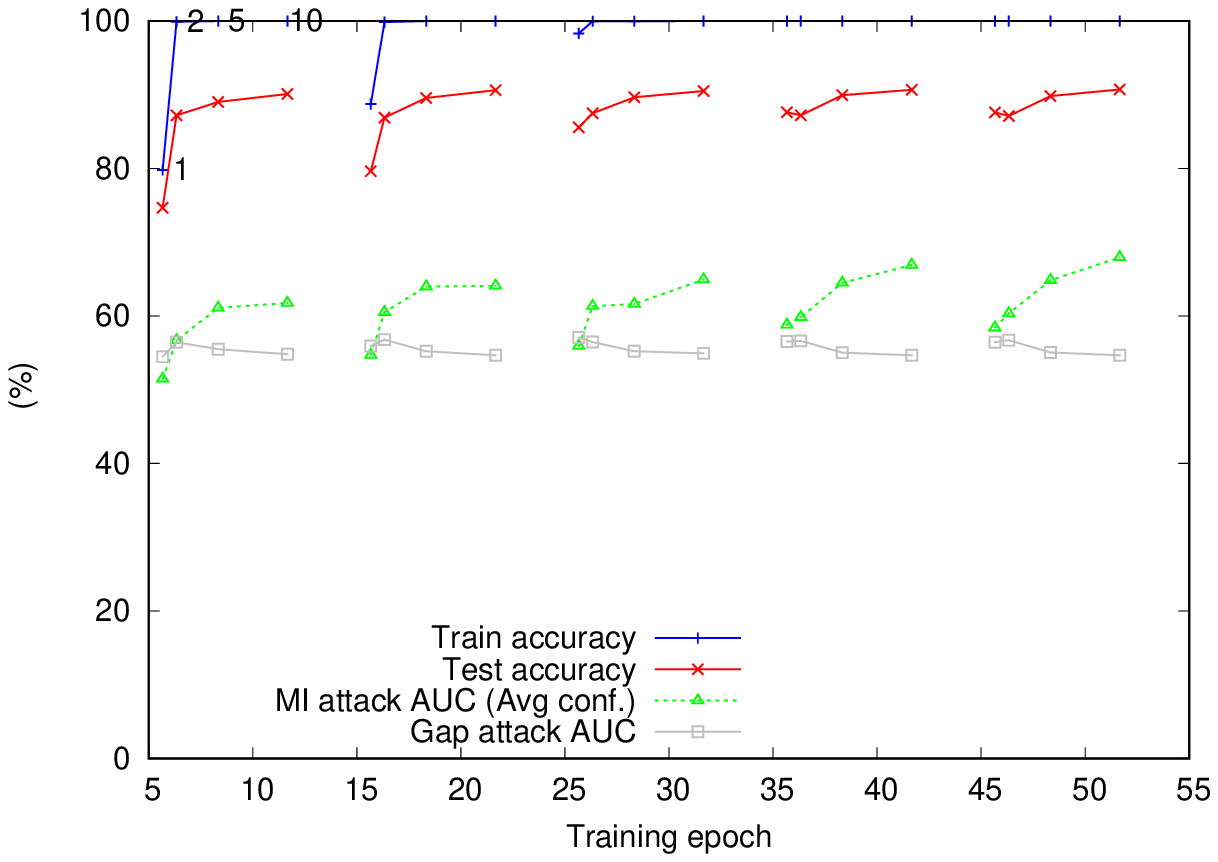}} \\
\subfloat[CIFAR10 (WResNet16-2)]{\includegraphics[width=0.29\linewidth]{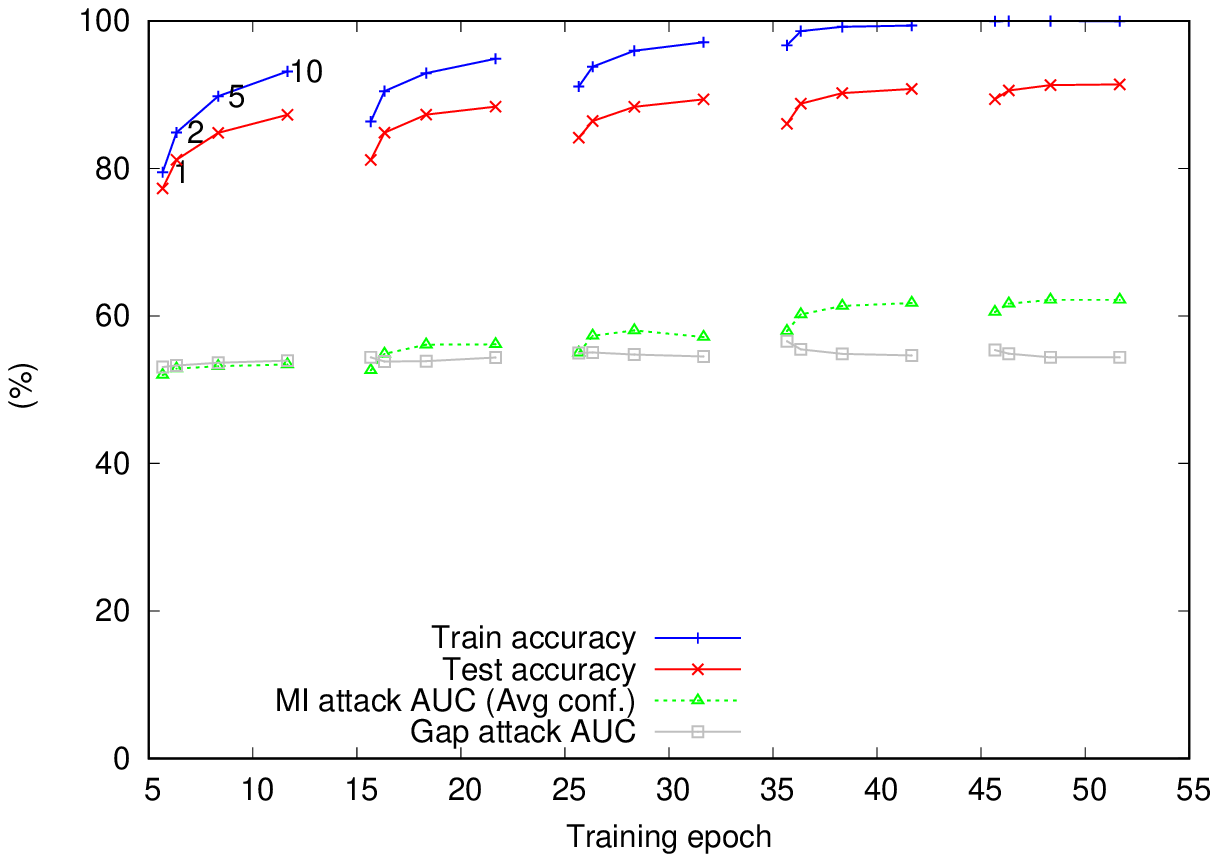}} 
  & \subfloat[CIFAR100 (AlexNet)]{\includegraphics[width=0.29\linewidth]{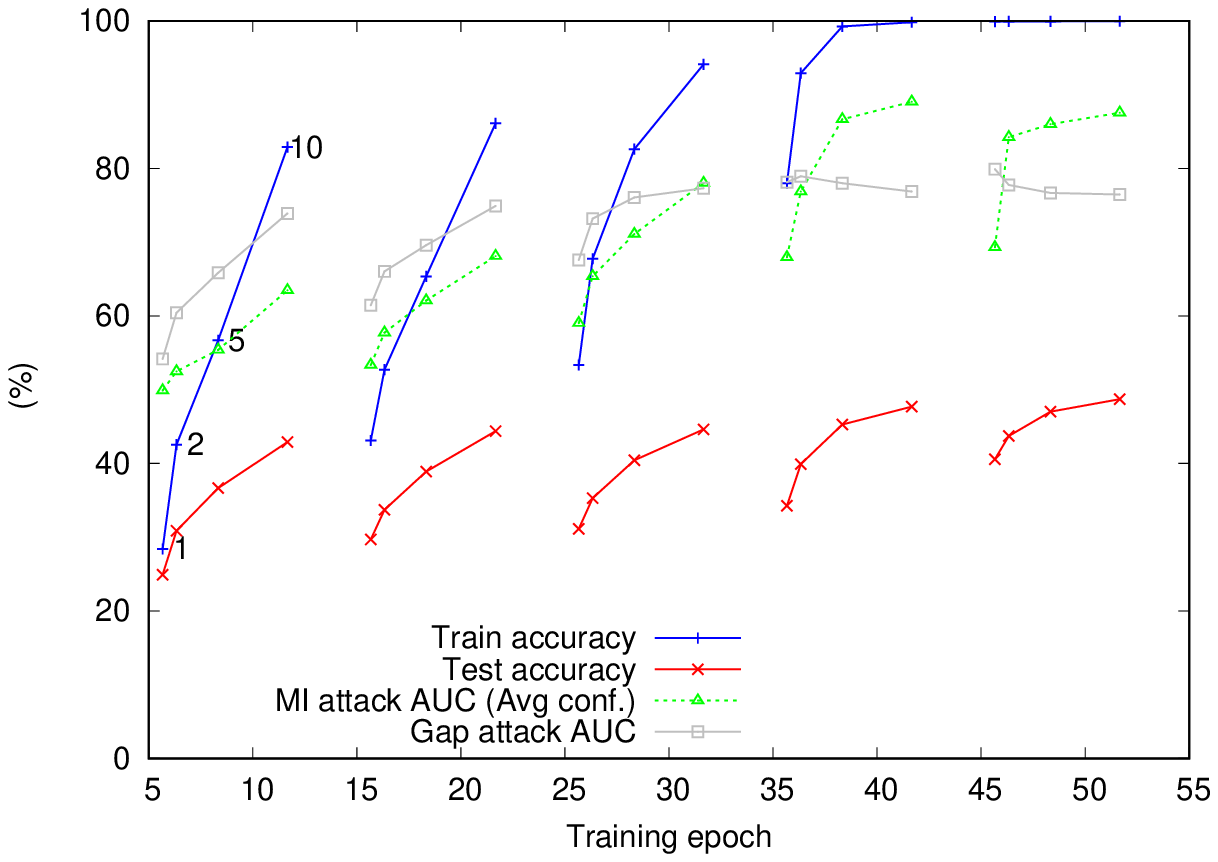}}
  & \subfloat[CIFAR100 (ResNet20)]{\includegraphics[width=0.29\linewidth]{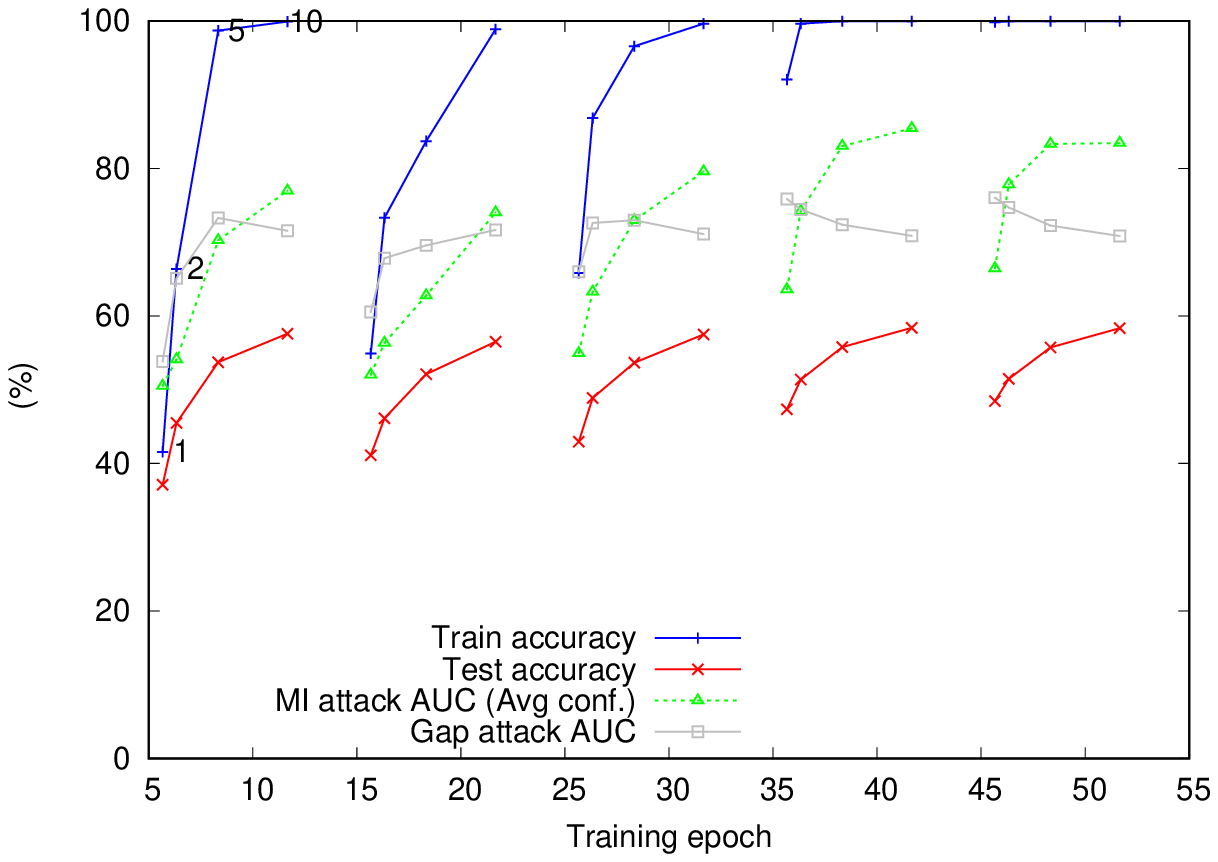}} \\
\subfloat[CIFAR100 (DenseNet100)]{\includegraphics[width=0.29\linewidth]{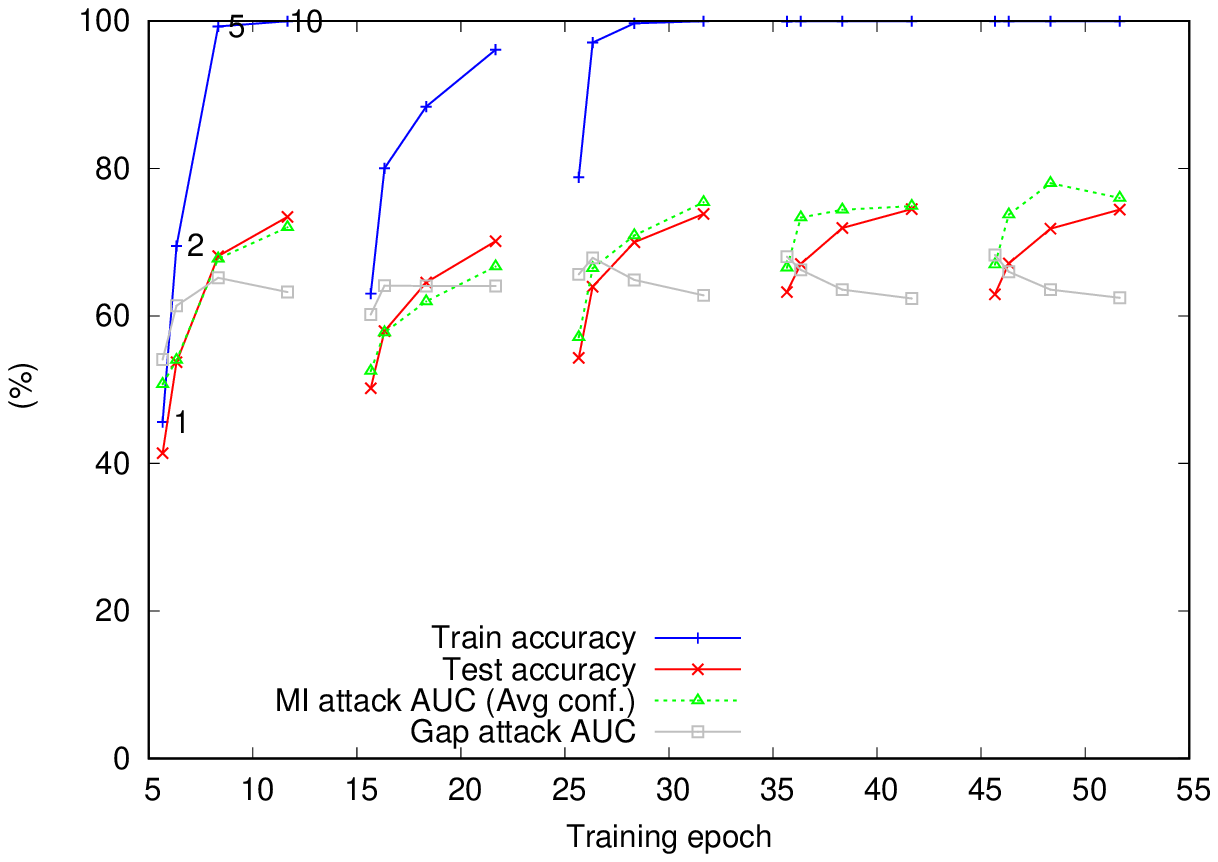}} 
  & \subfloat[CIFAR100 (WResNet16-2)]{\includegraphics[width=0.29\linewidth]{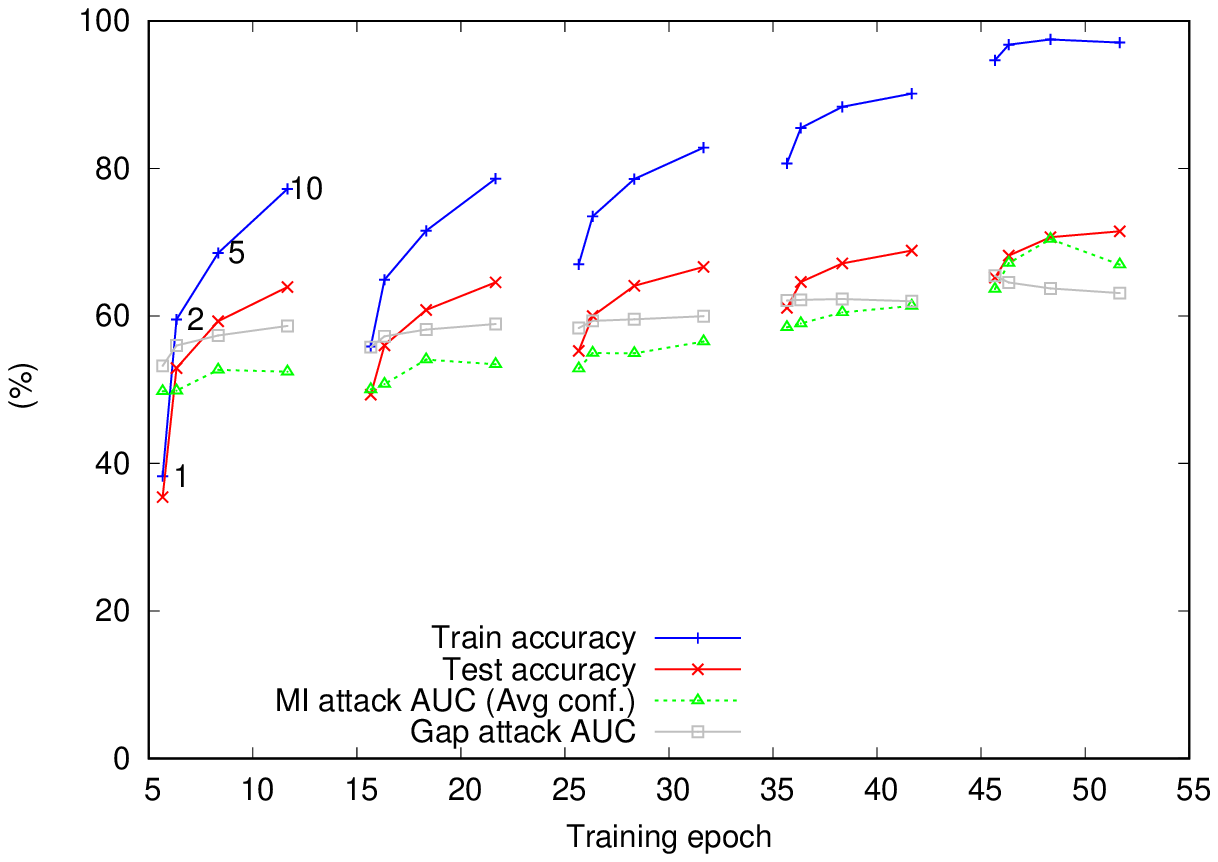}}
  & \subfloat[ImageNet (ResNet50)]{\includegraphics[width=0.29\linewidth]{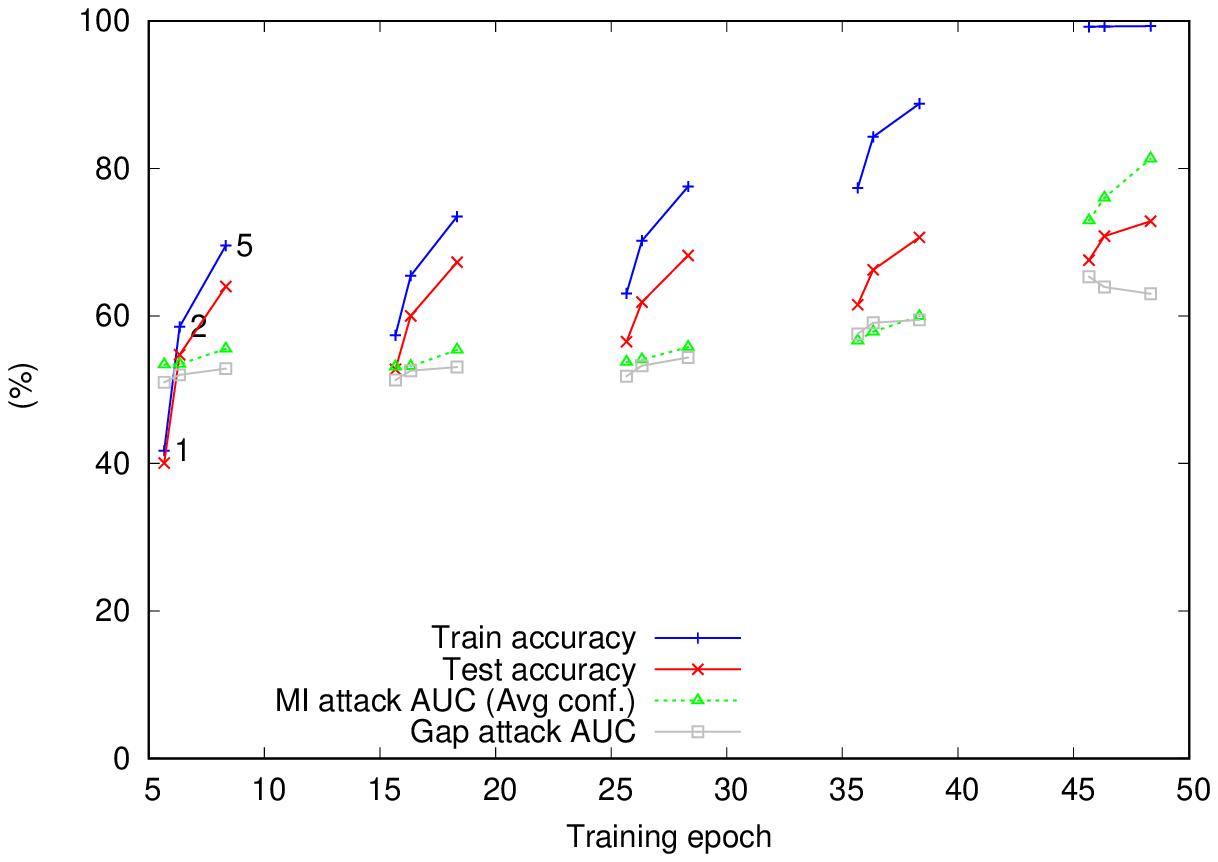}} \\
\end{tabular}
\end{tabularx}
\caption{Target models' accuracy and MI attacks' AUC across all datasets and models using snapshot ensemble \cite{huang2017snapshot}.}
\label{fig-all-epochs-cyclic}
\end{figure*}

In this section, we evaluate two state-of-the-art ensembling approaches, namely snapshot ensembles \cite{huang2017snapshot} and diversified ensemble networks \cite{zhang2020diversified}. 
For snapshot ensemble, we train several models on several datasets for 500 epochs and restart the cycle every 50 epochs, similar to the original paper \cite{huang2017snapshot}. Note that the goal of our evaluation is show the accuracy-privacy trade-off, not to achieve the highest accuracy possible. Due to this reason and limited time we had, we did not perform an exhaustive hyper-parameter tuning. Nevertheless, similar trade-off can be observed in Figure~\ref{fig-all-epochs-cyclic}.

\begin{figure*}
\def\tabularxcolumn#1{m{#1}}
\begin{tabularx}{\linewidth}{@{}cXX@{}}
\begin{tabular}{ccc}
\subfloat[CIFAR10 (AlexNet)]{\includegraphics[width=0.29\linewidth]{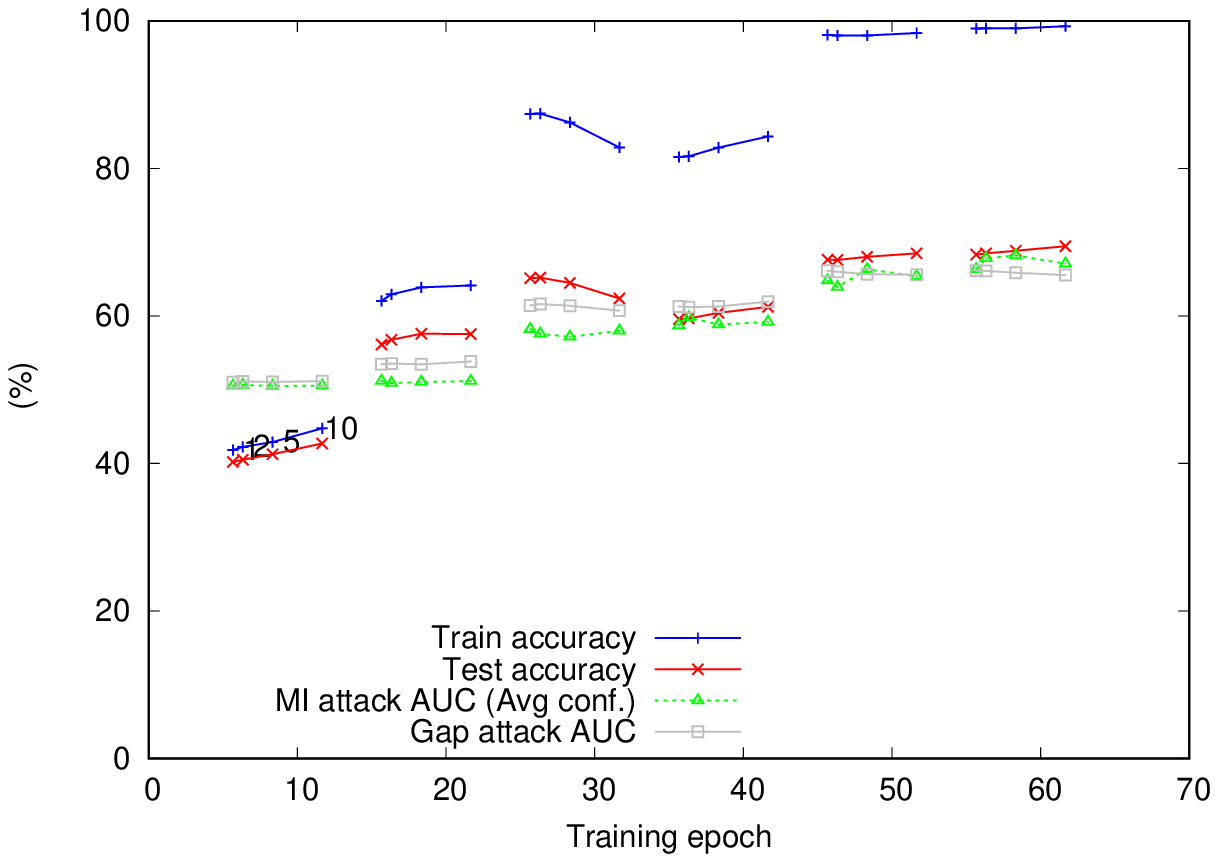}} 
  & \subfloat[CIFAR10 (ResNet20)]{\includegraphics[width=0.29\linewidth]{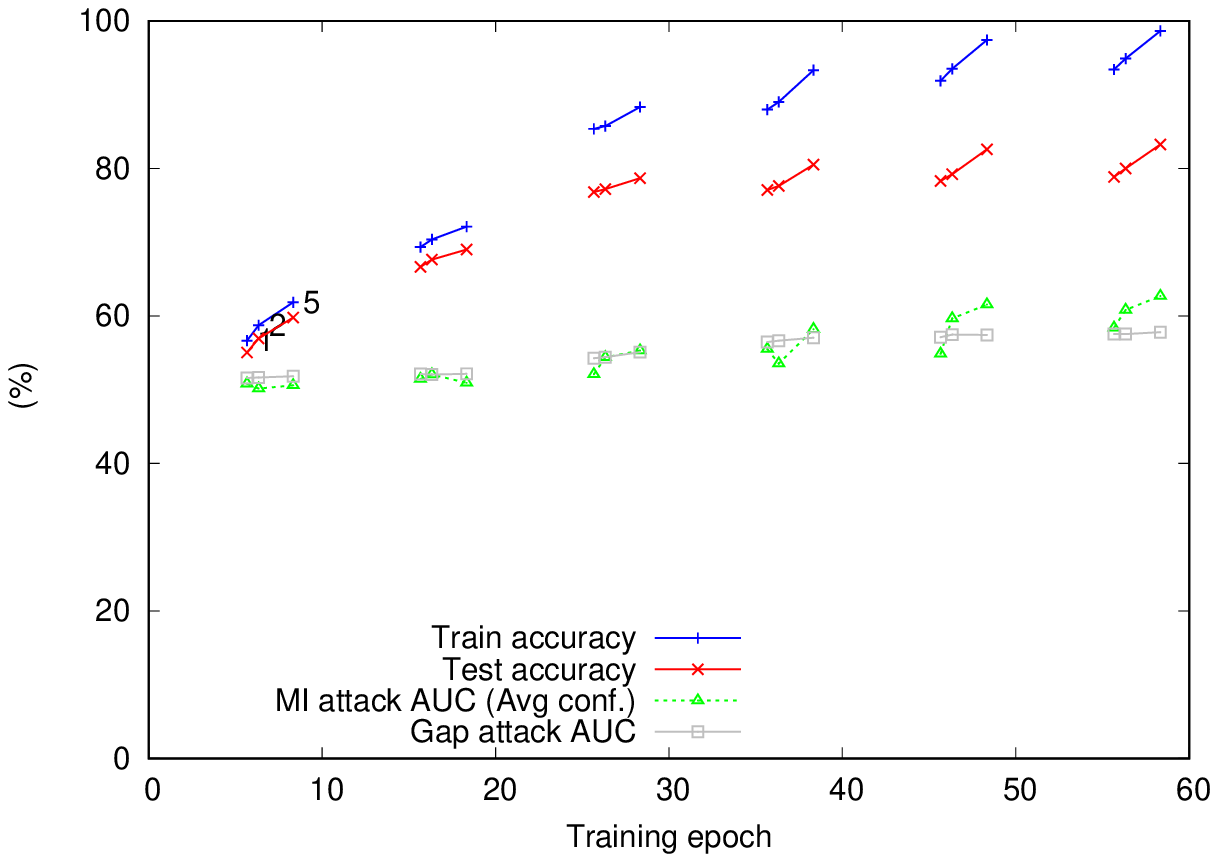}}
  & \subfloat[CIFAR100 (AlexNet)]{\includegraphics[width=0.29\linewidth]{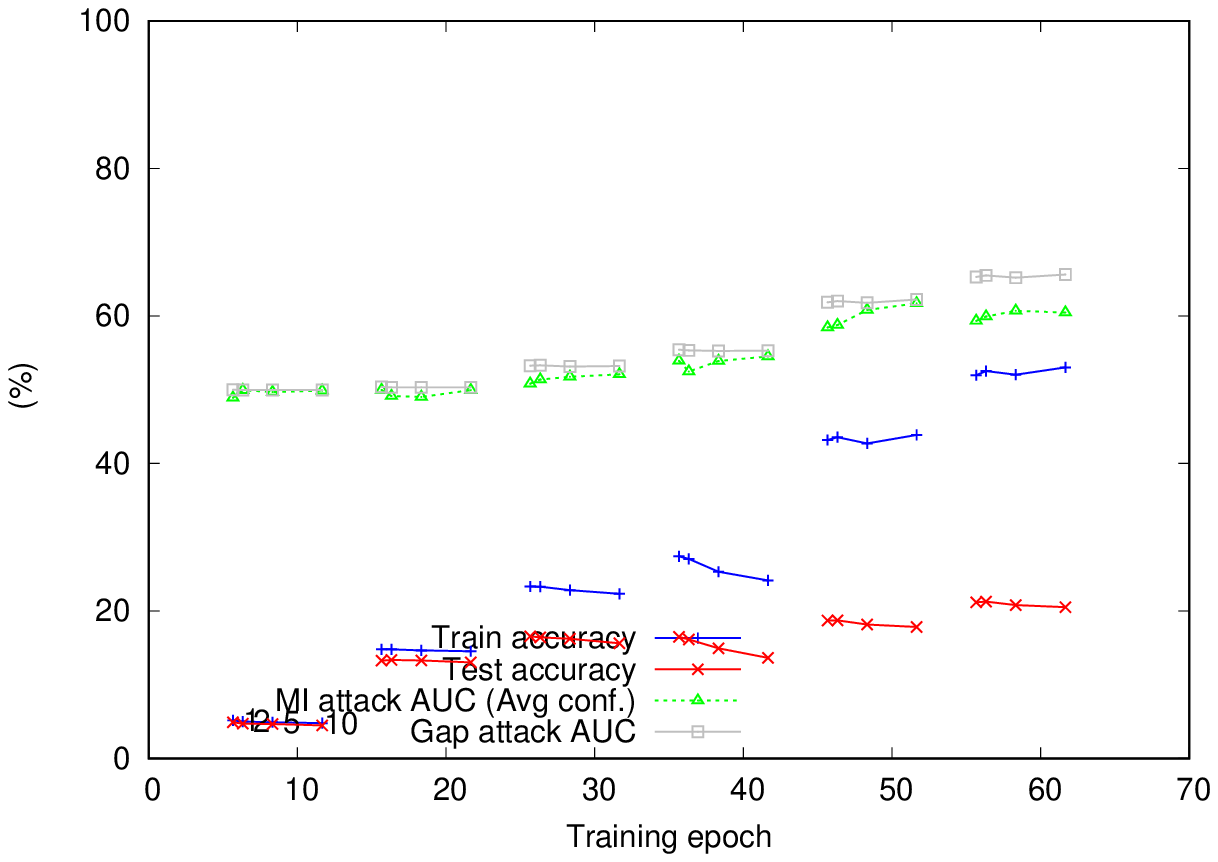}} \\
\subfloat[CIFAR100 (ResNet20)]{\includegraphics[width=0.29\linewidth]{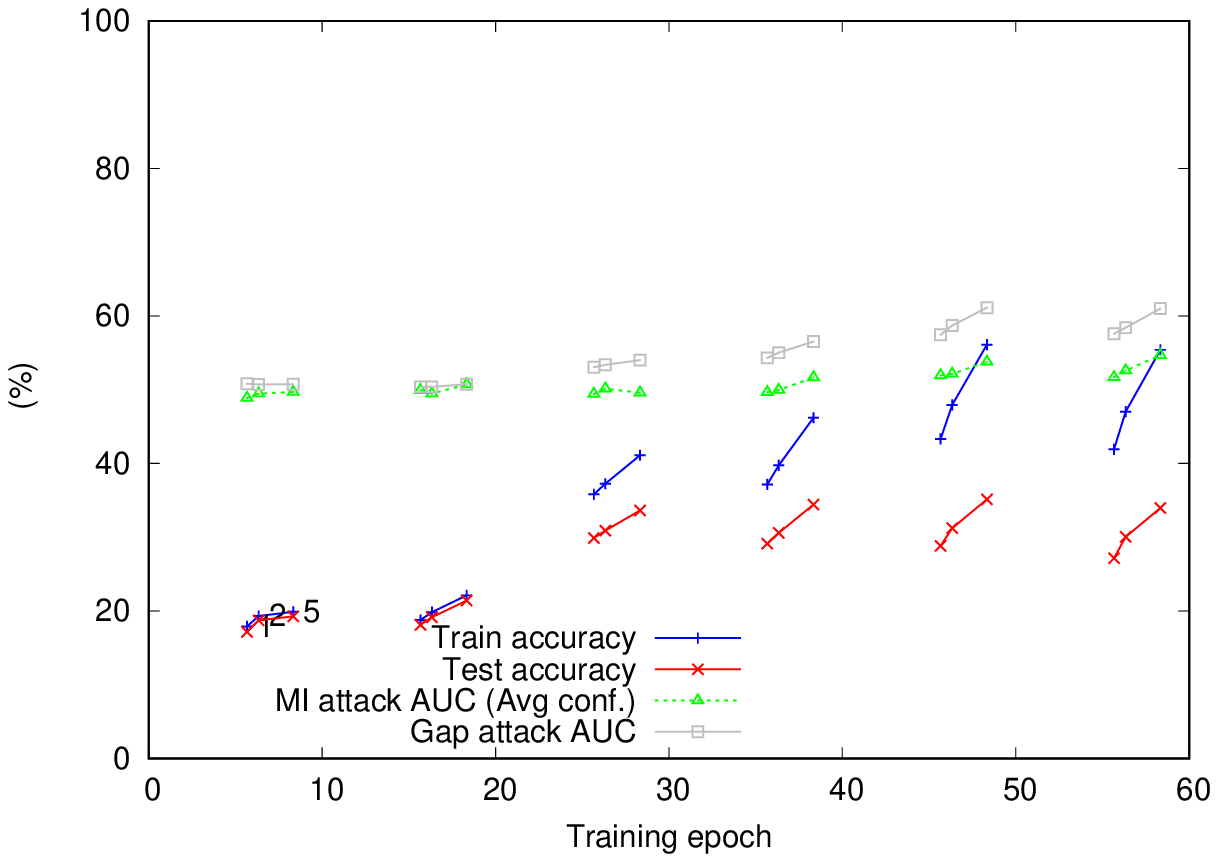}} 
  & \subfloat[SVHN (AlexNet)]{\includegraphics[width=0.29\linewidth]{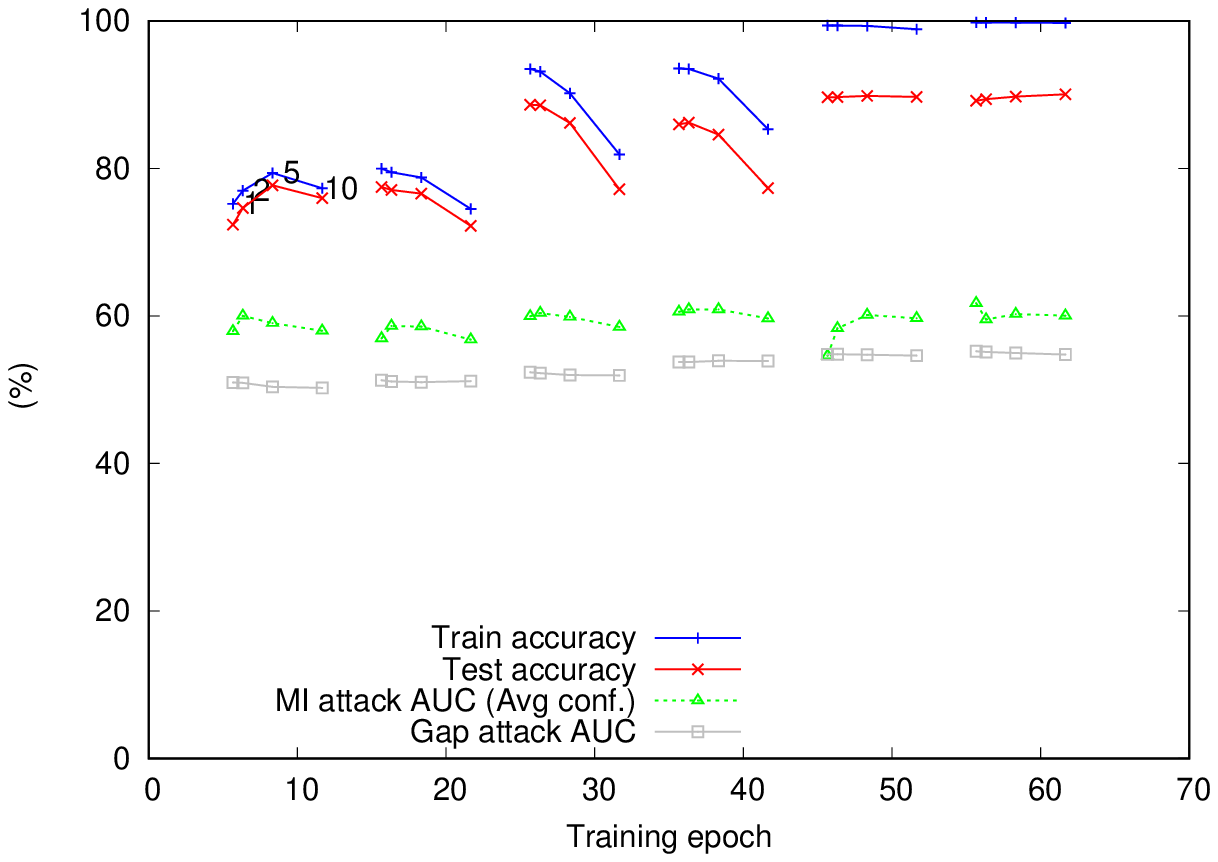}}
  & \subfloat[SVHN (ResNet20)]{\includegraphics[width=0.29\linewidth]{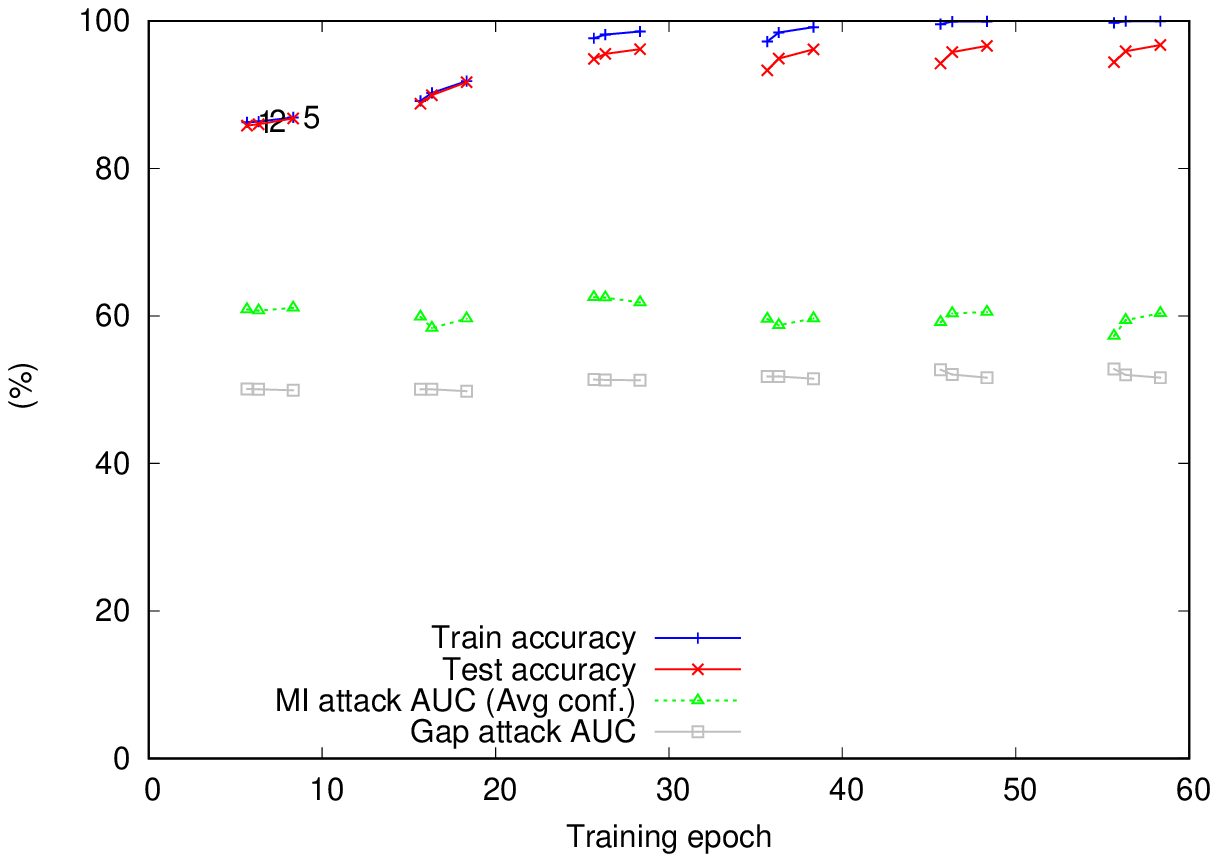}} \\
\end{tabular}
\end{tabularx}
\caption{Target models' accuracy and MI attacks' AUC across all datasets and models using diversified ensemble networks \cite{zhang2020diversified}.}
\label{fig-all-epochs-diversified}
\end{figure*}

We also conduct the same experiment with diversified ensemble networks \cite{zhang2020diversified}. The original paper used pre-trained VGG and ResNet models. We did not use pre-trained models for two reasons: 1) It make comparison with other approaches unfair, and 2) It may interfere with the membership inference analysis. We find that by using randomly initialized models to start training, $L_d$ varies significantly and prevents the optimization to converge. Therefore, we add a weight to the $L_d$ term to reduce its effect on the entire loss. We use 0.01 for CIFAR10 and SVHN and 0.001 for CIFAR100. For the shared layer, we use a fully-connected layer of size 128 followed by batch normalization and ReLU activation. We use SGD to train models for 60 epochs while dropping the learning at each 20 epochs by 0.1. We could not achieve the exact same results as reported in the paper for two main reasons: 1) we did not use pre-trained models in the ensemble, and 2) many hyper-parameters and implementation details are not reported in the original paper. We could not find a set of hyper-parameters and conditions to consistently achieve higher accuracy when increasing the number of models. This was also reported in the original paper where they found that more models in the diversified ensemble do not always improve accuracy. One penitential reason is that training base models in diversified neural networks are not independent. 
This means when the number of models in the diversified neural network is increased, there are significantly more parameters to train, but the number of epochs is constant. So, it is expected that diversified neural network with less models can sometimes outperform diversified neural network with more models if the number of training epoch is fixed. Nevertheless, as shown in Figure~\ref{fig-all-epochs-diversified}, in cases where accuracy increases, the membership inference attack effectiveness also increases.

\begin{figure*}
\def\tabularxcolumn#1{m{#1}}
\begin{tabularx}{\linewidth}{@{}cXX@{}}
\begin{tabular}{ccc}
\subfloat[CIFAR10 (AlexNet)]{\includegraphics[width=0.29\linewidth]{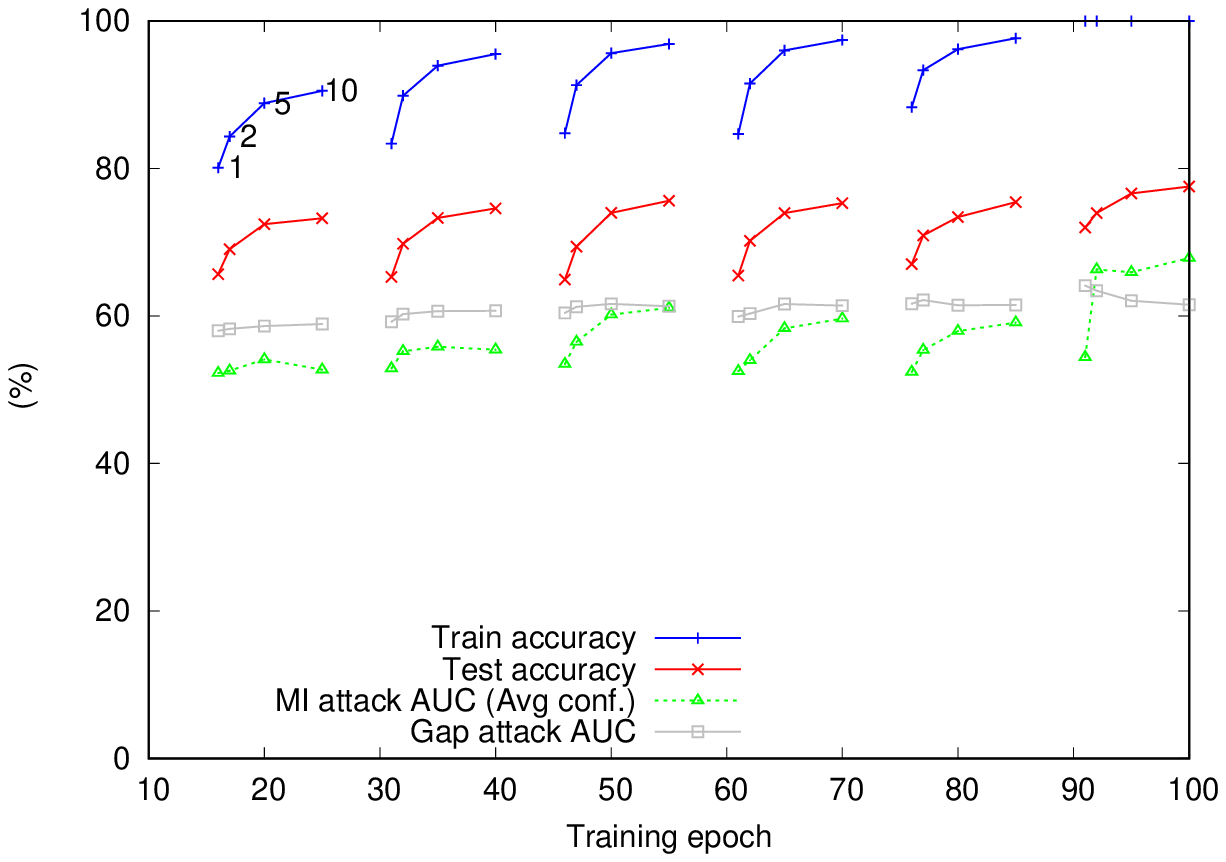}} 
  & \subfloat[CIFAR10 (ResNet20)]{\includegraphics[width=0.29\linewidth]{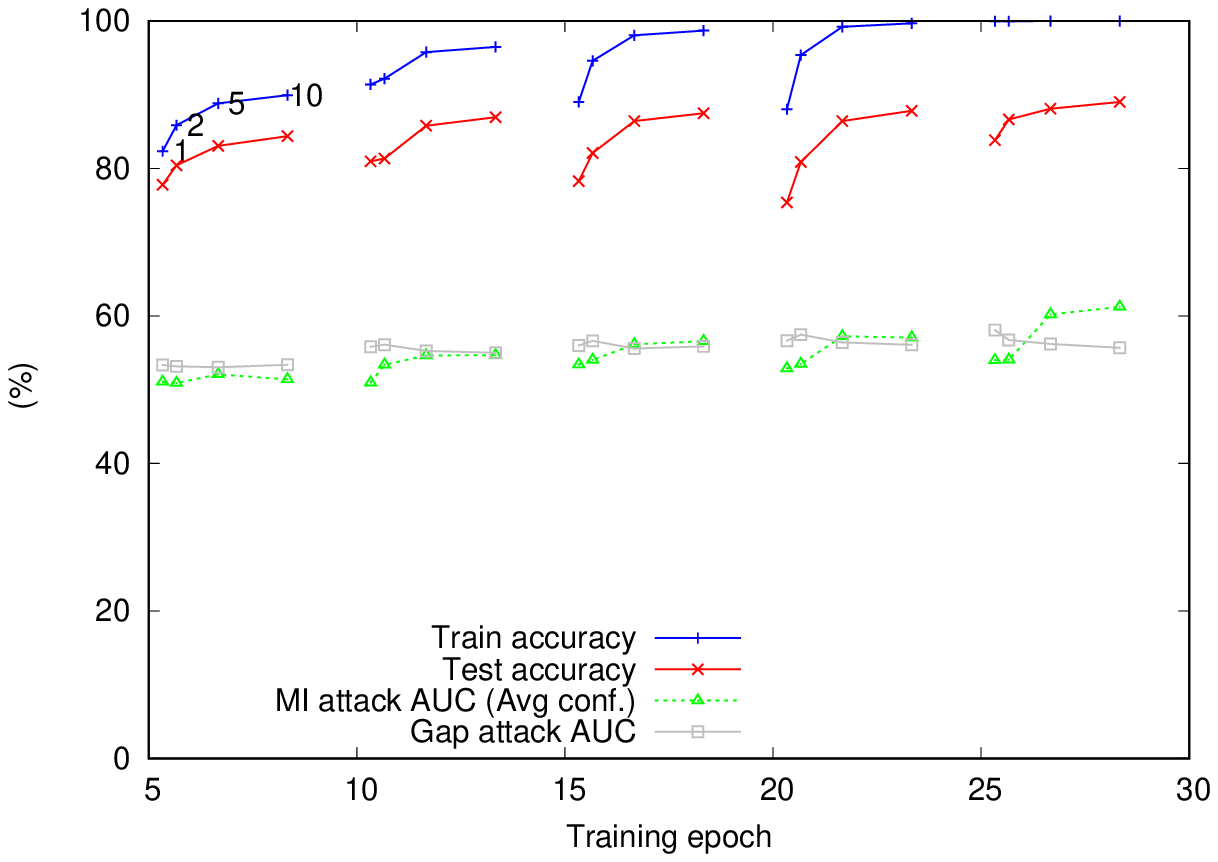}}
  & \subfloat[CIFAR100 (AlexNet)]{\includegraphics[width=0.29\linewidth]{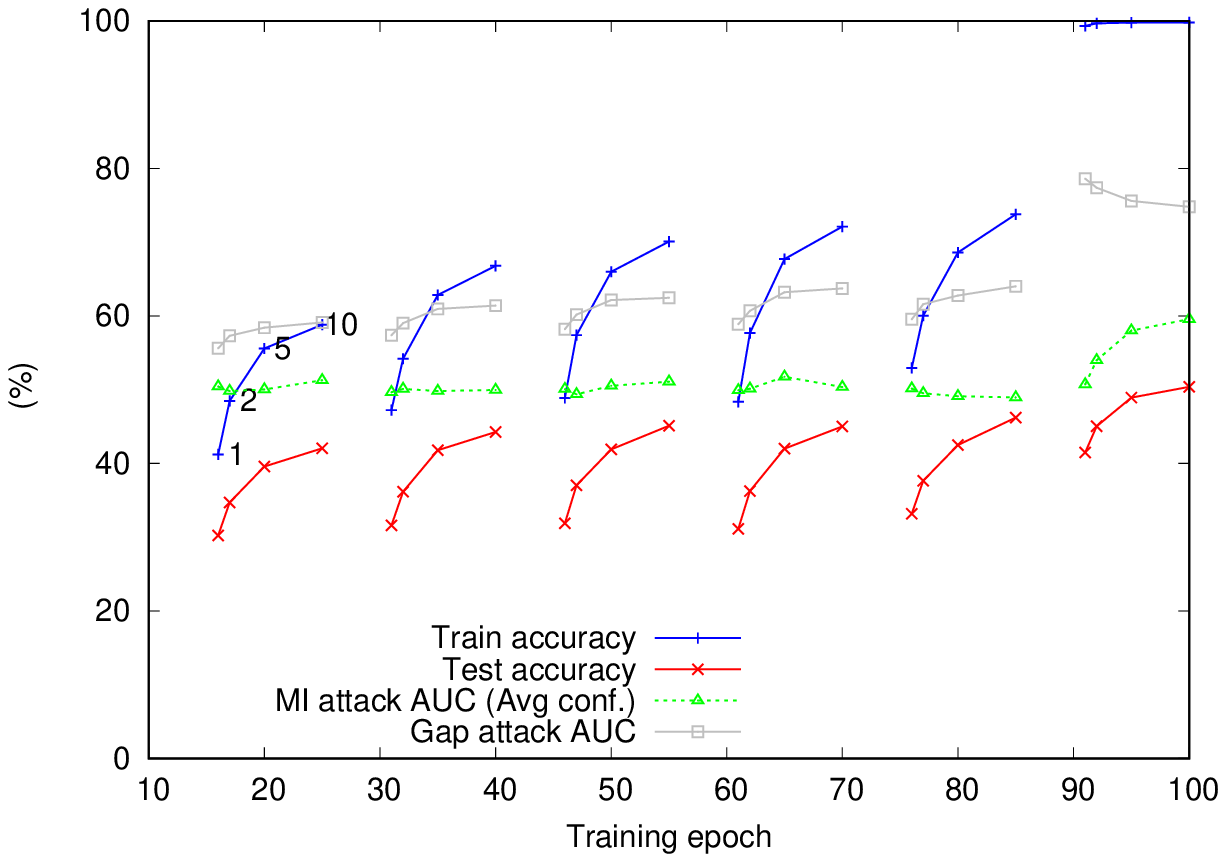}} \\
\subfloat[CIFAR100 (ResNet20)]{\includegraphics[width=0.29\linewidth]{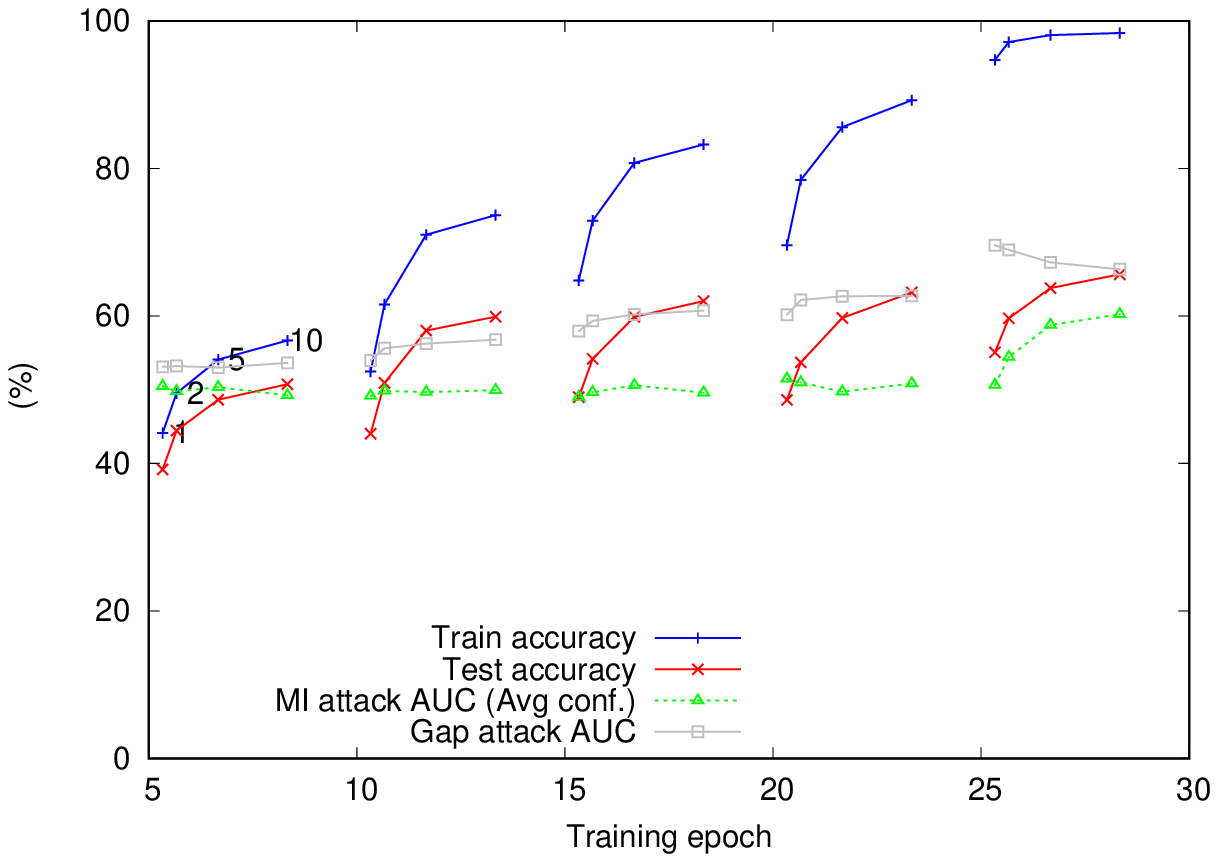}} 
  & \subfloat[SVHN (AlexNet)]{\includegraphics[width=0.29\linewidth]{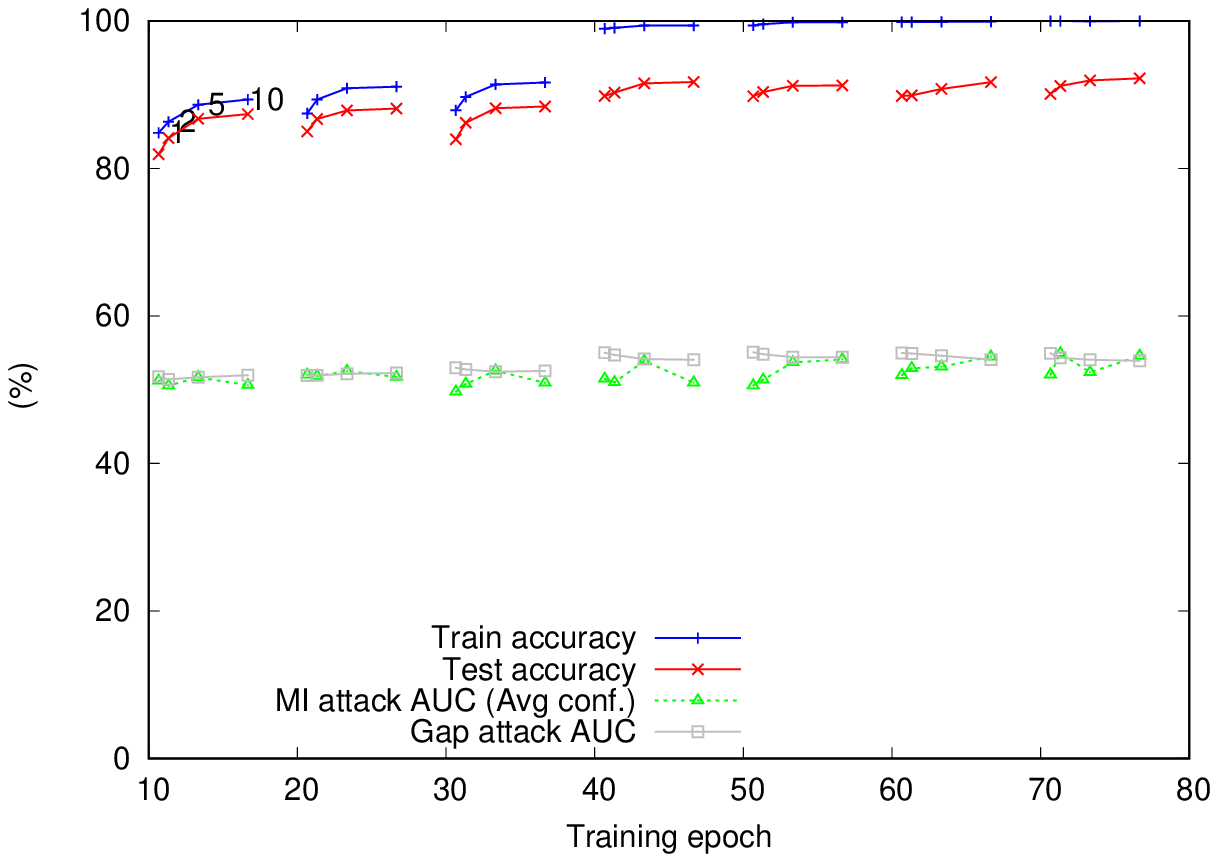}}
  & \subfloat[SVHN (ResNet20)]{\includegraphics[width=0.29\linewidth]{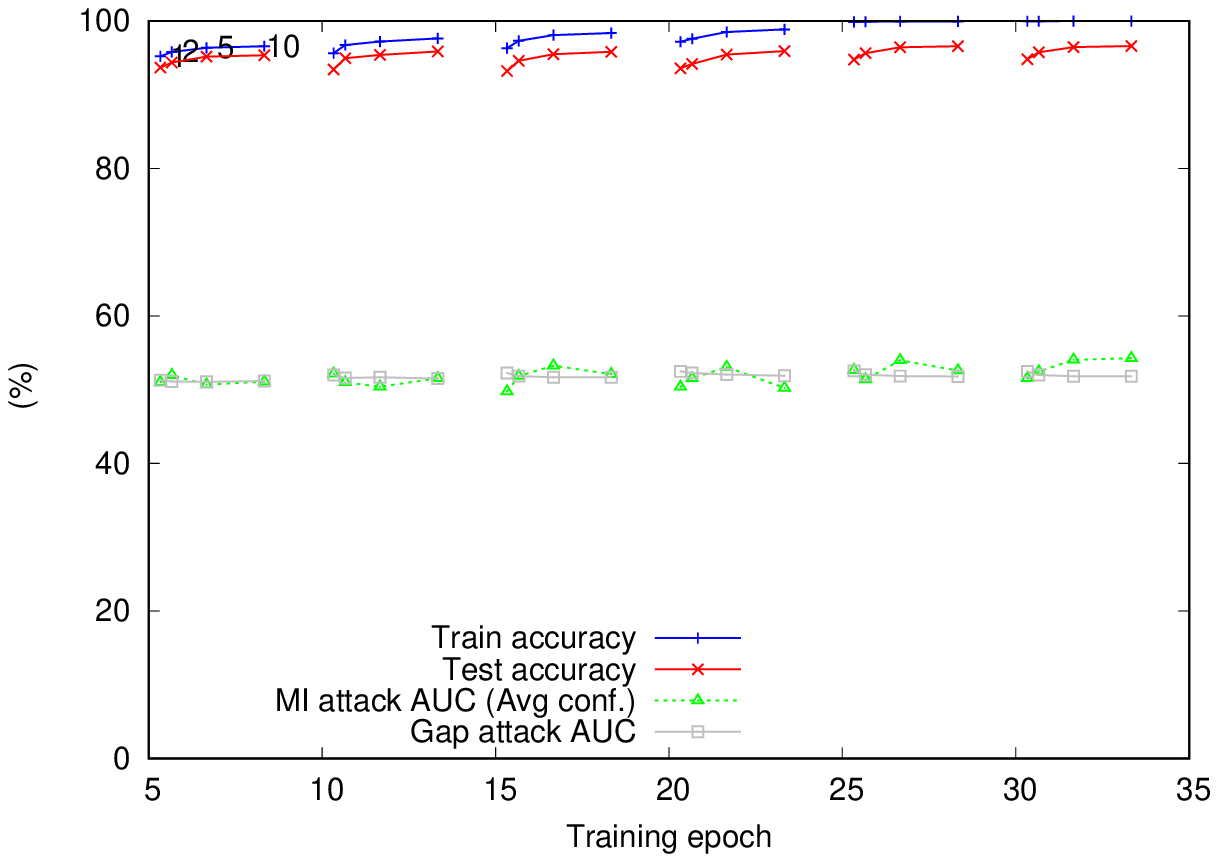}} \\
\end{tabular}
\end{tabularx}
\caption{Target models' accuracy and MI attacks' AUC across all datasets and models using weighted averaging ensemble networks.}
\label{fig-all-epochs-stacking}
\end{figure*}

\subsection{Weighted Averaging}
\label{appendix-all-epochs-stacking}

In this section, we evaluate weighted averaging of deep models. We focused on image CIFAR10, CIFAR100, and SVHN datasets. We trained each model with random initialization and all hyper-parameters similar to Section \ref{sec-setup}. Here, we use stochastic gradient descent (SGD) using the entire training set to learn the weight associated with each model. As shown in Figure~\ref{fig-all-epochs-stacking}, we observed similar accuracy-privacy trade-off.

\end{document}